\documentclass[sigconf]{acmart}

\usepackage{amssymb}
\usepackage{graphicx}
\usepackage{amsmath}

\usepackage{color}
\usepackage{bm}
\usepackage{algorithm}
\usepackage{algorithmic}
\usepackage{multirow}
\usepackage{subfigure}
\usepackage{amsthm}
\usepackage{booktabs}

\usepackage{mathrsfs}
\usepackage{makecell}
\usepackage{diagbox}
\usepackage{setspace}

\usepackage{balance}
\usepackage{float}
\usepackage{subeqnarray}
\usepackage{cases}
\usepackage{tabularx}
\usepackage{lscape}

\usepackage{mathtools}
\usepackage{wasysym}
\usepackage{arydshln} 
\usepackage{hyperref}

\hypersetup{colorlinks = true, 
	linkcolor = blue, 
	urlcolor = blue,
	citecolor = blue} 

\AtBeginDocument{%
  }

\copyrightyear{2023}
\acmYear{2023}
\setcopyright{acmlicensed}
\acmConference[MM '23] {Proceedings of the 31st ACM International Conference on Multimedia}{October 29--November 3, 2023}{Ottawa, ON, Canada.}
\acmBooktitle{Proceedings of the 31st ACM International Conference on Multimedia (MM '23), October 29--November 3, 2023, Ottawa, ON, Canada}
\acmISBN{979-8-4007-0108-5/23/10}
\acmDOI{10.1145/3581783.3612323}

\begin{document}

\title{A Generalized Physical-knowledge-guided Dynamic Model for Underwater Image Enhancement}

\author{Pan Mu}
\affiliation{
	\institution{College of Computer Science and Technology, Zhejiang University of Technology}
	\city{}
	\country{}
}
\email{panmu@zjut.edu.cn}

\author{Hanning Xu}
\affiliation{
	\institution{College of Computer Science and Technology, Zhejiang University of Technology}
		\city{}
	\country{}
}
\email{hanningxu@zjut.edu.cn}

\author{Zheyuan Liu}
\affiliation{
	\institution{College of Computer Science and Technology, Zhejiang University of Technology}
		\city{}
	\country{}
}
\email{zheyuanliu@zjut.edu.cn}

\author{Zheng Wang}
\affiliation{
	\institution{College of Computer Science and Technology, Zhejiang University of Technology}
		\city{}
	\country{}
}
\email{zhengwang@zjut.edu.cn}

\author{Sixian Chan}
\affiliation{
	\institution{College of Computer Science and Technology, Zhejiang University of Technology}
		\city{}
	\country{}
}
\email{sxchan@zjut.edu.cn}

\author{Cong Bai}
\authornote{Corresponding author.}
\affiliation{
	\institution{College of Computer Science and Technology, Zhejiang University of Technology}
		\city{}
	\country{}
}
\email{congbai@zjut.edu.cn}


\begin{teaserfigure}
	\includegraphics[width=1\textwidth]{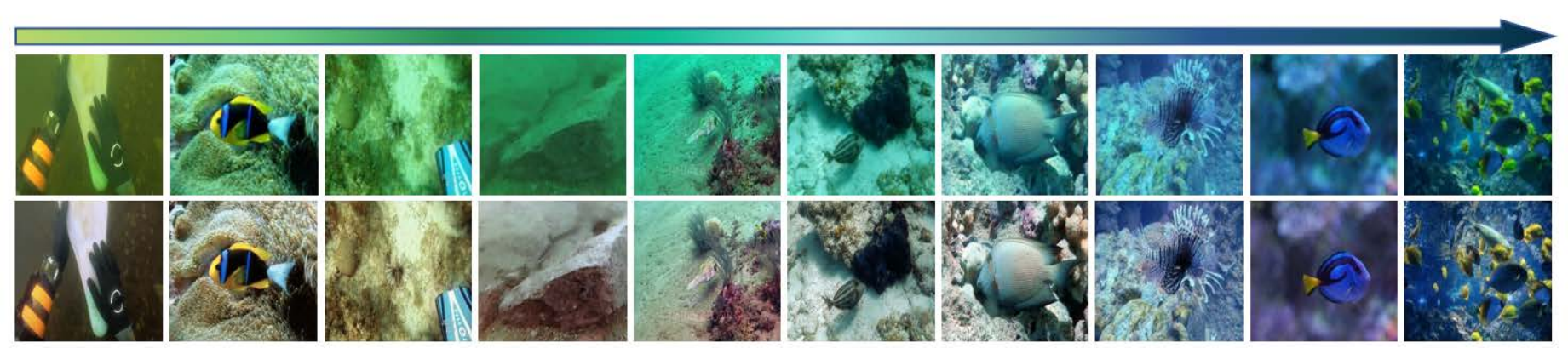}
	\caption{Visual results of our GUPDM on various underwater image types. The first row displays a series of degraded underwater images ranging from yellow to blue and the second row shows the enhanced results with our method. }\label{fig:first_figure} 
\end{teaserfigure}

\begin{abstract}
Underwater images often suffer from color distortion and low contrast resulting in various image types, due to the scattering and absorption of light by water. While it is difficult to obtain high-quality paired training samples with a generalized model. To tackle these challenges, we design a Generalized Underwater image enhancement method via a Physical-knowledge-guided Dynamic Model (short for GUPDM), consisting of three parts: Atmosphere-based Dynamic Structure (ADS), Transmission-guided Dynamic Structure (TDS), and Prior-based Multi-scale Structure (PMS). In particular, to cover complex underwater scenes, this study changes the global atmosphere light and the transmission to simulate various underwater image types (e.g., the underwater image color ranging from yellow to blue) through the formation model. We then design ADS and TDS that use dynamic convolutions to adaptively extract prior information from underwater images and generate parameters for PMS. These two modules enable the network to select appropriate parameters for various water types adaptively. Besides, the multi-scale feature extraction module in PMS uses convolution blocks with different kernel sizes and obtains weights for each feature map via channel attention block and fuses them to boost the receptive field of the network. The source code will be available at  \href{https://github.com/shiningZZ/GUPDM}{https://github.com/shiningZZ/GUPDM}. 
	
\end{abstract}



\maketitle

\section{Introduction}
Due to the scattering and absorbing effects of the water on light, underwater images generally suffer from low contrast, color shift and blur. The quality of underwater images can directly or indirectly affect the accuracy of underwater robots in performing tasks such as detection, segmentation, tracking and classification. Therefore, Underwater Image Enhancement (UIE) plays an important role in underwater tasks. 

The existing UIE methods can be coarsely divided into two categories: traditional and deep learning-based ones. Traditional UIE methods includes prior-based \cite{galdran2015automatic,drews2013UDCP,2018formation,2012wavelength,2016histogram_distribution,akkaynak2019sea,2019Sea-thru,2017light_absorption,2017attenuation-curve} and model-free \cite{2012fusion,ghani2015contrast,ancuti20193C,2010pixel-range-stretching,2017multi-scale_retinex} approaches. Prior-based ones utilize rich priors and estimate the parameters of underwater image formation model to generate the enhanced images. 
For example,~\cite{drews2013UDCP,2018formation,2012wavelength,2016histogram_distribution} focus on the estimation of medium transmittance to restore underwater images. Although these methods make full use of prior information, they are less practical in complex scenarios, which often leads to over-enhancement of images and thus cannot be applied to real underwater scenes. Model-free methods often rely on the spatial relationship between pixel values of the original underwater image to improve the brightness, contrast and saturation of the image, such as Gray World \cite{1990GrayWorld}, Max RGB \cite{1977MaxRGB} and White Balance\cite{liu1995whitebalance}. However, these methods tend to ignore the details and depth information of the image, resulting in artifacts and poor adaptation to the complex underwater degradations.

Deep learning has introduced new strategies for UIE task~\cite{li2020UWCNN,fu2022puieNet,li2021Ucolor,2022USUIR,Mu2022StructureInferredBM,guo2022URanker,huang2023Semi-UIR,jiang2022topal,liu2022tacl,lin2021global,jiang2022bilevel,lin2021global}. 
Those methods can achieve complex non-systematic end-to-end modeling or combine physical priors with networks to solve the existing problems. 
These methods have better feature representation capacity, benefiting from large data they utilized to train. 
Hypernet~\cite{ha2016hypernetworks,lin2020MTL,yin2022conditional,liu2020generic,liu2022general} is one of the great works among  deep learning-based approaches, which uses a relatively small network to generate weights for target network. These methods have better feature representation capacity, as they leverage large data to train. 
However, the main challenge is the high cost and difficulty of acquiring large-scale underwater datasets, which forces most methods to use small-scale datasets. 
Consequently, many GAN-based approaches focus on unsupervised UIE~\cite{li2019FGAN,guo2019DenseGAN,Islam2020FUnIE-GAN,Fabbri2018UGAN,jiang2022bilevel} models, aiming to synthesize underwater images by learning a realistic representation of underwater conditions from unlabeled images. Nevertheless,  a significant gap  between the synthesized underwater images and the real ones still remains, in terms of plausibility and scene diversity.

\begin{figure}
	\includegraphics[width=0.48\textwidth]{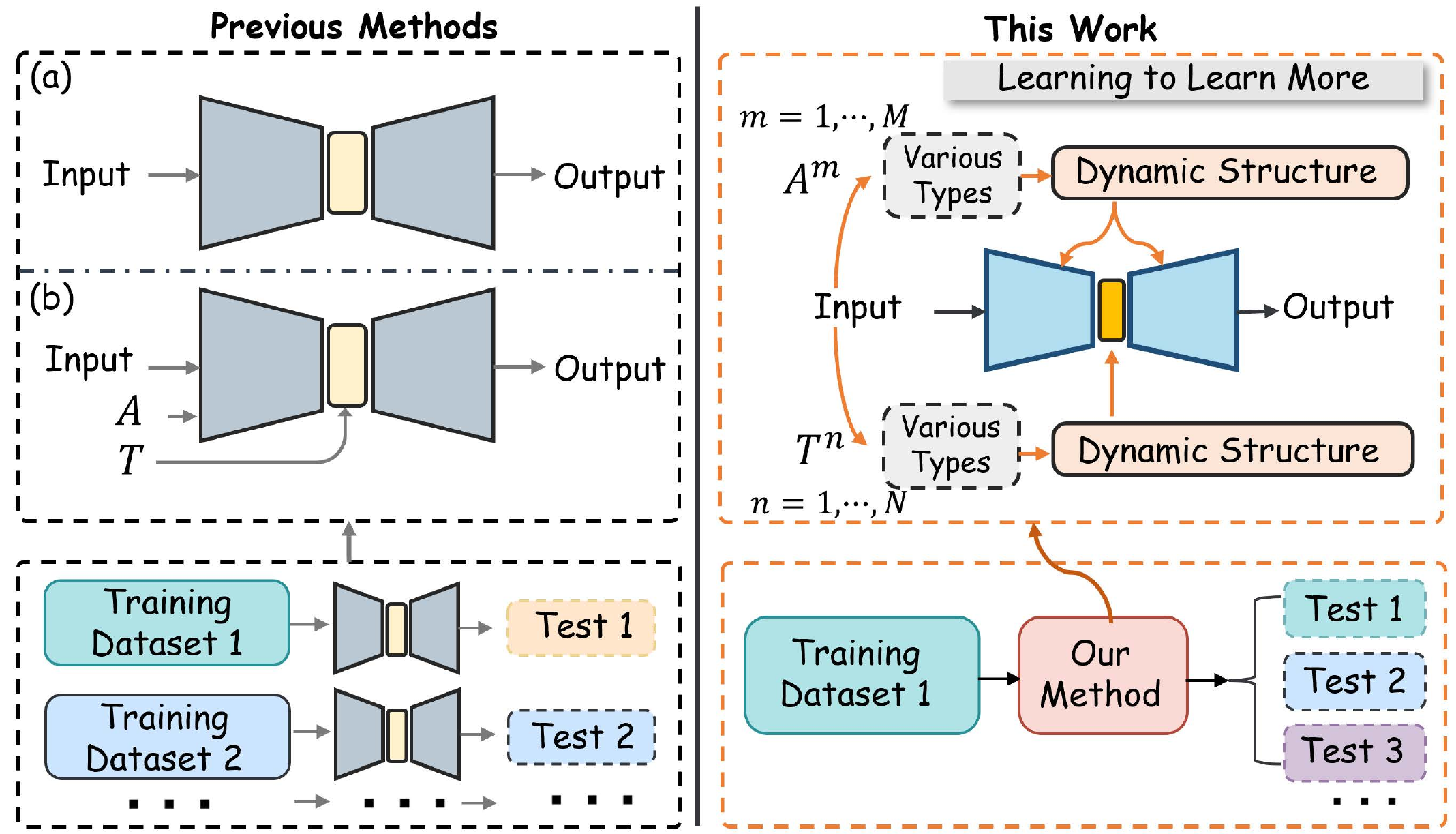}
	\caption{Schematic description of previous deep-learning methods (left) and our model (right).}\label{fig:Comp_Methods} 
\end{figure}

Overall, we summarize the challenges existing in underwater image enhancement as follows. \textbf{1)} Underwater images often suffer from color distortion and low contrast, due to the scattering and absorption of light by water. Moreover, the water quality and the distance of light transmission also affect the image clarity, making the underwater images blurry. \textbf{2)} The scarcity of high-quality paired training samples constrains the performance of learning-based models. \textbf{3)} Model generalization is a critical task for UIE task, which is important but has been neglected by many researches, as shown in Fig.~\ref{fig:Comp_Methods}. For example, a model obtained through training one dataset maybe not suitable for another water types. 

\begin{figure*}
	\includegraphics[width=0.95\textwidth]{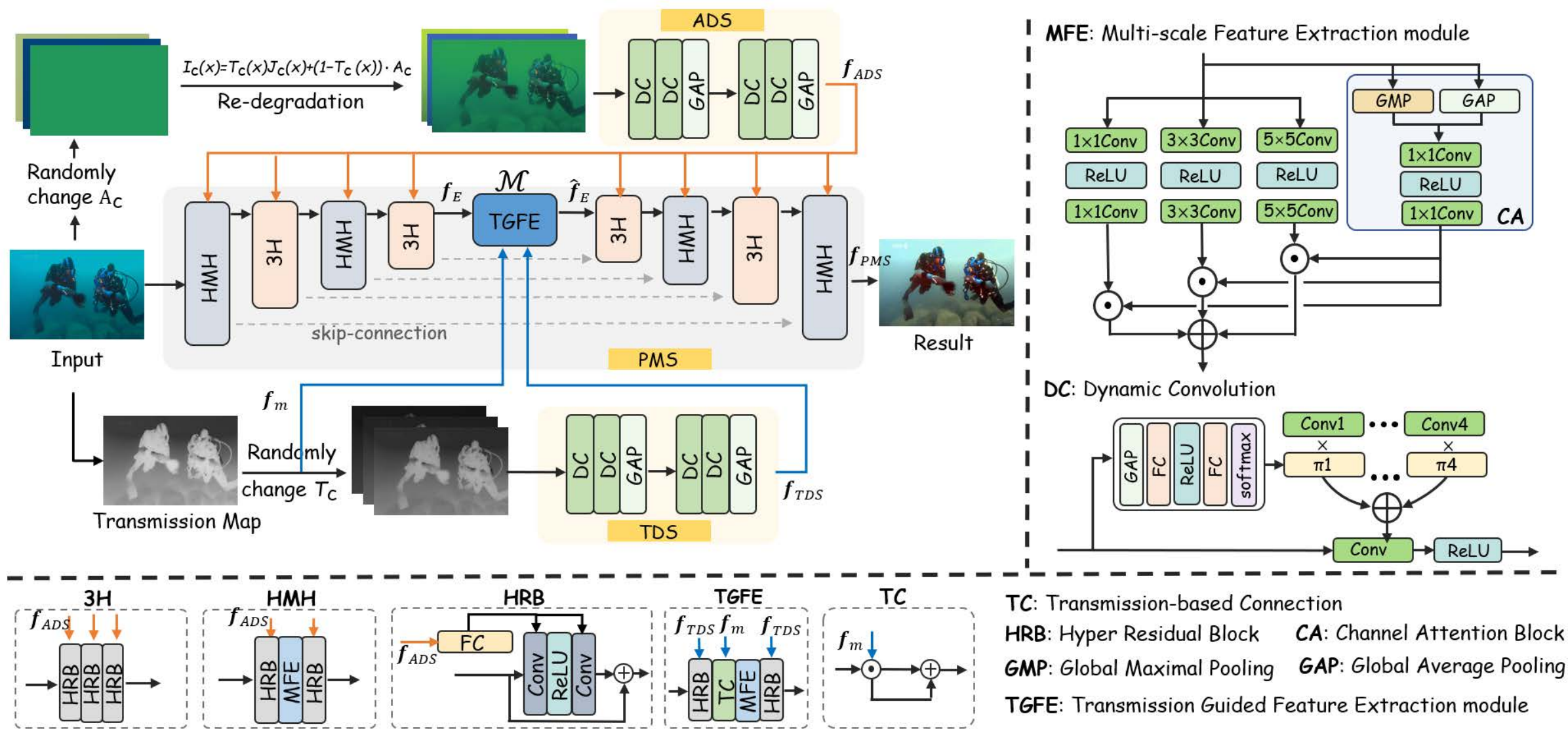}
	\caption{ Overall illustration of the proposed GUPDM framework. GUPDM is composed of three components: Atmosphere-based Dynamic Structure (ADS), Transmission-guided Dynamic Structure (TDS) and Prior-based Multi-scale Structure (PMS). }\label{fig:framework} 
\end{figure*}

\textbf{Contributions:} To tackle these challenges, this work designs a \textbf{G}eneralized \textbf{U}nderwater image enhancement method via \textbf{P}hysical-knowledge-guided \textbf{D}ynamic \textbf{M}odel (short for \textbf{GUPDM}), consisting of three parts: Atmosphere-based Dynamic Structure (ADS), Transmission-guided Dynamic Structure (TDS) and Prior-based Multi-scale Structure (PMS), as shown in Fig.~\ref{fig:framework}. 
In particular, to cover complex underwater scenes, this study varies the global atmosphere and the transmission,  reaching various underwater image types (e.g., the underwater image color ranging form yellow to blue) through the formation model. Then, we design a ADS that uses dynamic convolutions to adaptively extract prior information of underwater images and generate parameters for PMS. 
In addition, to force our model to pay more attention to various details in images, this work introduces a TDS to enable our network to adaptively select suitable parameters. Thus, the whole network can  obtain appropriate parameters according to the water type of underwater image, making the proposed model more robust. Besides, the Multi-scale Feature Extraction (MFE) module in PMS uses convolution blocks with different kernel sizes and obtain weights for each feature map via channel attention block and then fuses them to boost the receptive field of network. 
We summarize the main contributions as follows:

\begin{itemize}
	\item This work designs a generalized underwater image enhancement method via physical-knowledge-guided (i.e., various atmosphere and transmission) dynamic model to adaptively enhance the underwater images with different water types (as shown in Fig.~\ref{fig:first_figure}). 

	\item To cover complex underwater scene, by varying global atmosphere light, we design a ADS that uses dynamic convolutions to adaptively extract prior information of underwater images and generates parameters for base network structure. This also allows the entire model to adjust the parameters according to the degradation level of input image, improving the generalization capacity of network. 
	
	\item The proposed TDS uses medium transmittance-based prior to prompt the network to pay more attention to  areas with the most quality degradation, enabling the network to  adaptively select appropriate parameters according to water quality. 
	
	\item Extensive experiments demonstrate that our method achieves superior performance and the best generalization capacity, reaching the state-of-the-art level on multiple test datasets.
\end{itemize}

\section{Related Works}
We briefly review previous related works regarding the traditional UIE models and deep learning-based UIE approaches.

\textbf{Traditional UIE Models:} 
The Traditional UIE Models can be summarized as prior-based approaches and model-free approaches. The prior-based approaches are grounded on physical model intended to estimate the parameters of underwater image formation model by using visual cues and then reconstruct a clean image by applying these parameters in reverse. For instance, in~\cite{galdran2015automatic}, researchers propose a variant of DCP which uses red channel information to estimate the transmission map of underwater images.  UDCP~\cite{drews2013UDCP} estimates the transmission map by considering the blue and green color channels. Moreover, Sea-thru~\cite{akkaynak2019sea} estimates the back scatter and attenuation coefficient by using RGBD images as input. 

Another type of traditional methods is model-free approach which improve the chromatic aberration and contrast by modifying the overall pixel values of  underwater images, including histogram equalization \cite{pizer1990contrast}, white balance \cite{liu1995whitebalance} and Retinex \cite{rahman1996retinex}. To fully enhance the image details and colors, \cite{ghani2015contrast} proposes a method that integrates global and local contrast stretching. To handle the artifacts caused by the severely uneven color spectrum distribution, 3C~\cite{ancuti20193C} reconstructs the lost channel based on opponent color. These traditional methods can improve the visual effect to some extent. However, these methods may result in color distortions and artifacts when encountered with sophisticated illumination conditions.

\textbf{UIE Approaches based on Deep-learning:} 
Deep learning methods are trained on large scale underwater images and can automatically extract relevant features from them to improve the quality of enhanced images. For example, methods use Generative Adversarial Network (GAN) for image enhancement includes FGAN~\cite{li2019FGAN}, DenseGAN~\cite{guo2019DenseGAN}, UGAN~\cite{Fabbri2018UGAN} and FUnIE-GAN~\cite{Islam2020FUnIE-GAN}. The main purpose of GAN strategy is to expand the source of pairing data through generative networks, but there is still a lack of high-quality training samples that truly match real underwater scenarios and diverse degradation. Due to the nonavailability of ground truth high-quality images, a novel probabilistic network PUIE-Net~\cite{fu2022puieNet} is proposed to learn the enhanced distribution of degraded underwater images. These end-to-end methods can produce visually pleasing results. However, they usually require a specific model for each dataset and lack generalization ability and flexibility in handling different underwater scenarios, due to the complexity of underwater environments. The methods trained on real data can produce visually pleasing results. However, they cannot restore the color and structure of specific objects well and tend to produce inauthentic results since the reference images are not the actual ground truths. 
To tackle these problems, some models~\cite{liu2019compounded,liu2018learning,li2020UWCNN,li2021Ucolor,qian2022real} integrate the priors like transmission map and atmosphere light to deal with the environmental information. 
For instance, UWCNN~\cite{li2020UWCNN} is trained on synthesize datasets with underwater scene prior. Instead of estimating the parameters of underwater imaging model, UWCNN directly reconstructs clear enhanced images. Ucolor~\cite{li2021Ucolor} uses medium transmission-guided multi-color embedding to solve the color casts and low contrast issues. 
DDNet~\cite{qian2022real} uses medium transmission maps and global atmosphere light to form the haze and scene-adaptive convolutions.

\section{Our Developed Framework}
This section detailed introduces the developed framework as shown in Fig.~\ref{fig:framework}. Subsection~\ref{subsec:motivation} first present the motivation and the problem formulation of this work. Subsection~\ref{subsec:network} describes the detailed structure of GUPDM and Subsection~\ref{subsec:loss} presents the loss functions and training procedure. 

\subsection{Motivation and Problem Formulation}\label{subsec:motivation}

Following Akkaynak’s light scattering model~\cite{akkaynak2019sea}, the degraded underwater images can be expressed as 
\begin{equation}\label{eq:degraded_model}
	\mathbf{I}_{c}(x)=\mathbf{J}_{c}(x)e^{-\bm{\beta} d(x)}+\left(1-e^{-\bm{\beta} d(x)}\right)  \mathbf{A}_{c}, c \in\{r, g, b\}
\end{equation}
where $x$ indicates the spatial location of each pixel, $\mathbf{I}_{c}$ is the observed image, $\mathbf{J}_{c}$ is the restored haze-free image and $\mathbf{A}_{c}$ means the global background light, $d(x)$ is the scene depth at pixel $x$ and $\bm{\beta}$ is the channel-wise extinction coefficient depending on the water quality. 
$\mathbf{T}_c(x):=e^{-\bm{\bm{\beta}} d(x)}$ is the medium transmission map representing the percentage of scene spoke brightness that reaches the camera after reflection from point $x$ in the underwater scene, which also reflect the water type.

\begin{figure}
	\includegraphics[width=0.45\textwidth]{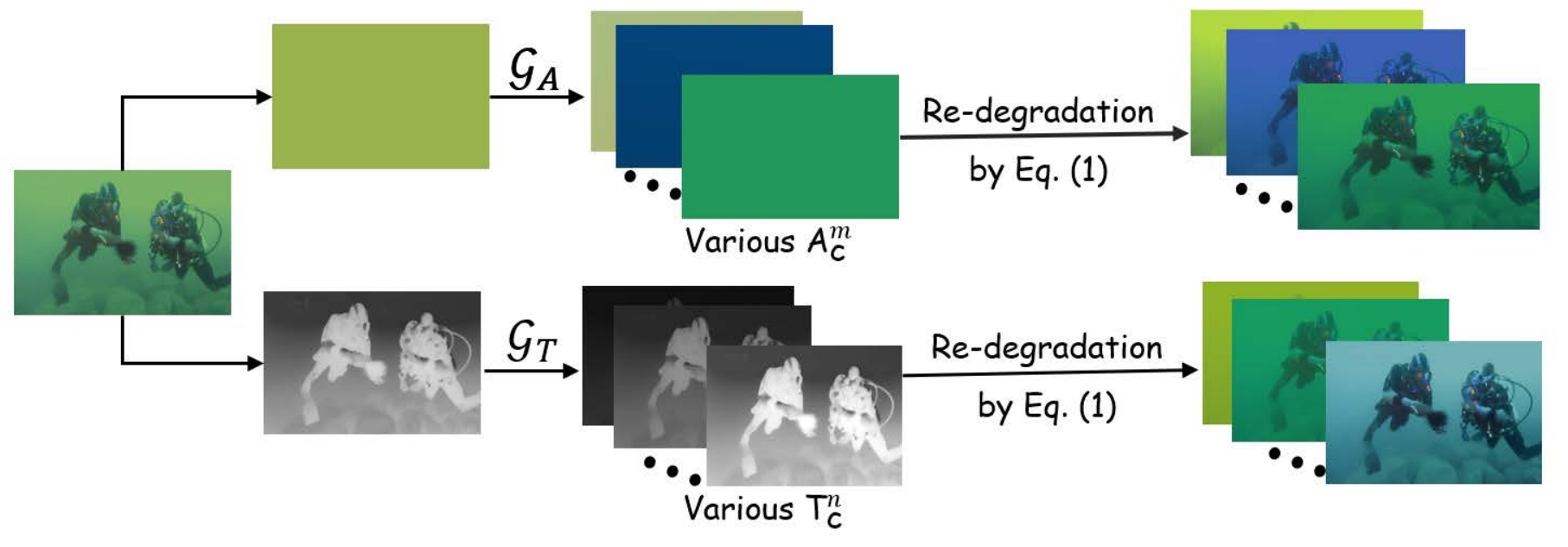}
	\caption{Generating various re-degraded underwater images. 
	}\label{fig:Re_degradation} 
\end{figure}

Due to the color cast caused by varying light attenuation with different wavelengths and haze effect caused by scattering, there exist various types of degraded underwater images (see Fig.~\ref{fig:Re_degradation}). 
In other words, we can roughly simulate complex underwater scene with the global atmosphere light $\mathbf{A}_c$ and the transmission map $\mathbf{T}_c$. Inspired by aforementioned information, we change the original input underwater images by adjusting the two parameters. 
We aim to reduce the color cast of underwater images through making such changes so that our encoders can learn more information about the color. This operation can also help to augment the data to better handle the deficient data problem of underwater image. Thus, we generate the input images as is shown in Fig.~\ref{fig:Re_degradation}. We summarize the re-degradation types as $\mathbf{I}_{c}(x)\stackrel{(a),(b)}{\longrightarrow}\mathbf{I}_{c}^{m,n}(x)$, where
\begin{itemize}
	\item[(a).] \textit{Varing $\mathbf{A}_c\to\mathbf{A}_c^m$, based on Eq.~\eqref{eq:degraded_model}, we obtain $\mathbf{I}_{c}^m(x)$};
	\item[(b).] \textit{If we further vary $\mathbf{T}_c(x)\to\mathbf{T}_c^n(x)$, we obtain $\mathbf{I}_{c}^{n}(x)$}.
\end{itemize}

Motivated by the above analysis, this work aims to restore various underwater images using a generalized model. Therefore, the key is to enable the designed network ot estimate the features of varied types. We design an Atmosphere-based Dynamic Structure $\mathcal{A}$ (i.e., ADS) and a Transmission-based  Dynamic Structure  $\mathcal{T}$ (i.e., TDS) to jointly guide the Prior-based Multi-scale Structure $\mathcal{N}$ (i.e., PMS). Actually, the $\mathcal{N}$ acts as a base network structure for UIE while $\mathcal{A}$ and $\mathcal{T}$ are two hyper-guided modules with different level. 
Thus, we formulate this problem in the following forms:
\begin{equation}\label{eq:bi-level}
	\begin{array}{l}
		\min\limits_{\bm{\theta},\bm{\phi}}L\left(\widehat{\mathbf{J}}_c(x),\bar{\mathbf{J}}_c(x);\mathbf{I}_{c}^{m}(x),\mathbf{I}_{c}^{n}(x)\right)\\	
		s.t., \bm{\omega}\in\arg\min\limits_{\bm{\omega}}F(\widetilde{\mathbf{J}}_c(x);\mathbf{I}_{c}(x)),
	\end{array}
\end{equation}
where $\bm{\theta},\bm{\phi}$ and $\bm{\omega}$ are the parameters of module ADS, TDS and PMS, respectively; $L$ and $F$ are the loss functions; $\widetilde{\mathbf{J}}_c(x)$, $\widehat{\mathbf{J}}_c(x)$ and $\bar{\mathbf{J}}_c(x)$ are respectively estimated by:
\begin{equation*}
	\begin{array}{l}
		\widetilde{\mathbf{J}}_c(x)\leftarrow \mathcal{N}(\bm{\omega};\mathcal{A}(\bm{\theta},\mathbf{A}_{c}),\mathcal{T}(\bm{\phi},\mathbf{T}_{c});\mathbf{I}_c(x)),\\
		\widehat{\mathbf{J}}_c(x)\leftarrow \mathcal{N}(\bm{\omega};\mathcal{A}(\bm{\theta},\mathbf{A}_{c}^{m}),\mathcal{T}(\bm{\phi},\mathbf{T}_{c}(x);\mathbf{I}_{c}(x)),\\
		\bar{\mathbf{J}}_c(x)\leftarrow\mathcal{N}(\bm{\omega};\mathcal{A}(\bm{\theta},\mathbf{A}_{c}^{m}),\mathcal{T}(\bm{\phi},\mathbf{T}_{c}^{n}(x);\mathbf{I}_{c}(x)).
	\end{array}
\end{equation*}

\subsection{Physical Knowledge-guided Dynamic Network}\label{subsec:network}

In this part, we  introduce the developed model in detail, as shown in Fig.~\ref{fig:framework}. The proposed GUPDM mainly consists of three parts: atmosphere-based dynamic structure, transmission-guided dynamic structure and prior-based multi-scale structure. ADS and TDS use dynamic convolutions to adaptively extract prior information from underwater images and generate parameters for PMS. 
We combine the meta-learning strategy to train our physical knowledge-based networks. This procedure can prompt our GUPDM to attend to different water types. Thus, the whole network can adaptively select appropriate parameters according to the water types of input images, achieving more robustness and generalization capacity.

\textbf{Atmosphere-guided Dynamic Structure.} 
With the degraded underwater model in Eq.~\eqref{eq:degraded_model}, this work varies the atmosphere light $\mathbf{A}_c$ to different values, i.e., $\mathbf{A}_c^m,\ m=1,\cdots, M$, obtaining $M$ degraded underwater types:
\begin{equation*}
	\mathbf{I}_c^m=\mathcal{G}_\mathbf{A}(\mathbf{A}_c)=\lambda_{c}^m\mathbf{A}_{c}. 
\end{equation*}
where $c \in \{r, g, b\}$ and $\lambda_{c}^m\in (0.3,0.6)$ is randomly generated degradation level. 
Then, to adaptively extract atmosphere-based prior information, we design ADS (i.e., $\mathcal{A}(\bm{\theta})$) with dynamic convolutions. We formulate this procedure as:
\begin{equation*}
	\mathbf{f}_{\text{ADS}}= \mathcal{A}(\bm{\theta},\mathbf{A}_c^m;\{\mathbf{I}_c^m\}),\ m=1,\cdots,M.
\end{equation*}

The ADS consists of duplicated blocks composed of two Dynamic Convolutions (DC)~\cite{chen2020dynamic} and one Global Average Pooling (GAP) module. The schematic of DC is depicted in Fig.\ref{fig:framework} where DC aggregates multiple parallel convolution kernels dynamically based upon their attentions, boosting the model complexity without increasing the network depth or width.

\begin{algorithm}[t]
	\renewcommand{\algorithmicrequire}{\textbf{Input:}}
	\renewcommand{\algorithmicensure}{\textbf{Output:}}
	\caption{GUPDM Updating Framework}\label{alg:GUPDM}
	\begin{algorithmic}[1]
		
		\REQUIRE Paired underwater data, necessary initialization parameters.\\
		\STATE $t = 1$;
		\WHILE{Not Converge}
		\STATE Use a batch of images $\mathbf{I}_c(x)$ as input :
		\STATE $\widetilde{\mathbf{J}}_c(x)\leftarrow \mathcal{N}(\bm{\omega};\mathcal{A}(\bm{\theta},\mathbf{A}_{c}),\mathcal{T}(\bm{\phi},\mathbf{T}_{c});\mathbf{I}_c(x))$		
		\STATE Evaluate the total loss in 
 $ L_{\text{total}}=L( \widehat{\mathbf{J}},\mathbf{I}_{\text{gt}};\mathbf{I}_c(x) ) $
		\STATE $\#$ Update the parameters $\bm{\omega}$ of PMS
		\STATE $\bm{\omega} \leftarrow \bm{\omega}  -\rho_0 \nabla_{\bm{\omega} } L_{\text{total}}$

		\IF{$(t \ Mod \ t_0)=0$}
		\STATE  Generating $M$ patches degraded underwater images by $\mathcal{G}_\mathbf{A}$ as input $\mathbf{I}_{c}^{m}(x)$:
		\STATE $\widehat{\mathbf{J}}_c(x)\leftarrow \mathcal{N}(\bm{\omega};\mathcal{A}(\bm{\theta},\mathbf{A}_{c}^{m}),\mathcal{T}(\bm{\phi},\mathbf{T}_{c}(x);\mathbf{I}_{c}(x))$
		\STATE Evaluate the total loss in 
		\STATE 
		 $L_{\text{total}}^A=\sum_{m=1}^{M}L(\widehat{\mathbf{J}},\mathbf{I}_{\text{gt}};\mathbf{I}_c^m(x),\mathbf{I}_c(x))$
		\STATE $\#$ Update the parameters $\bm{\theta}$ of ADS
		\STATE $\bm{\theta} \leftarrow \bm{\theta}  -\rho_1 \nabla_{\bm{\theta} } L_{\text{total}}^A$
		\ENDIF
		
		\IF{$(t \ Mod \ t_1)=0$}
		\STATE  Generating $N$ patches degraded underwater images by $\mathcal{G}_\mathbf{T}$ as input $\mathbf{I}_{c}^{n}(x)$:
		\STATE $\bar{\mathbf{J}}_c(x)\leftarrow \mathcal{N}(\bm{\omega};\mathcal{A}(\bm{\theta},\mathbf{A}_{c}^{m}),\mathcal{T}(\bm{\phi},\mathbf{T}_{c}^{n}(x);\mathbf{I}_{c}(x))$
		\STATE Evaluate the total loss in 
		\STATE
		 $L_{\text{total}}^T=\sum_{n=1}^{N}L(\bar{\mathbf{J}},\mathbf{I}_{\text{gt}};\mathbf{I}_c^n(x),\mathbf{I}_c(x))$
		\STATE $\#$ Update the parameters $\bm{\phi}$ of TDS
		\STATE $\bm{\phi} \leftarrow \bm{\phi}-\rho_2 \nabla_{\bm{\phi}} L_{\text{total}}^T$
		\ENDIF
		\STATE $t = t+1$
		
		\ENDWHILE 
	\end{algorithmic}
\end{algorithm}

\textbf{Transmission-guided Dynamic Structure.} 
We first estimate the  transmission map $\mathbf{T}_c(x)$ via general dark channel prior (UDCP~\cite{drews2013UDCP}). 
Since the transmission $\mathbf{T}_c(x)$ can reflect water types, we vary it through multiplying $\mathbf{T}_c(x)$ with randomly generated coefficients:
\begin{equation*}
	\mathbf{T}_c^n(x)=\mathcal{G}_\mathbf{T}(\mathbf{T}_c(x))=\gamma_c^n\cdot\mathbf{T}_c(x) , n=1,\cdots,N,
\end{equation*}
where $\gamma_c^n\in(0.5,1.1)$ is the random coefficient. 
We then introduce a TDS (i.e., $\mathcal{T}(\bm{\phi})$) to enable our model to adaptively select appropriate parameters according to water types while paying more attention to various image details. We formulate this step as:
\begin{equation*}
	\mathbf{f}_{\text{TDS}}= \mathcal{T}(\bm{\phi},\mathbf{T}_c^n;\{\mathbf{I}_c^{n}\}),\ n=1,\cdots,N. 
\end{equation*}
Specifically, $\mathbf{f}_{\text{TDS}}$ and the transmission map $\mathbf{f}_{\text{m}}$ are integrated in Transmission Guided Feature Extraction module $\mathcal{M}$ (i.e., TGFE). The TDS has the same base structure with ADS.

\begin{table*}[t]
	\centering
	\caption{Quantitative comparisons on datasets with reference. We employ experiment on four datasets and three metrics, and achieve the best performance under most settings. We highlight the best result in black bold and underline the second ones.}
	\label{tab:comparison_supervised}
	\renewcommand\arraystretch{1}
	\resizebox{\linewidth}{!}{
		\begin{tabular}{c|ccccccccccc|c}
			\hline
			\toprule
			Datasets & Metrics & UDCP \cite{drews2013UDCP}  & Fusion \cite{2012fusion} & Water-Net \cite{li2019WaterNet-UIEB-C60}& UGAN \cite{Fabbri2018UGAN}  & FUnIEGAN \cite{Islam2020FUnIE-GAN}   & Ucolor \cite{li2021Ucolor} & USUIR \cite{2022USUIR}& PUIE\_net \cite{fu2022puieNet} &Uranker-NU$^2$Net \cite{guo2022URanker} &Semi-UIR \cite{huang2023Semi-UIR} & Ours    \\
			\hline
			\midrule
			
			& PSNR$\uparrow$  & 16.38 & 17.61 & 20.14         & 21.89      & 18.43 & 22.23      & 18.2  & 17.7  & \underline{23.92} & 18.7           & \textbf{25.48} \\
			Test-E515     & SSIM$\uparrow$   & 0.64  & 0.75  & 0.68          & 0.8        & 0.76  & 0.83 & 0.74  & 0.76  & \underline{0.84}  & 0.73           & \textbf{0.86}  \\
			
			& MSE$\downarrow$   &1990  & 1331  & 826           & 556        & 1115  & 492        & 1153  & 1424  & \underline{368}   & 1015           & \textbf{254}   \\
			
			\hline
			& PSNR$\uparrow$   & 13.05 & 17.6  & 19.11         & 20.51      & 16.81 & 15.52      & 20.64 & 19.21 & 20.47       & \textbf{22.42} & \underline{21.53}    \\
			
			Test-U90     & SSIM$\uparrow$   & 0.62  & 0.77  & 0.79          & 0.79       & 0.74  & 0.67       & 0.85  & 0.87  & 0.85        & \textbf{0.89}  & \underline{0.86}     \\
			
			& MSE$\downarrow$    & 3779  & 1331  & 1220          & 911        & 1778  & 2217       & 716   & 966   & 879         & \textbf{522}   & \underline{721}      \\
			
			\hline
			& PSNR$\uparrow$   & 12.66 & 14.48 & 17.73         & 24.43      & 19.43 & 17.92      & 19.35 & 19.61 & \underline{24.66} & 22.36          & \textbf{25.36} \\
			
			Test-L504    & SSIM$\uparrow$   & 0.62  & 0.79  & 0.82          & 0.86       & 0.81  & 0.75       & 0.85  & 0.86  & \underline{0.92}  & 0.86           & \textbf{0.93}  \\
			
			& MSE$\downarrow$    & 4529  & 3501  & 1361          & \underline{285}  & 891   & 1596       & 933   & 840   & 296         & 522            & \textbf{261}   \\
			
			\hline
			& PSNR$\uparrow$   &   18.26 & 14.58 & 22.46         & 21.93      & 18.35 & 21.79      & 17.87 & 17.09 & \underline{23.09} & 18.68          & \textbf{24.34} \\

			Test-U120  & SSIM$\uparrow$   &   0.72  & 0.54  & \textbf{0.79} & \underline{0.77} & 0.73  & 0.76       & 0.72  & 0.72  & \underline{0.77}  & 0.72     & \textbf{0.79}  \\
			
			& MSE$\downarrow$    & 1249      &   2968     &     458      &   525    &  1172     &495 &   1222     & 1633  & \underline{370}   & 1005           & \textbf{297}     \\ 
			\bottomrule
		\end{tabular}
	}
\end{table*}

\begin{table*}[tb]
	\centering
	\caption{Averaged unsupervised scores (i.e., PS, UIQM, UCIQE, NIQE and URanker) on three real-world underwater datasets without reference images (i.e., Test-C60, Test-R300 and Test-S16).}
	\resizebox{\linewidth}{!}{
		\begin{tabular}{c|ccccc|ccccc|ccccc}
			\Xhline{1.2pt}
			\multirow{2}{*}{Datasets} & \multicolumn{5}{c|}{Test-C60} & \multicolumn{5}{c|}{Test-R300}& \multicolumn{5}{c}{Test-S16}\\
			&      PS$\uparrow$ & UIQM$\downarrow$ & UCIQE$\uparrow$ & NIQE$\downarrow$ & Uranker $\uparrow$ &      PS$\uparrow$ & UIQM$\downarrow$ & UCIQE$\uparrow$ & NIQE$\downarrow$ & Uranker$\uparrow$&      PS$\uparrow$ & UIQM$\downarrow$ & UCIQE$\uparrow$ & NIQE$\downarrow$ & Uranker$\uparrow$\\
			\hline
			\midrule
			Water-Net \cite{li2019WaterNet-UIEB-C60} & 6.45          & 2.86          & 27.39          & \underline{4.97}    & 0.93          & 4.45          & 2.58          & 27.49          & 4.91          & 2.17          & 6.18          & 2.11          & 24.66          & 6.41          & 1.99          \\
			FUnIEGAN \cite{Islam2020FUnIE-GAN}& 4.58          & 4.54          & 30.05          & 6.21          & 1.67          & 4.22          & 4.00          & 28.58          & 4.89          & 2.25          & 6.51          & 3.27          & 27.81          & 6.49          & 2.02          \\
			USUIR \cite{2022USUIR}   & 3.94          & 4.71          & 32.04          & 5.74          & 1.14          & 3.66          & 3.34          & 20.34 & 4.68    & 1.83          & 5.31          & 3.30          & 28.24          & \underline{6.37}          & 0.42          \\	
			Fusion \cite{2012fusion}   & 4.73          & 2.67          & 32.26          & 5.29          & 1.24          & 3.18          & 3.06          & \textbf{31.58} & \underline{4.36}    & 2.34          & 7.86          & 2.01          & 31.35          & 12.19         & 0.91          \\
			Ucolor \cite{li2021Ucolor}  & 6.26          & 4.31          & 24.27          & 5.34          & 0.63          & 4.40          & 3.36          & 20.48          & 4.83          & 0.77          & \textbf{8.90} & 3.11          & 27.51          & 11.85         & 0.58          \\
			UDCP \cite{drews2013UDCP}     & \textbf{6.66} & 1.54          & \textbf{32.55} & 5.44          & 0.62          & \underline{4.88}    & 2.30          & 28.72          & 5.41          & 0.46          & 7.47          & 1.02          & \textbf{33.92} & 9.05          & 0.26          \\
			UGAN \cite{Fabbri2018UGAN}    & 4.53          & 4.69          & 31.08          & 6.81          & \underline{1.91}    & 4.03          & \underline{4.29}    & 29.44          & 5.22          & \underline{2.61}    & 6.01          & 3.29          & 28.26          & 6.98          & 2.02          \\
			PUIE\_NET \cite{fu2022puieNet}& 4.54          & 3.64          & 27.25          & 6.22          & 1.05          & 4.27          & 4.00          & 26.83          & 4.90          & 1.77          & 6.99          & \underline{3.34}    & 27.81          & 7.97          & 0.99          \\
			Uranker-NU$^2$Net \cite{guo2022URanker}  & 4.26          & 4.57          & 28.90          & 5.79          & 1.18          & 4.65          & 4.18          & 28.63          & 4.69          & 2.30          & 5.84          & 3.26          & 30.17          & 6.66          & \underline{2.11}    \\
			Semi-UIR \cite{huang2023Semi-UIR} & 3.97          & \underline{4.74}    & 30.63          & 5.77          & 1.69          & 4.52          & 4.11          & 29.23          & 4.52          & 2.45          & 5.79          & 2.63          & 30.00          & 6.43          & 1.98          \\

			Ours      & \underline{6.45}    & \textbf{4.88} & \underline{32.36}    & \textbf{4.76} & \textbf{1.94} & \textbf{4.95} & \textbf{4.35} & \underline{31.05}    & \textbf{4.26} & \textbf{2.64} & \underline{8.06}    & \textbf{3.52} & \underline{31.78}    & \textbf{6.18} & \textbf{2.22}
			\\
			\Xhline{1pt}
		\end{tabular}
	}
	\label{tab:comparison_unsupervised}
\end{table*}

\textbf{Prior-based Multi-scale Structure.} 
This structure is a basic part of our underwater image enhancement. First, to mine the physical knowledge and features of depth-texture information at different scales, we adopt the prior-based multi-scale structure to estimate the preliminary pixel feature $\mathbf{f}_{\text{PMS}}$ in this branch. 
\begin{equation*}
	\mathbf{J}_c(x)\leftarrow \mathcal{N}(\bm{\omega};\mathcal{A}(\bm{\theta},\mathbf{A}_c),\mathcal{T}(\bm{\phi},\mathbf{T}_c);\mathbf{I}_c(x)). 
\end{equation*}

As for $\mathcal{N}$, it mainly consists of a series of HMH and 3H modules with a tailored Transmission-Guided Feature Extraction (TGFE) module at its bottleneck. 
HMH comprises a Hyper Residual Block (HRB), a Multi-scale Feature Extraction module (MFE) and another HRB, while 3H consists of three HRBs. 
The HRB converts input parameters (i.e. $\mathbf{f}_{\text{ADS}}$ or $\mathbf{f}_{\text{TDS}}$) to weights of convolution block via a Fully Connected (FC) layers. 
The detailed structure of the MFE module is shown in top right of Fig.~\ref{fig:framework}. 
To improve the receptive field of the framework, we utilize three convolution blocks with different kernel sizes $1 \times 1$, $3 \times 3$ and $5 \times 5$,  generating three feature maps. We then employ global max pooling and average pooling in channel attention block to automatically extract and fuse the main features from three feature maps at different scales. 
TGFE consists of a HRB, a TC, a MFE and another HRB sequentially, integrating the transmission feature $\mathbf{f}_{m}$ to tackle the  blur and haze.

\subsection{Loss Functions and Training Procedure}\label{subsec:loss}
We carefully design the training loss functions to guide the model to produce enhanced results with  minimum color artifacts, blurriness and the closest details to reference images. Firstly, we use  pretrained VGG16 network to extract the feature in $3^{th}, 8^{th}$ and $15^{th}$ layers to formulate the perceptional loss $L_{per}$:
\begin{equation}
	L_{\text{per}}(\mathbf{J},\mathbf{I}_{\text{gt}})=L_{\text{MSE }}\left(\operatorname{VGG}_{3,8,15}(\mathbf{J}), \mathrm{VGG}_{3,8,15}\left(\mathbf{I}_{\text{gt}}\right)\right),
\end{equation}
where $L_{\text{MSE }}$ denotes the Mean Square Error (MSE). 
We also impose the smooth $L_{1}-\operatorname{loss}$ as reconstruction loss since we notice that unnecessary interference often occurs from the background color:
\begin{equation}
	L_{1}(\mathbf{J},\mathbf{I}_{\text{gt}})=\frac{1}{n} \sum_{i=1}^{n}|\mathbf{J}(i)-\mathbf{I}_{\text{gt}}(i)|,
\end{equation}
where $n = H \times W$ is the overall pixel number. We furthermore apply the SSIM loss to focus more on the structural details. Thus, the total loss function can be summarized as follows:
\begin{equation}\label{loss}
	L(\mathbf{J},\mathbf{I}_{\text{gt}})=\ell_{1}+\lambda_{1} \mathcal{L}_{\text{SSIM }}+\lambda_{2} \mathcal{L}_{\text{per }}
\end{equation}
where $\lambda_{1}$ and $\lambda_{2}$ stand for the weights of each loss functions.

We illustrate the training procedure of proposed method in Algorithm ~\ref{alg:GUPDM}. In particular, we first train the base-net PMS with given atmosphere $\mathbf{A}_c$ and transmission $\mathbf{T}_c$ and loss function $L(\widetilde{\mathbf{J}},\mathbf{I}_{\text{gt}};\mathbf{I}_c(x))$. Next, we fix the base-net PMS  and update parameter of ADS (i.e., $\bm{\theta}$) which is optimized with loss $L(\widehat{\mathbf{J}},\mathbf{I}_{\text{gt}};\mathbf{I}_c^m(x),\mathbf{I}_c(x))$. Similarly, the parameter of TDS (i.e., $\bm{\phi}$ ) is updated through optimizing $L(\bar{\mathbf{J}},\mathbf{I}_{\text{gt}};\mathbf{I}_c^n(x),\mathbf{I}_c(x))$. 
The ADS, TDS and PMS are trained in an alternate manner until they converge.

\section{Experimental Results}
This section first introduces the implementation details in Subsection~\ref{subsec:implentation}. Then, to evaluate the performance when comparing with existing state-of-the-art approaches, a series of qualitative and quantitative assessments are conducted in Subsection~\ref{subsec:comp_SOTA}. To analyze the developed method, we conduct various ablation studies to verify the effectiveness of different branches in Subsection~\ref{subsec:ablation}. 
\emph{More experimental results are provided in our supplementary material}. 

\subsection{Implantation Details}\label{subsec:implentation}

\textbf{Datasets.} 
Our model is trained on LSUI~\cite{2021LSUI}, a real underwater dataset, which consisting 4500 training pairs and 504 test pairs (i.e. Test-504). In order to evaluate the effectiveness and robustness of our proposed method, we test it on six real world datasets: Test-U90 (i.e. UIEB \cite{li2019WaterNet-UIEB-C60}), Test-L504 (i.e. LSUI \cite{2021LSUI}), Test-U120 (i.e. UFO \cite{2020UFO}), Test-S16 (i.e. SQUID \cite{2020SQUID}), Test-R300 (i.e. RUIE \cite{2020RUIE}) and Test-C60 \cite{li2019WaterNet-UIEB-C60}, and a synthetic one: Test-E515 (i.e. EUVP \cite{2020EUVP}). 
There are two categories of testing datasets: those that have reference images (either real or generated by another method) and those that don’t. The first type includes Test-E515, Test-U90, Test-L504 and Test-U120 datasets, where Test-E515 uses synthetic images created by a CycleGAN-based model. The second type includes Test-C60, Test-R300 and Test-S16 datasets. Specifically, Test-C60 contains 60 difficult images from UIEB and Test-S16 has 16 images from its original dataset without references.

\begin{figure*}[t]
	\centering 
	\begin{tabular}{c@{\extracolsep{0.2em}}c@{\extracolsep{0.2em}}c@{\extracolsep{0.2em}}c@{\extracolsep{0.2em}}c@{\extracolsep{0.2em}}c@{\extracolsep{0.2em}}c@{\extracolsep{0.2em}}c@{\extracolsep{0.2em}}c}
		
		\rotatebox{90}{\quad\quad\ EUVP}
		&\includegraphics[width=0.118\textwidth]{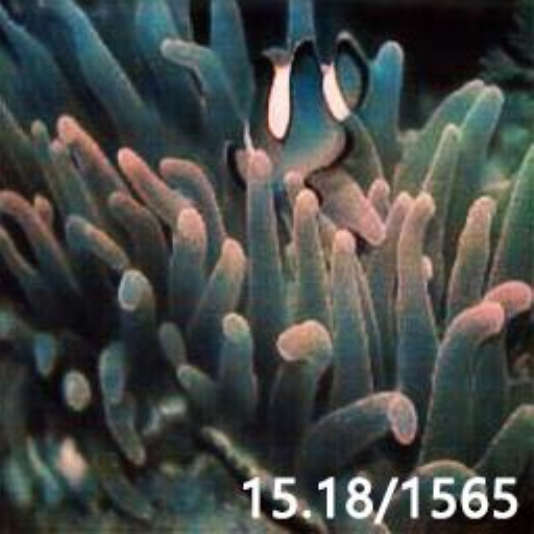}
		&\includegraphics[width=0.118\textwidth]{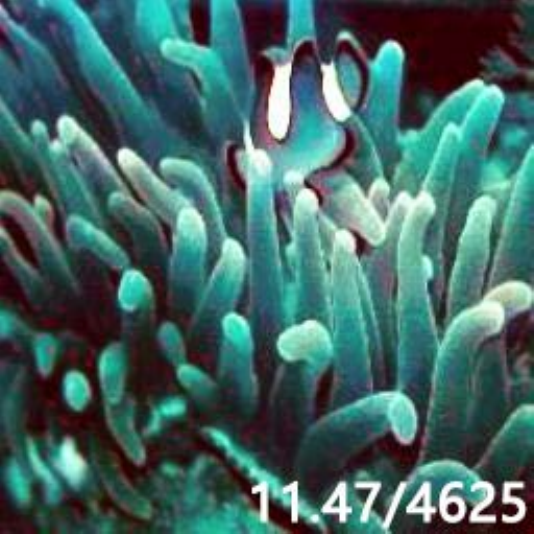}
		&\includegraphics[width=0.118\textwidth]{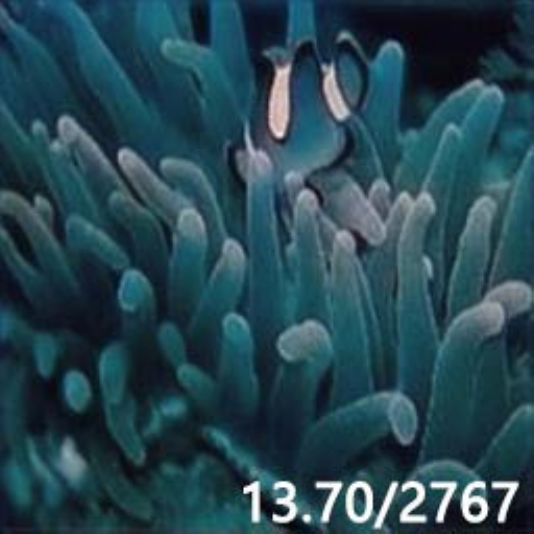}
		&\includegraphics[width=0.118\textwidth]{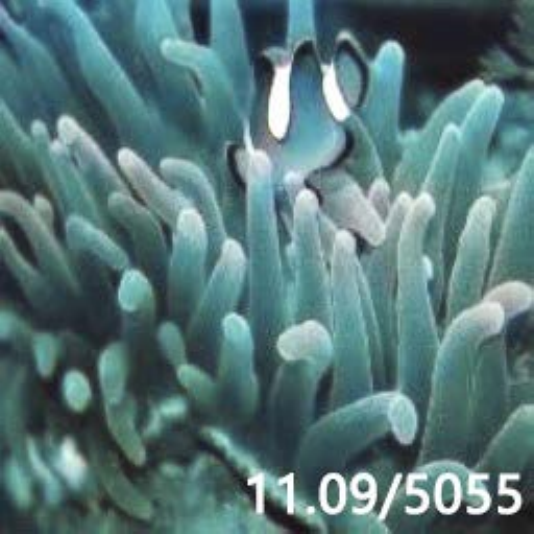}
		&\includegraphics[width=0.118\textwidth]{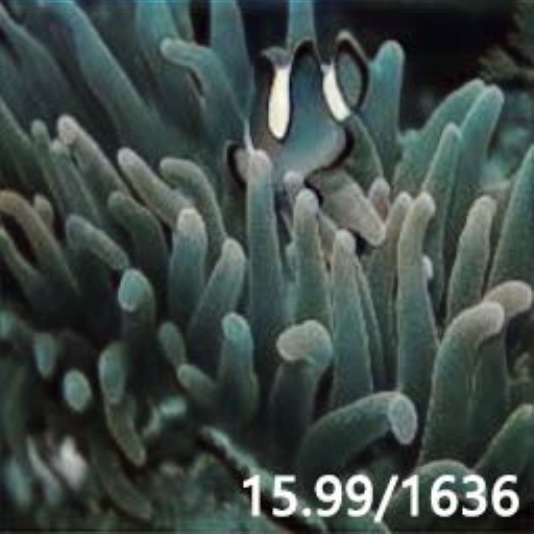}
		&\includegraphics[width=0.118\textwidth]{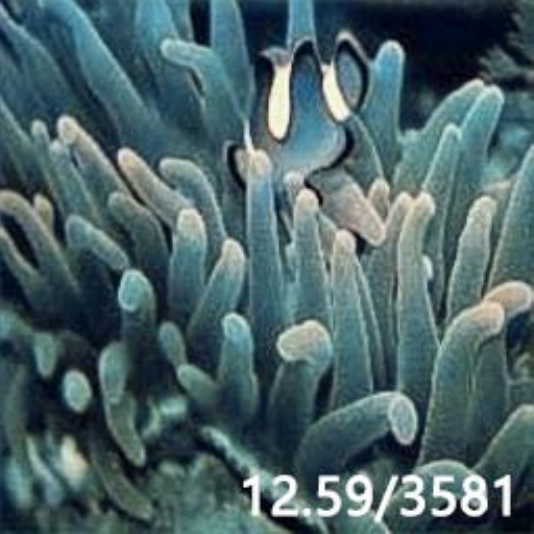}
		&\includegraphics[width=0.118\textwidth]{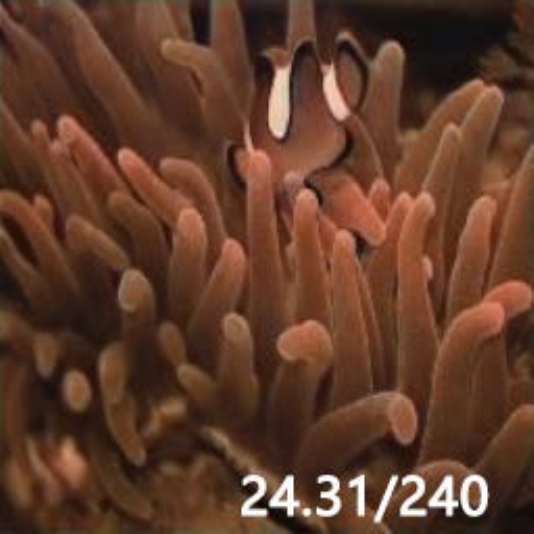}
		&\includegraphics[width=0.118\textwidth]{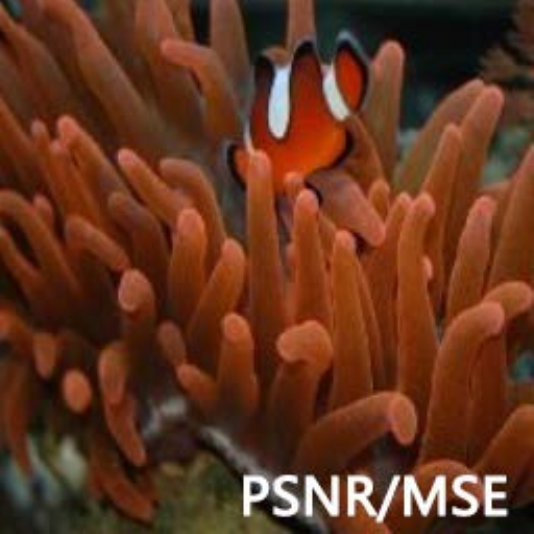}\\

		\rotatebox{90}{\quad\quad\ UIEB}
		&\includegraphics[width=0.118\textwidth]{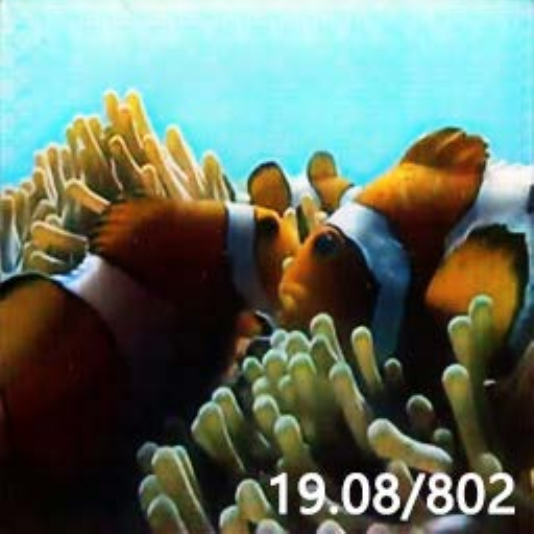}
		&\includegraphics[width=0.118\textwidth]{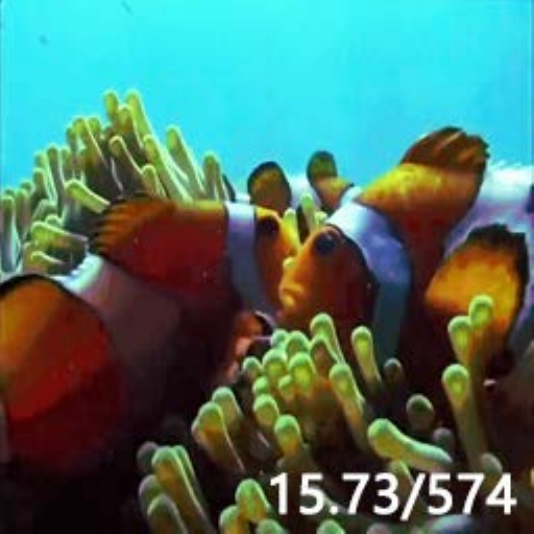}
		&\includegraphics[width=0.118\textwidth]{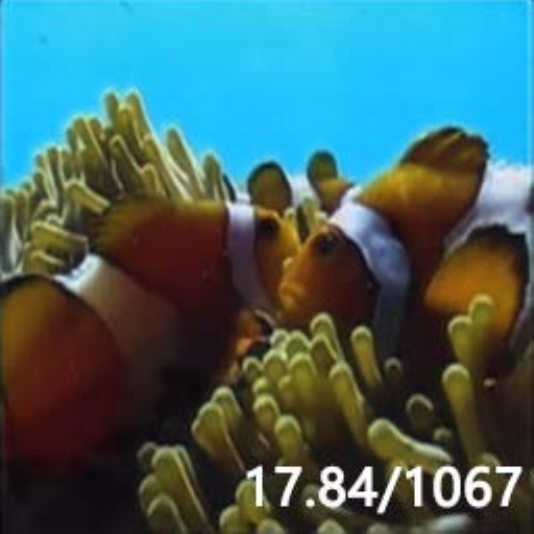}
		&\includegraphics[width=0.118\textwidth]{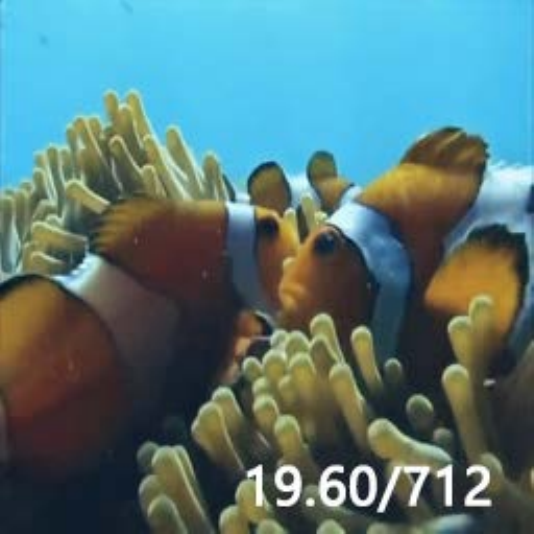}
		&\includegraphics[width=0.118\textwidth]{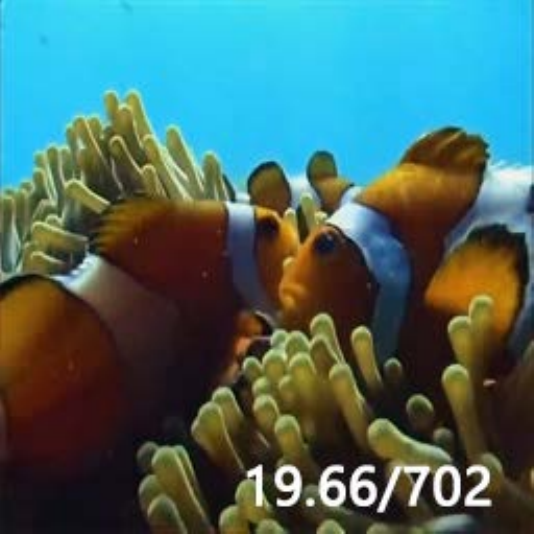}
		&\includegraphics[width=0.118\textwidth]{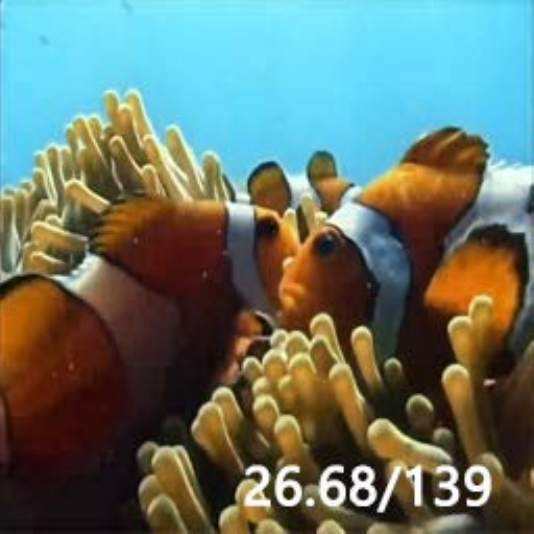}
		&\includegraphics[width=0.118\textwidth]{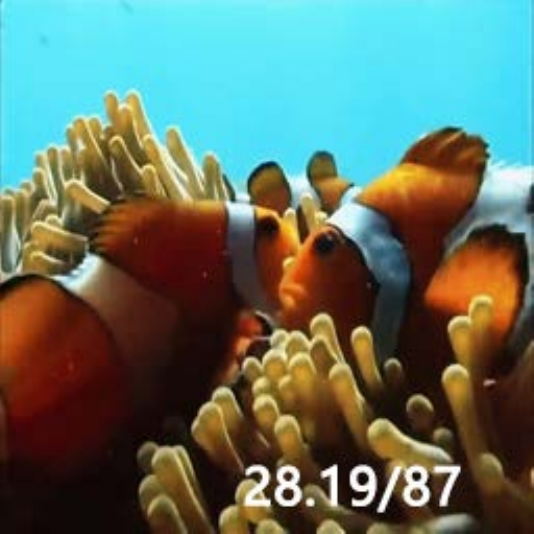}
		&\includegraphics[width=0.118\textwidth]{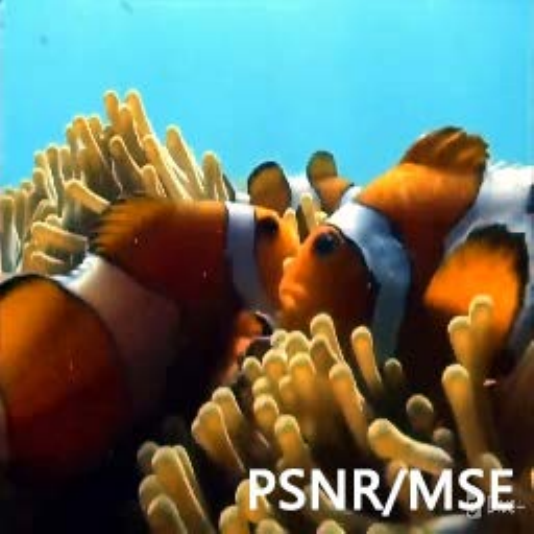}\\
		
		\rotatebox{90}{\quad\quad\ LSUI}
		&\includegraphics[width=0.118\textwidth]{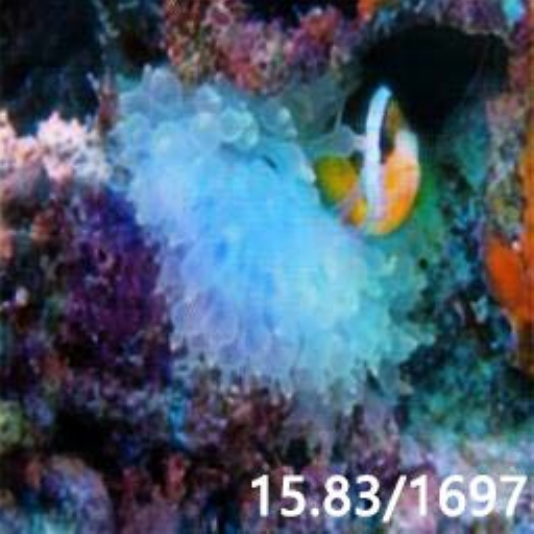}
		&\includegraphics[width=0.118\textwidth]{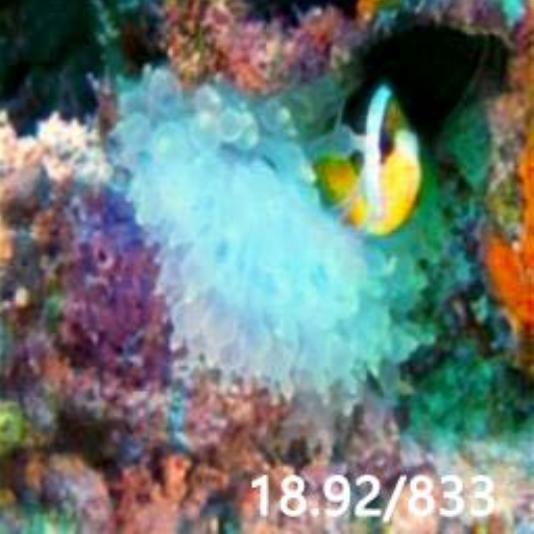}
		&\includegraphics[width=0.118\textwidth]{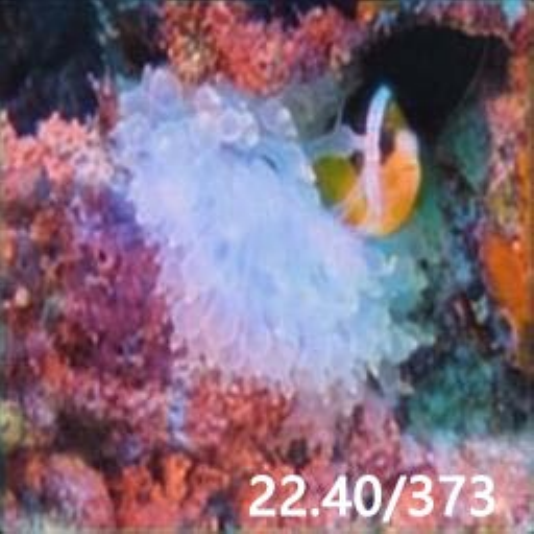}
		&\includegraphics[width=0.118\textwidth]{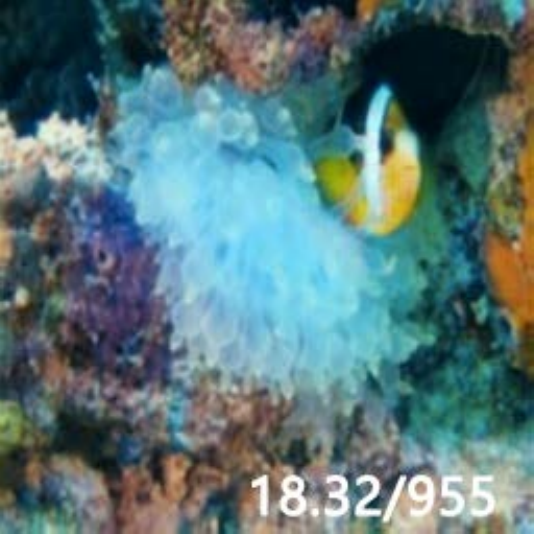}
		&\includegraphics[width=0.118\textwidth]{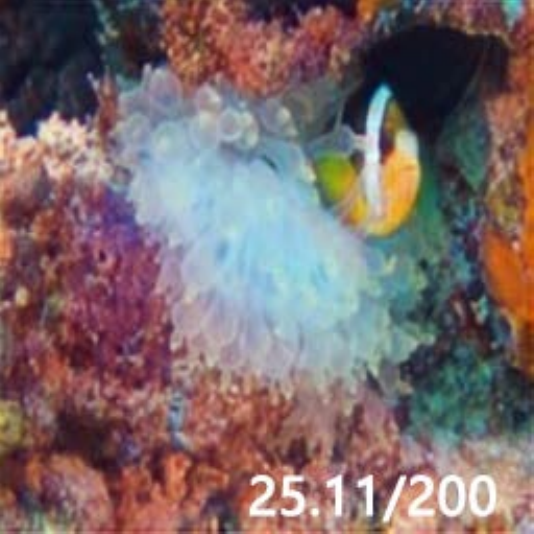}
		&\includegraphics[width=0.118\textwidth]{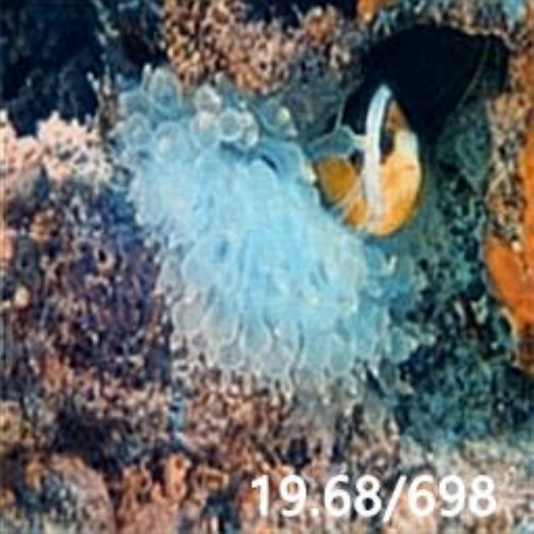}
		&\includegraphics[width=0.118\textwidth]{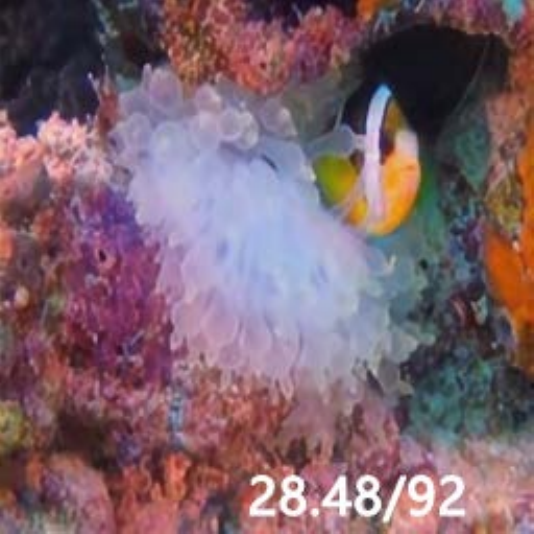}
		&\includegraphics[width=0.118\textwidth]{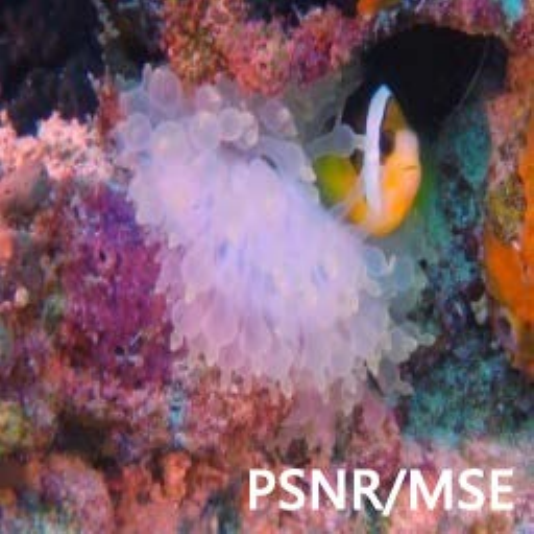}\\

		\rotatebox{90}{\quad\quad\ UFO}
		&\includegraphics[width=0.118\textwidth]{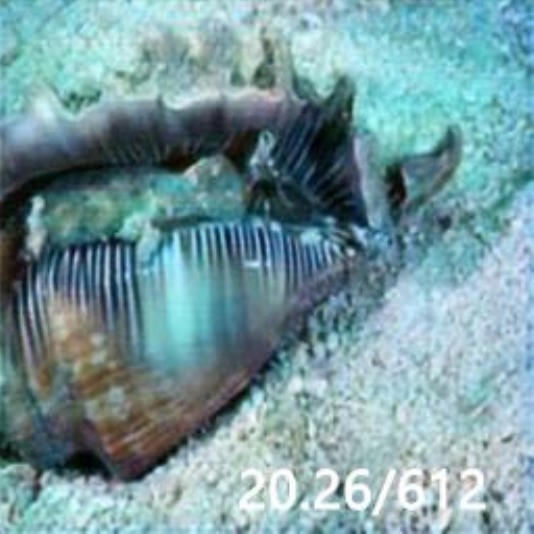}
		&\includegraphics[width=0.118\textwidth]{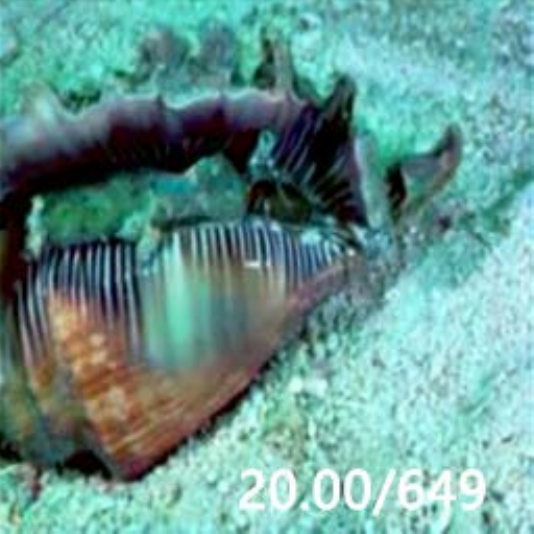}
		&\includegraphics[width=0.118\textwidth]{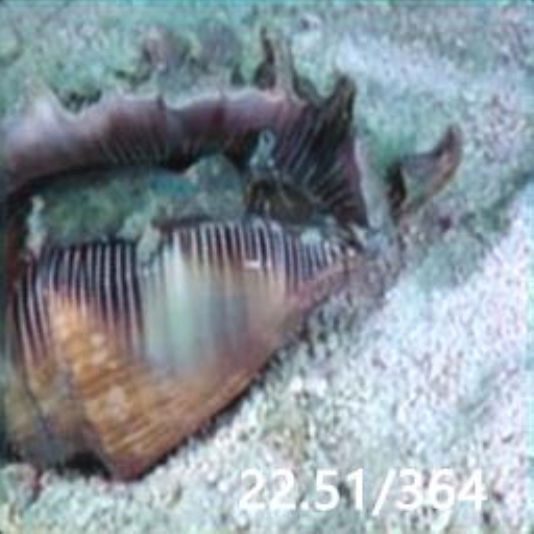}
		&\includegraphics[width=0.118\textwidth]{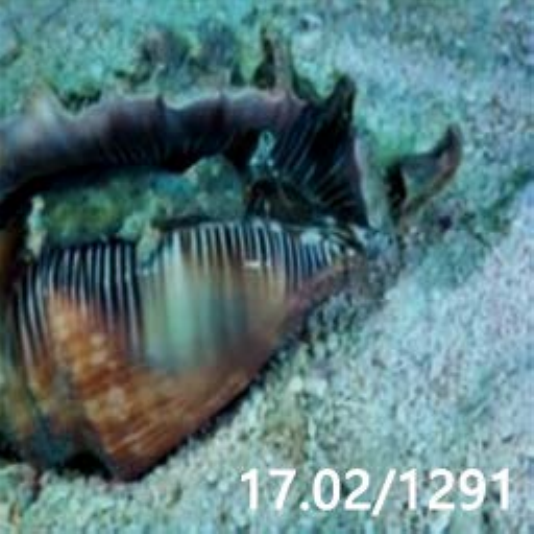}
		&\includegraphics[width=0.118\textwidth]{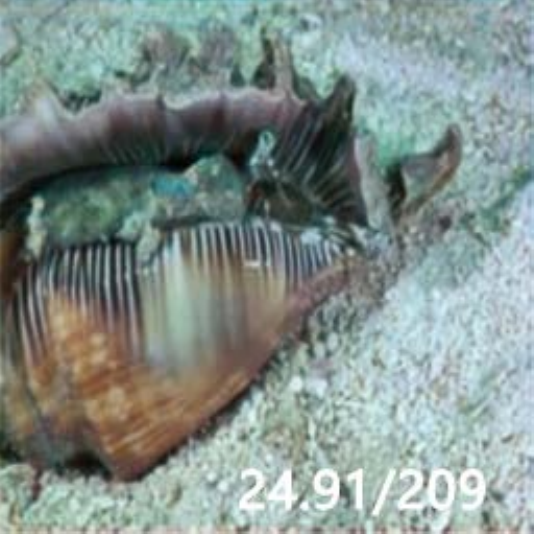}
		&\includegraphics[width=0.118\textwidth]{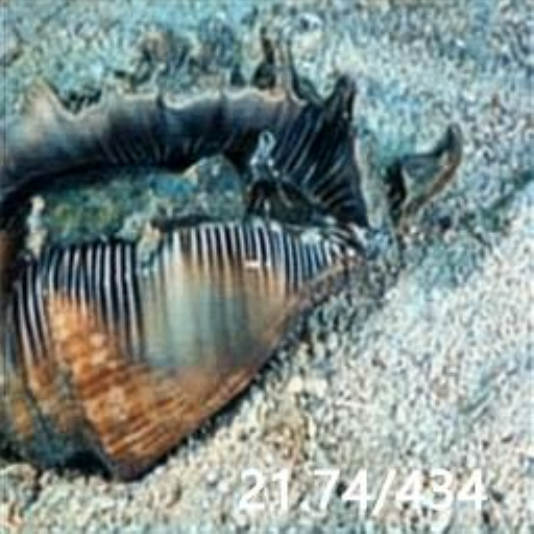}
		&\includegraphics[width=0.118\textwidth]{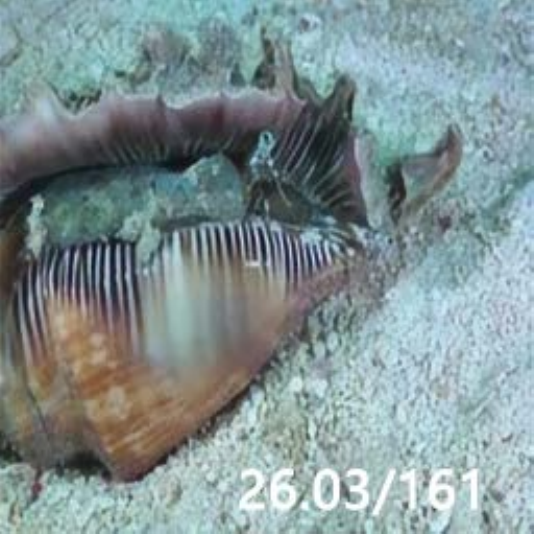}
		&\includegraphics[width=0.118\textwidth]{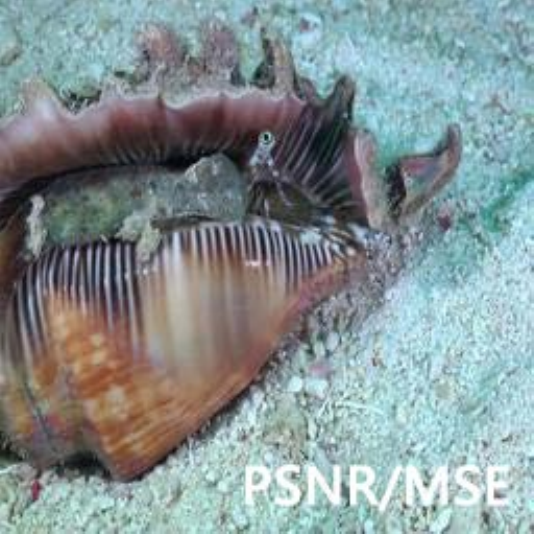}\\
		
		& \footnotesize FUnIEGAN & \footnotesize USUIR & \footnotesize  Ucolor& \footnotesize PUIENet  &  \footnotesize Uranker-NU$^2$Net  & \footnotesize Semi-UIR & \footnotesize Ours &\footnotesize Reference
	\end{tabular}
	\caption{Visual comparison with different methods on synthesis dataset (EUVP) and real-world datasets (UIEB, LSUI and UFO). Comparing with FUnIEGAN, USUIR, Ucolor, PUIENet, URanker, Semi-UIR, our developed method (i.e., SIM-Net) achieves an natural color and  preserves more details, performing the best visual quality.
	}\label{fig:syn_real} 
\end{figure*}

\begin{figure*}[t]
	\centering 
	\begin{tabular}{c@{\extracolsep{0.2em}}c@{\extracolsep{0.2em}}c@{\extracolsep{0.2em}}c@{\extracolsep{0.2em}}c@{\extracolsep{0.2em}}c@{\extracolsep{0.2em}}c@{\extracolsep{0.2em}}c@{\extracolsep{0.2em}}c}
		
		\rotatebox{90}{\quad\ Test-C60}
		&\includegraphics[width=0.118\textwidth]{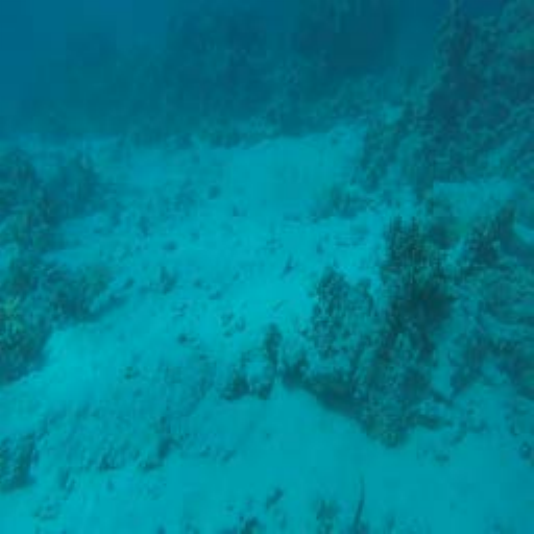}
		&\includegraphics[width=0.118\textwidth]{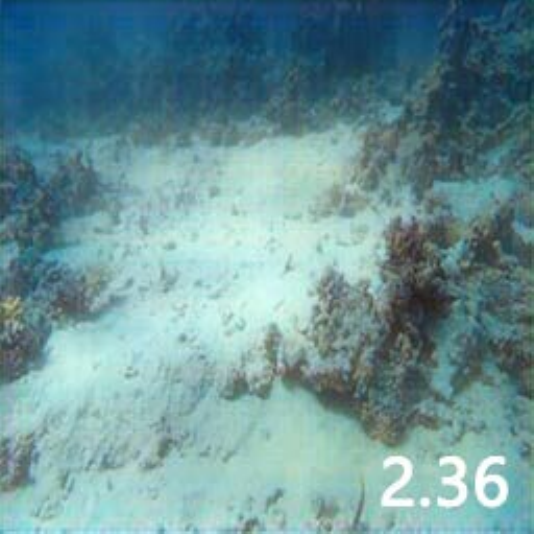}
		&\includegraphics[width=0.118\textwidth]{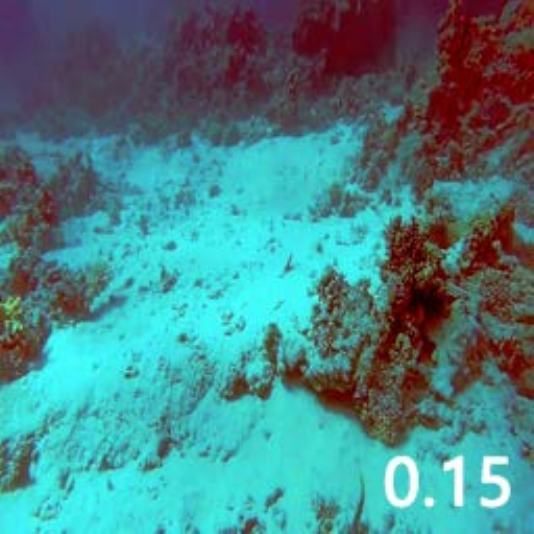}
		&\includegraphics[width=0.118\textwidth]{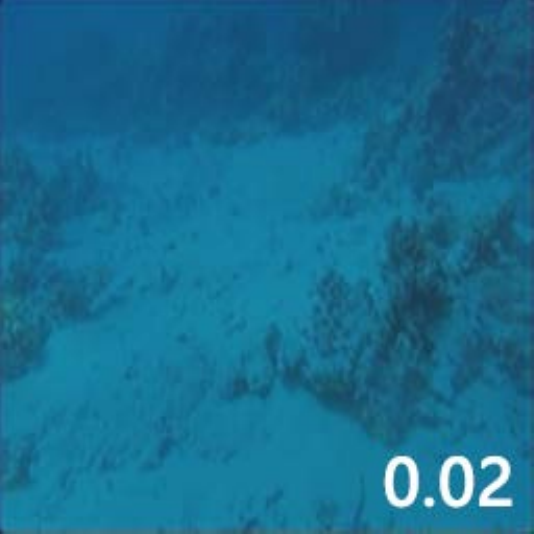}
		&\includegraphics[width=0.118\textwidth]{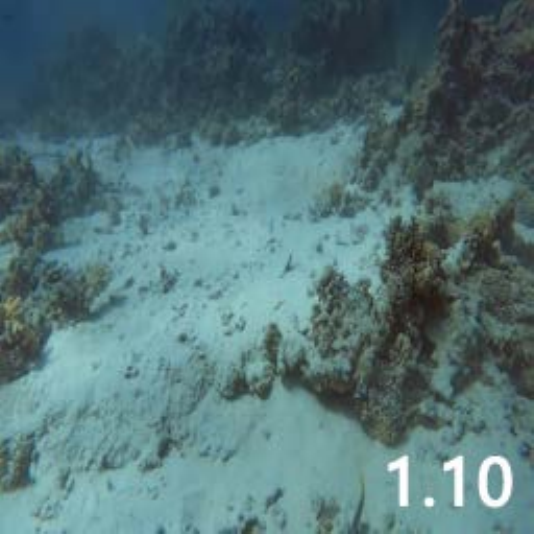}
		&\includegraphics[width=0.118\textwidth]{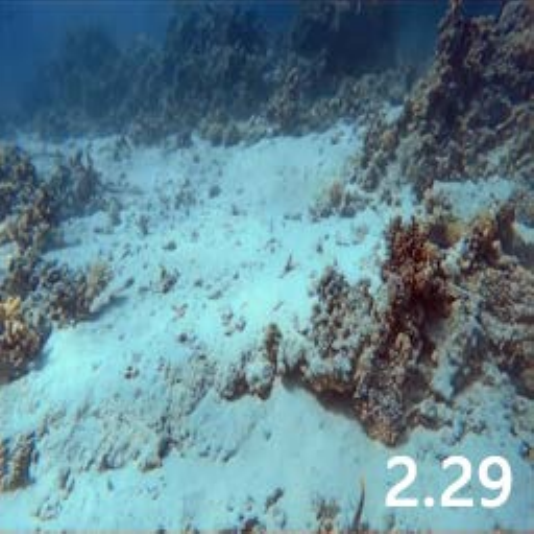}
		&\includegraphics[width=0.118\textwidth]{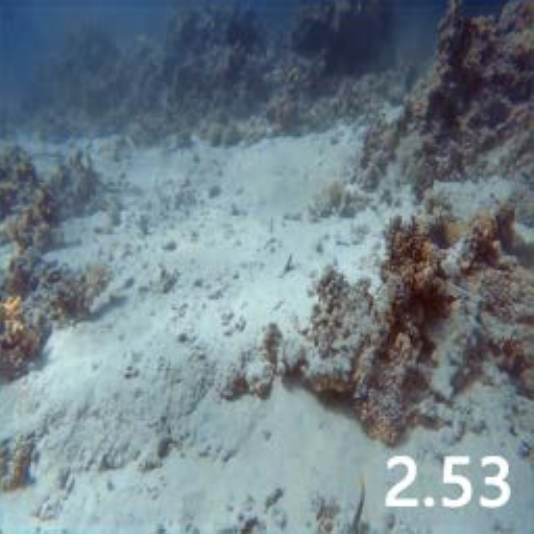}
		&\includegraphics[width=0.118\textwidth]{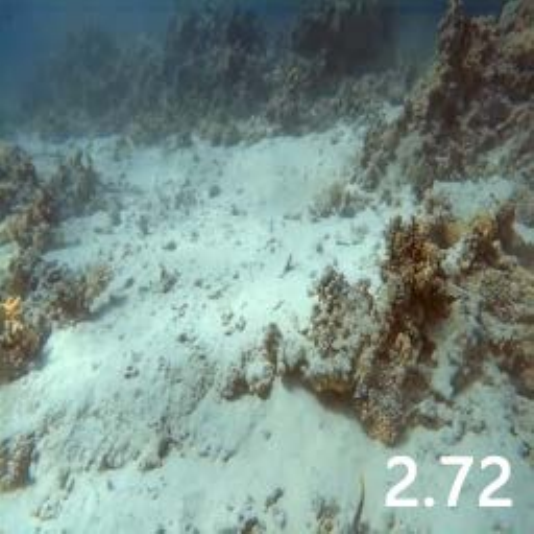}\\

		\rotatebox{90}{\quad Test-R300}
		&\includegraphics[width=0.118\textwidth]{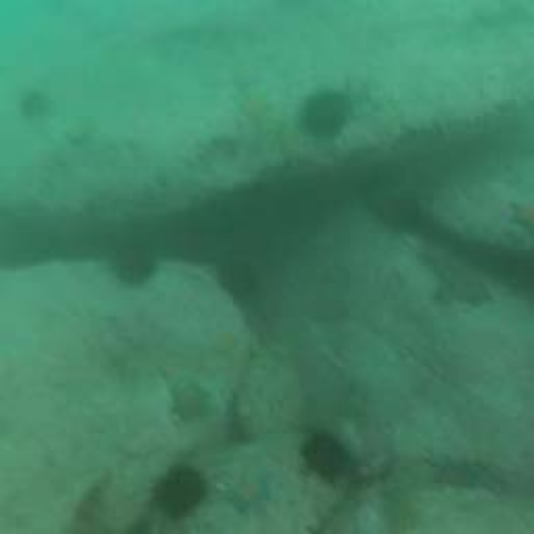}
		&\includegraphics[width=0.118\textwidth]{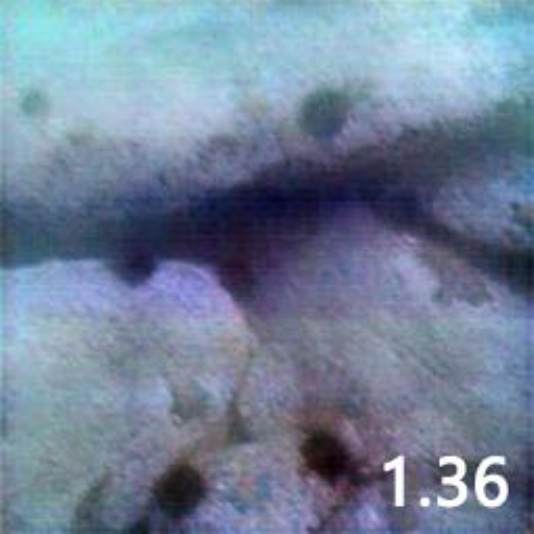}
		&\includegraphics[width=0.118\textwidth]{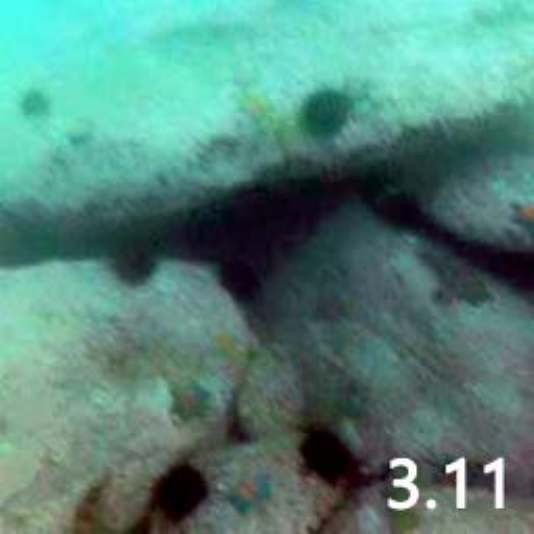}
		&\includegraphics[width=0.118\textwidth]{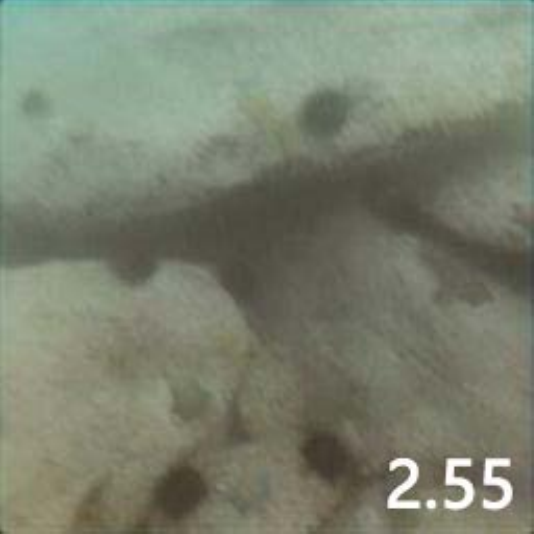}
		&\includegraphics[width=0.118\textwidth]{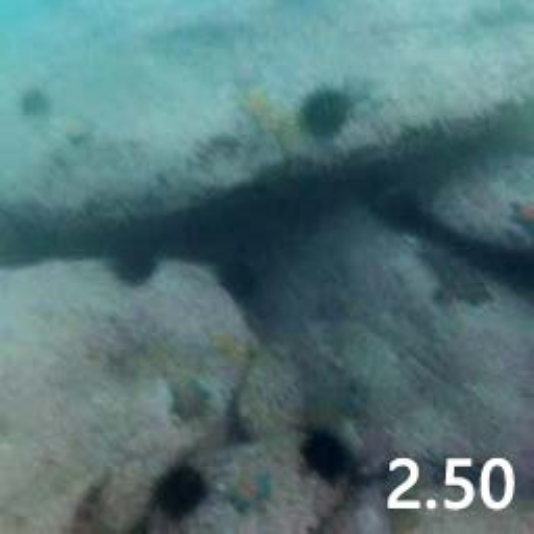}
		&\includegraphics[width=0.118\textwidth]{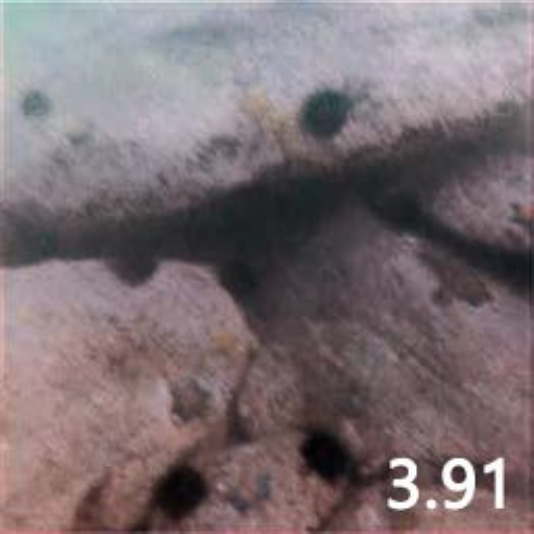}
		&\includegraphics[width=0.118\textwidth]{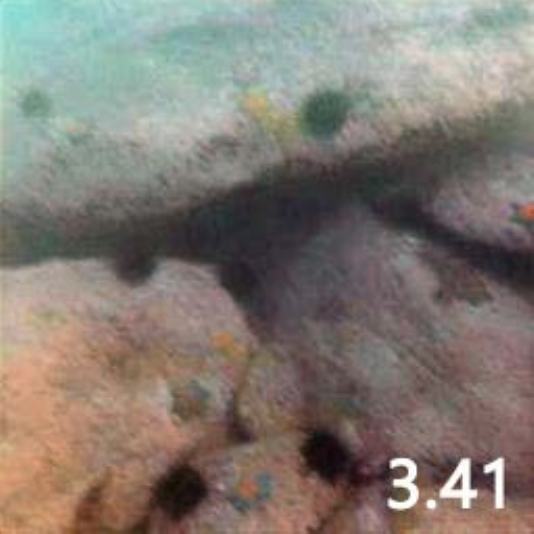}
		&\includegraphics[width=0.118\textwidth]{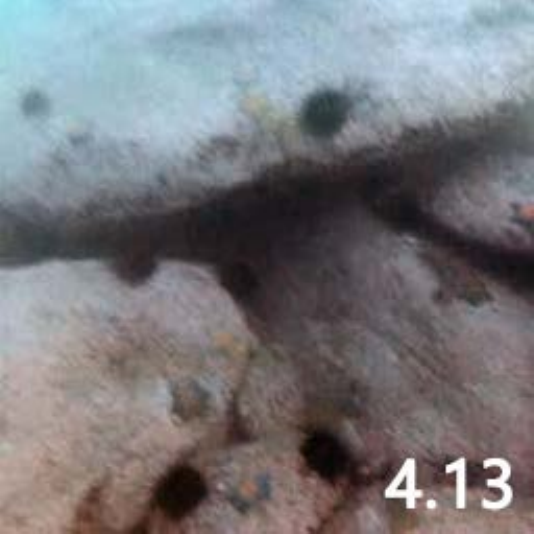}\\

		\rotatebox{90}{\quad Test-S16}
		&\includegraphics[width=0.118\textwidth]{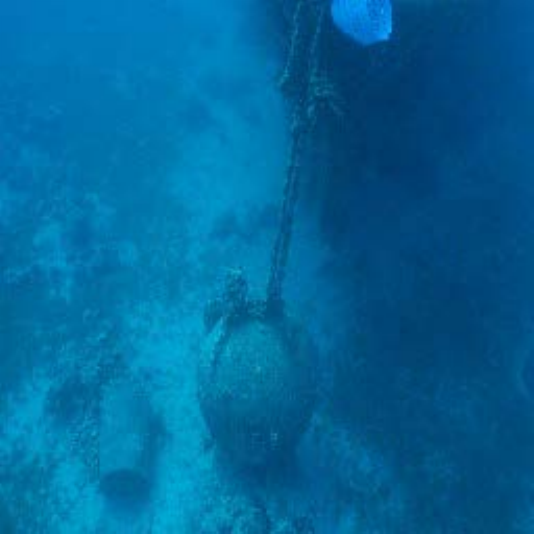}
		&\includegraphics[width=0.118\textwidth]{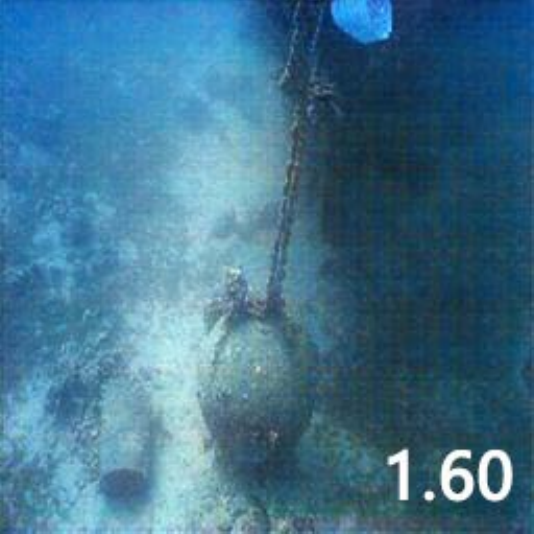}
		&\includegraphics[width=0.118\textwidth]{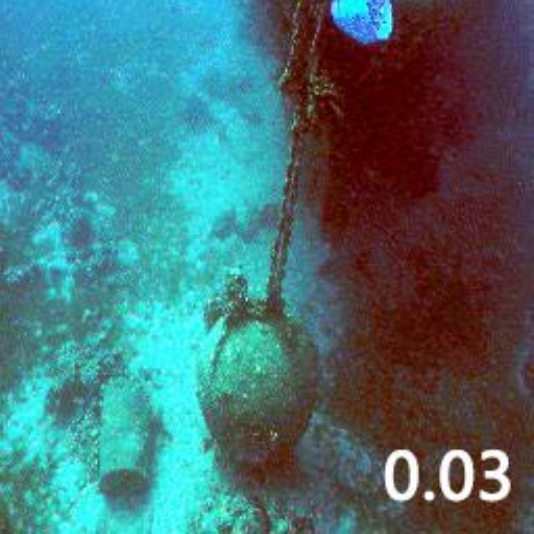}
		&\includegraphics[width=0.118\textwidth]{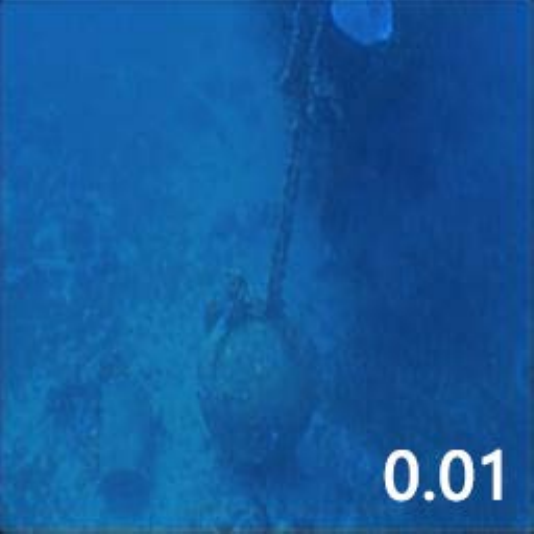}
		&\includegraphics[width=0.118\textwidth]{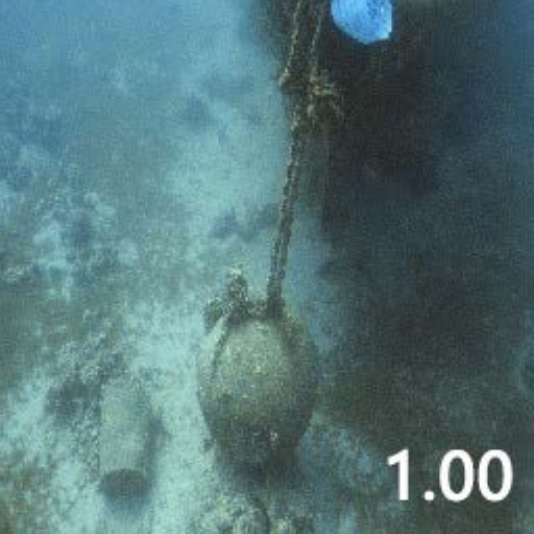}
		&\includegraphics[width=0.118\textwidth]{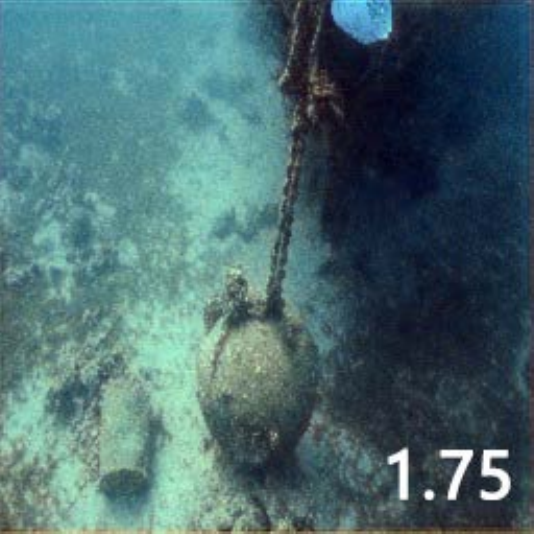}
		&\includegraphics[width=0.118\textwidth]{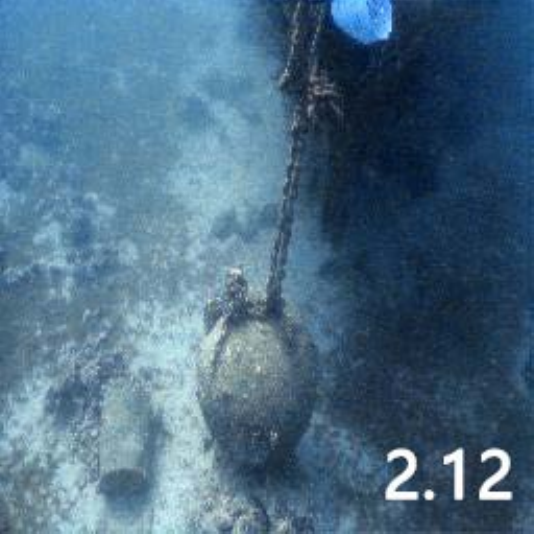}
		&\includegraphics[width=0.118\textwidth]{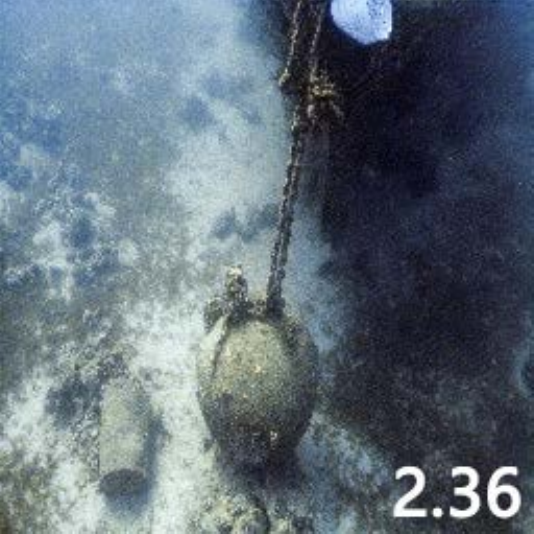}
		\\
		& \footnotesize Input 
		& \footnotesize FUnIEGAN & \footnotesize USUIR & \footnotesize  Ucolor& \footnotesize PUIENet  &  \footnotesize Uranker-NU$^2$Net  & \footnotesize Semi-UIR & \footnotesize Ours 
	\end{tabular}
	\caption{Visual comparison on datasets with no references (Test-C60, Test-R300,  Test-S16) with Uranker metric value annotated. Comparing with FUnIEGAN, USUIR, Ucolor, PUIENet, URanker, Semi-UIR, our developed method performs the best visual quality.
	}\label{fig:noreference} 
\end{figure*}

\textbf{Settings. } 
PyTorch is utilized to implement proposed network and we use NVIDIA RTX 3090 cards to perform all the experiments. 
We use Adam as optimizer while training all the models we compare. 
All the images are resized to $256 \times 256$. 
During the training process, we set the total epoch to 200, the batchsize to 8 and the initial learning rate of PMS, ADS and TDS are 1e-4, 1e-4 and 1e-6. Moreover, in Algorithm~\ref{alg:GUPDM} the update step $t_0$ is 10 and $t_1$ is 11. The coefficientes $\lambda_1, \lambda_2$ of Eq.~\ref{loss} are 0.04 and 0.02.

\textbf{Evaluation Metrics.} 
We adopt both reference-dependent and non-reference evaluation measurements to comprehensively assess the performance of our model. For reference-dependent metrics, we select Peak Signal to Noise Ratio (PSNR), Structural Similarity (SSIM) and Mean Square Error (MSE). 
As for the non-reference metrics, we employ Underwater Color Image Quality Evaluation (UCIQE \cite{2015UCIQE}), Underwater Image Quality Measure (UIQM \cite{2015UIQM}) and Perceptual Scores (PS) which reflect the visual quality of the image from a human perspective. We also adopt Natural Image Quality Evaluator (NIQE \cite{2012NIQE}), which evaluates the naturalness of generated images. Furthermore, we utilize URanker \cite{guo2022URanker}, an UIQA method based on the Transformer. We apply these metrics on real-world underwater challenge datasets.

\subsection{Comparing with Other UIE Methods}\label{subsec:comp_SOTA}

In this section, we make comparisons with traditional methods (i.e., Fusion~\cite{2012fusion} and UDCP~\cite{drews2013UDCP} ), GAN-based methods (i.e., UGAN~\cite{Fabbri2018UGAN}, FUnIEGAN~\cite{Islam2020FUnIE-GAN}) and CNN-based approaches (i.e, Water-Net~\cite{li2019WaterNet-UIEB-C60}, Ucolor~\cite{li2021Ucolor}, USUIR~\cite{2022USUIR}, PUIE\_net~\cite{fu2022puieNet}, Uranker-NU$^2$Net~\cite{guo2022URanker} and Semi-UIR~\cite{huang2023Semi-UIR}). 
All of the competitive models are retrained with their public codes on LSUI dataset for fair comparison. 

\textbf{Quantitative Comparisons.} 
We first evaluate the capacity of our method quantitatively and summarize the comparisons results in Tab.~\ref{tab:comparison_supervised} and Tab.~\ref{tab:comparison_unsupervised} where the best result is highlighted in black bold and the second one is marked underline. 
From Tab.~\ref{tab:comparison_supervised}, we can observe that, our proposed method ranks the first in Test-E515, Test-L504 and Test-U120 datasets in all three metrics. We achieve the  percentage gain of 6.5\%, 2.8\% and 5.4\% regarding PSNR and obtain leading 30.9\%, 8.4\% and 19.7\% percentage improvement regarding MSE metric in three datasets, respectively, indicating our results are more consistent with ground-truth images and our superior performance compared to other methods. 
Besides, we are only slightly behind Semi-UIR in Test-U90 dataset, with a minor gap. 
Moreover, Tab.~\ref{tab:comparison_unsupervised} demonstrates that we can achieve the best result in UIQM, NIQE and Uranker metrics, and gain the second performance in all other evaluation cases. These indicate that our method is equipped with superior performance when compared to other methods on challenge dataset using unsupervised metrics.

\textbf{Qualitative Comparisons. }  
We display the results on synthesized datasets in Fig.~\ref{fig:syn_real} and the images from datasets with no references in Fig.~\ref{fig:noreference} where we manually pick one  picture from each dataset. Besides, we annotate  the corresponding PSNR, MSE results of each image in the right bottom in Fig.~\ref{fig:syn_real} and the Uranker metric in Fig.~\ref{fig:noreference}. 
In general, our method provides images with more natural color, abundant details, and less blurs. We have the closest overall chroma to references and the best metrics as in all four rows in  Fig.~\ref{fig:syn_real}. 
In Fig.~\ref{fig:noreference}, FGAN, USUIR, Ucolor and PUIENet fail to predict the plausible color, suffering from over or under enhancement results. Uranker-NU$^2$Net and Semi-UIR achieve good performance but obtain lower value of Uranker metric as depicted in the right bottom. Our method attains the best results both visually and metrically.

\begin{figure*}[t]
	\centering 
	\begin{tabular}{c@{\extracolsep{0.2em}}c@{\extracolsep{0.2em}}c@{\extracolsep{0.2em}}c@{\extracolsep{0.2em}}c@{\extracolsep{0.2em}}c@{\extracolsep{0.2em}}c@{\extracolsep{0.2em}}c}

		&\includegraphics[width=0.135\textwidth]{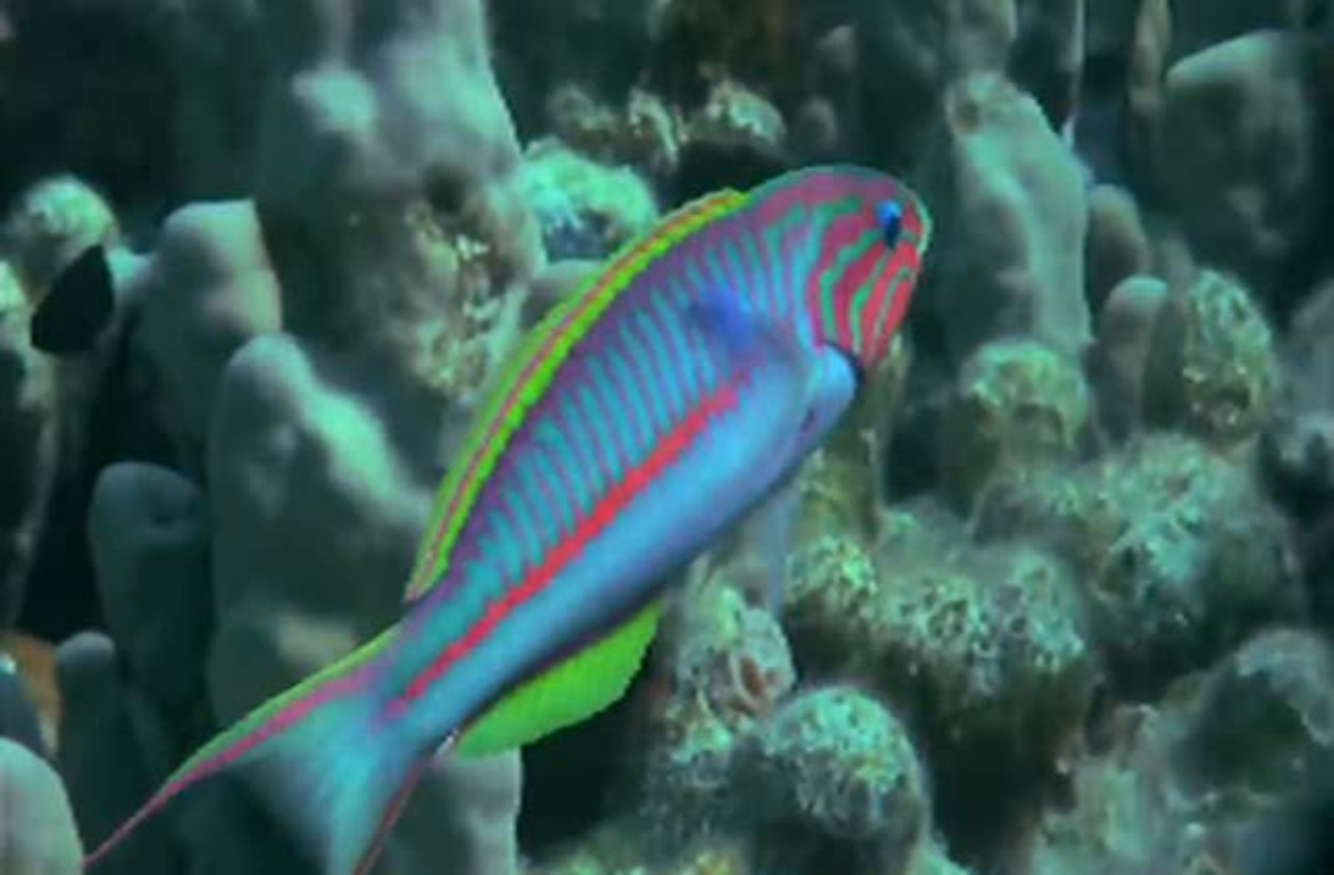}
		&\includegraphics[width=0.135\textwidth]{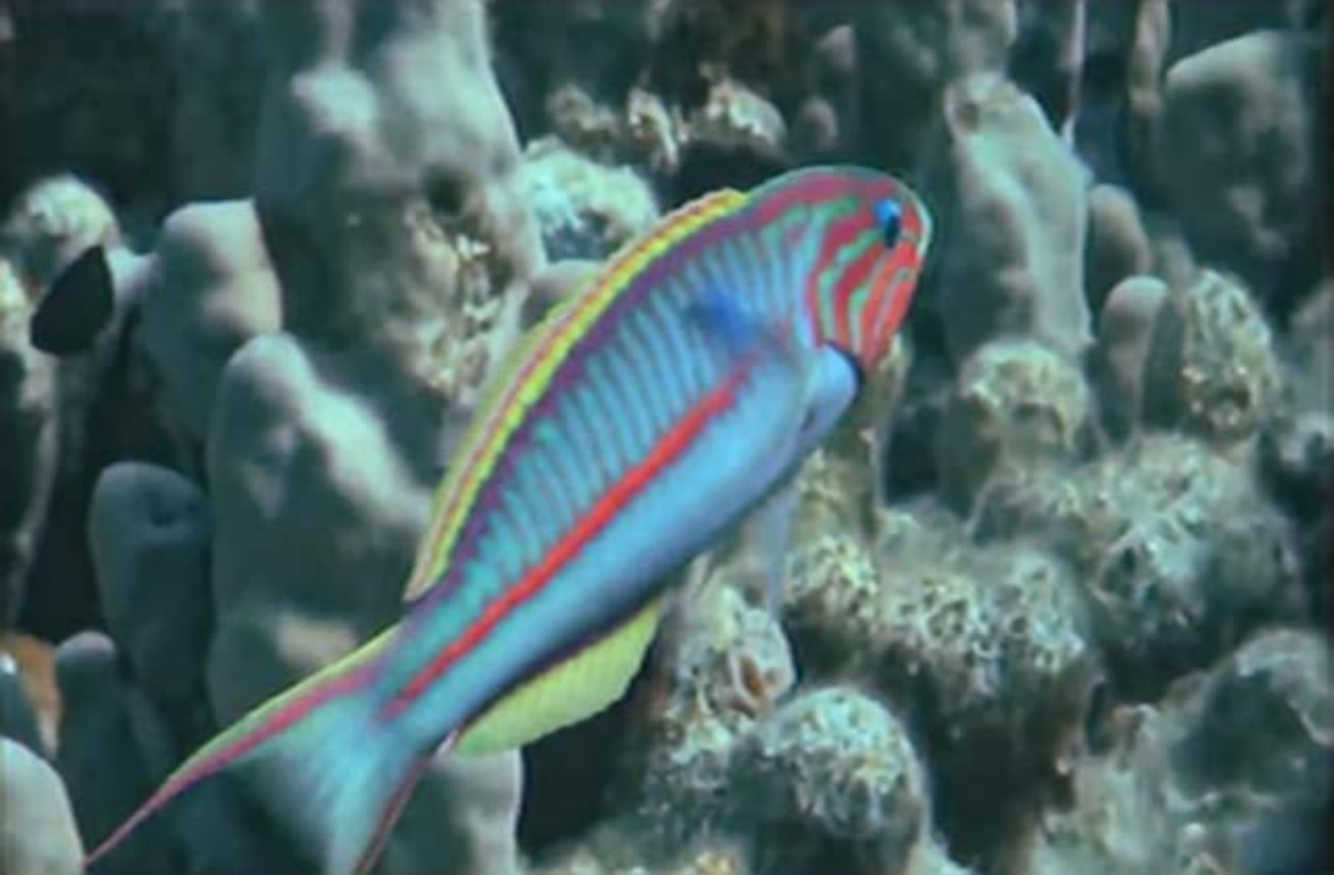}
		&\includegraphics[width=0.135\textwidth]{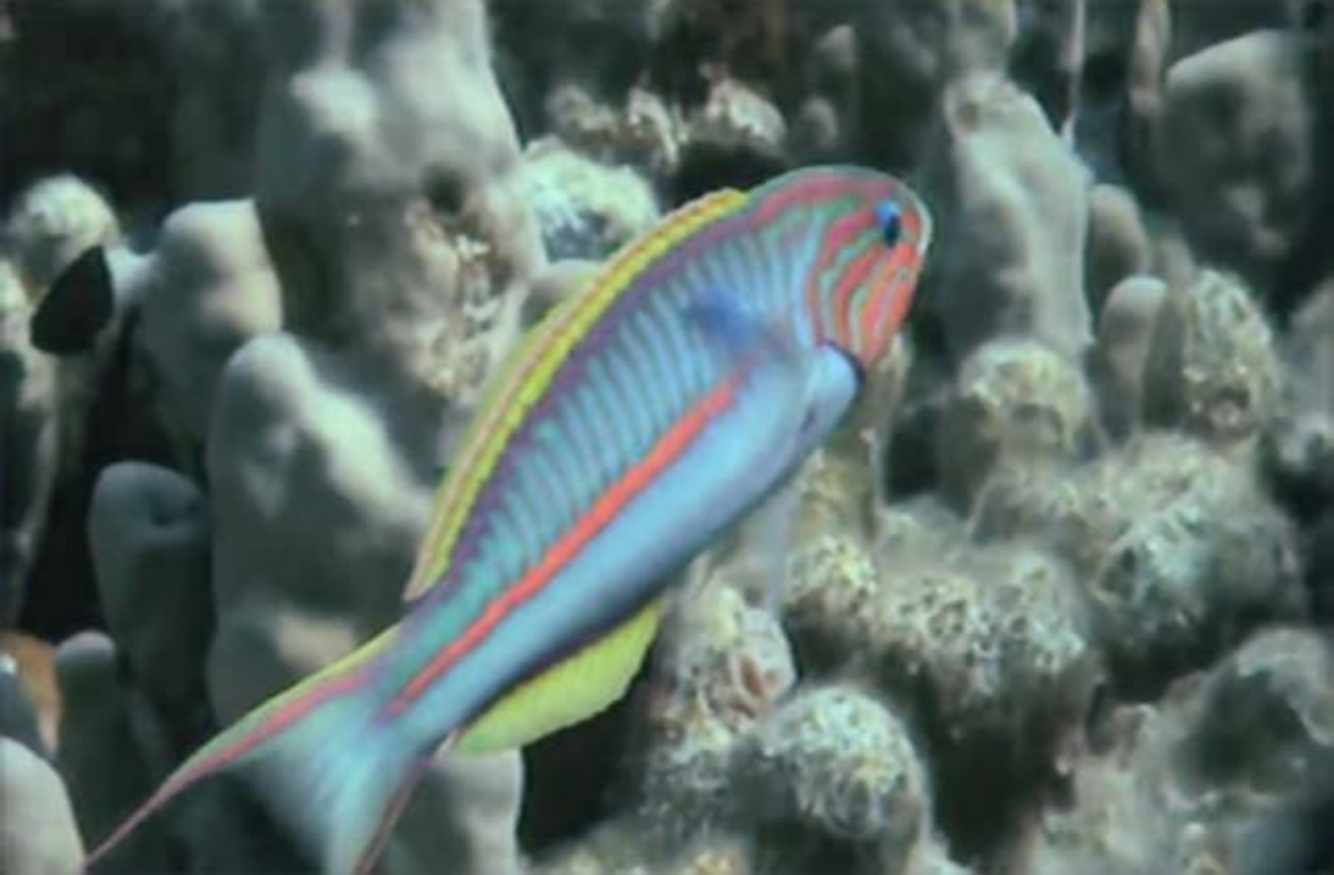}
		&\includegraphics[width=0.135\textwidth]{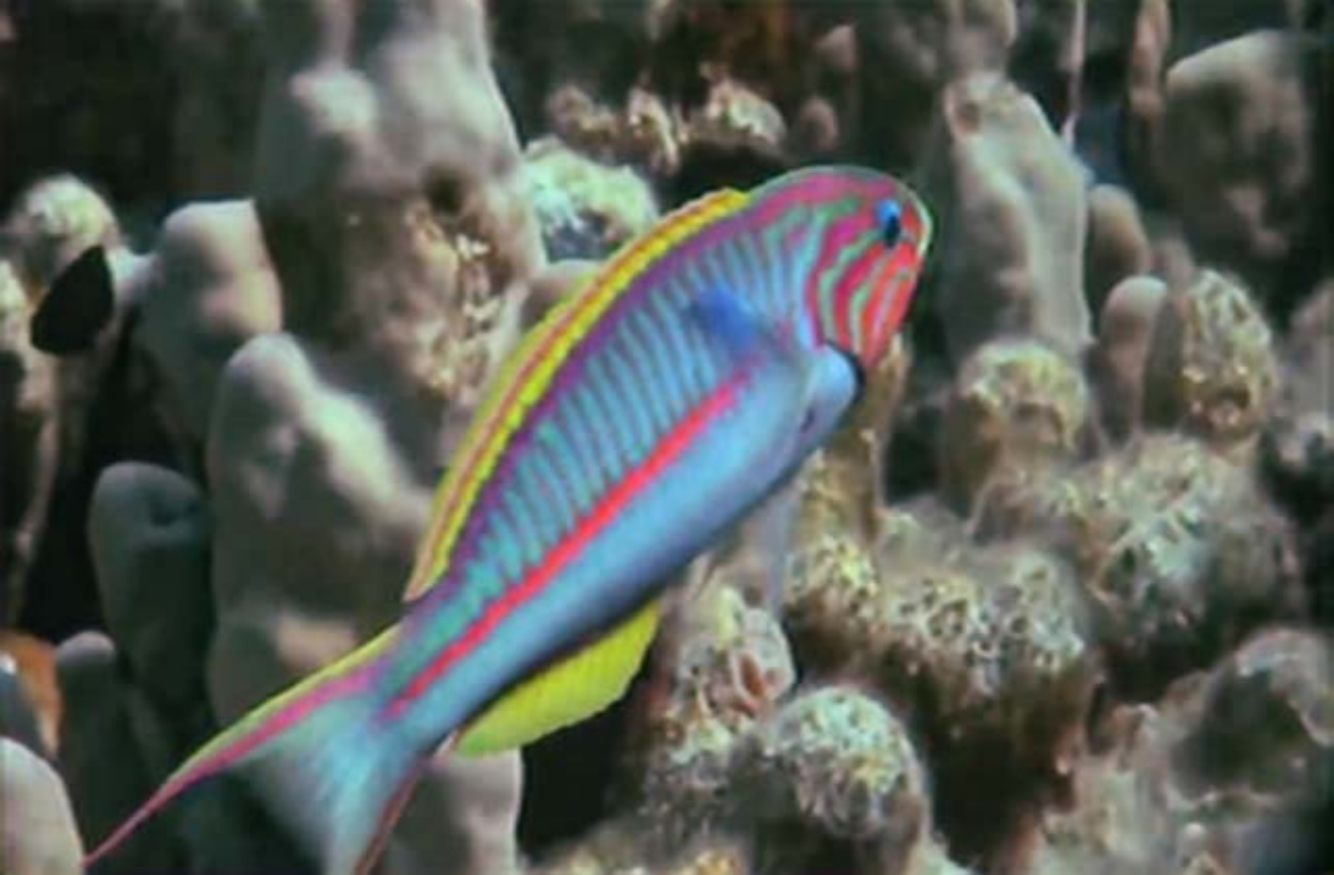}
		&\includegraphics[width=0.135\textwidth]{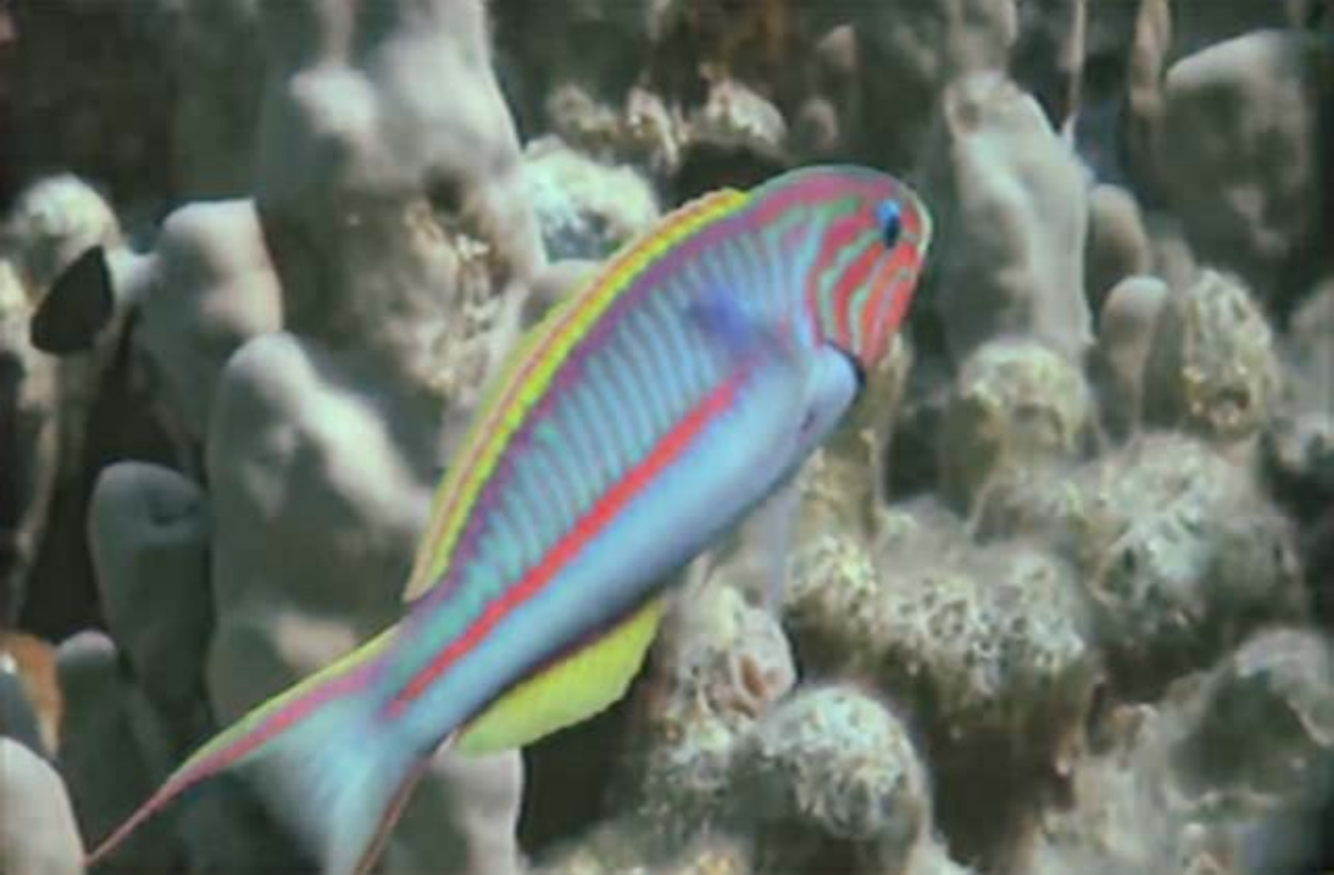}
		&\includegraphics[width=0.135\textwidth]{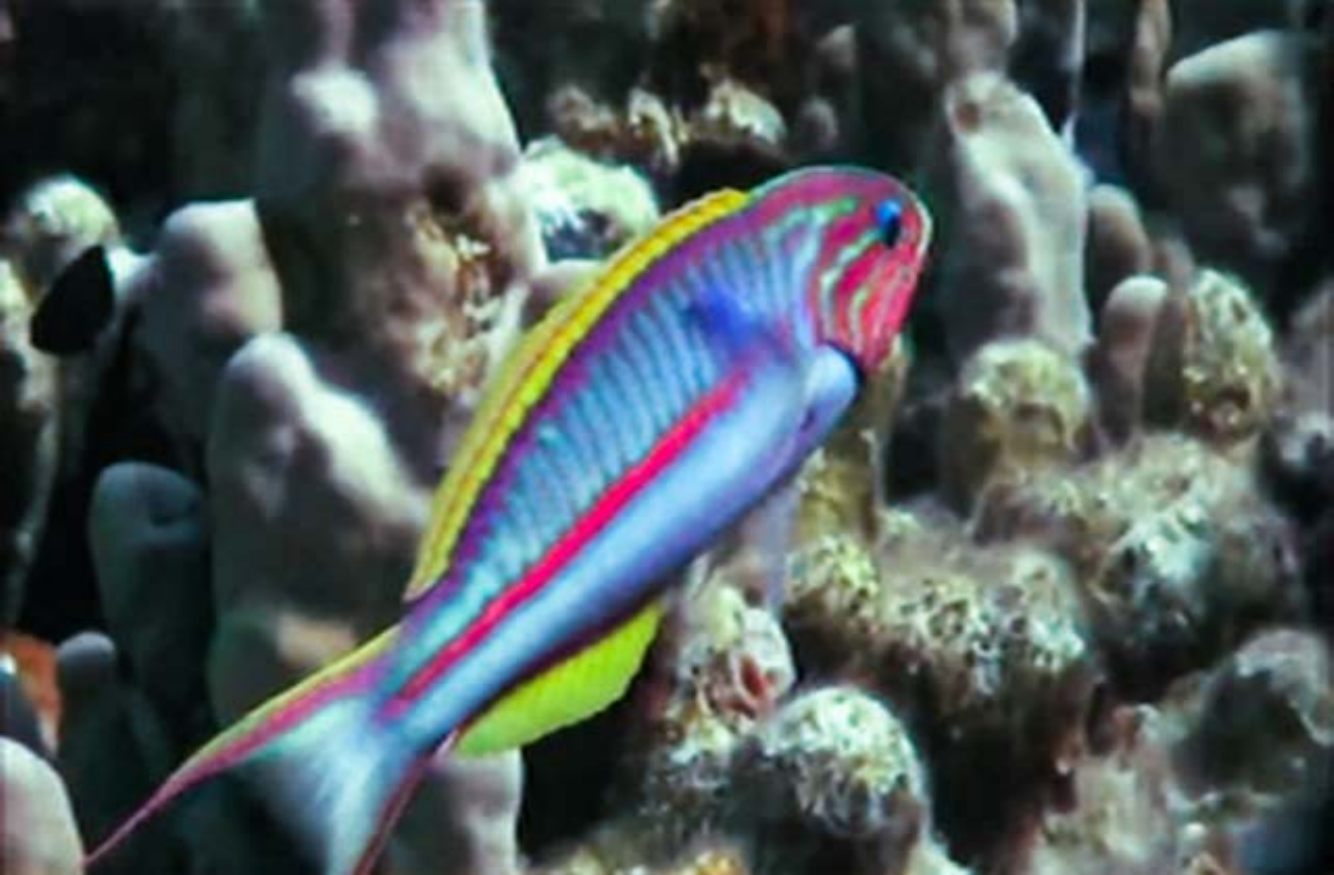}
		&\includegraphics[width=0.135\textwidth]{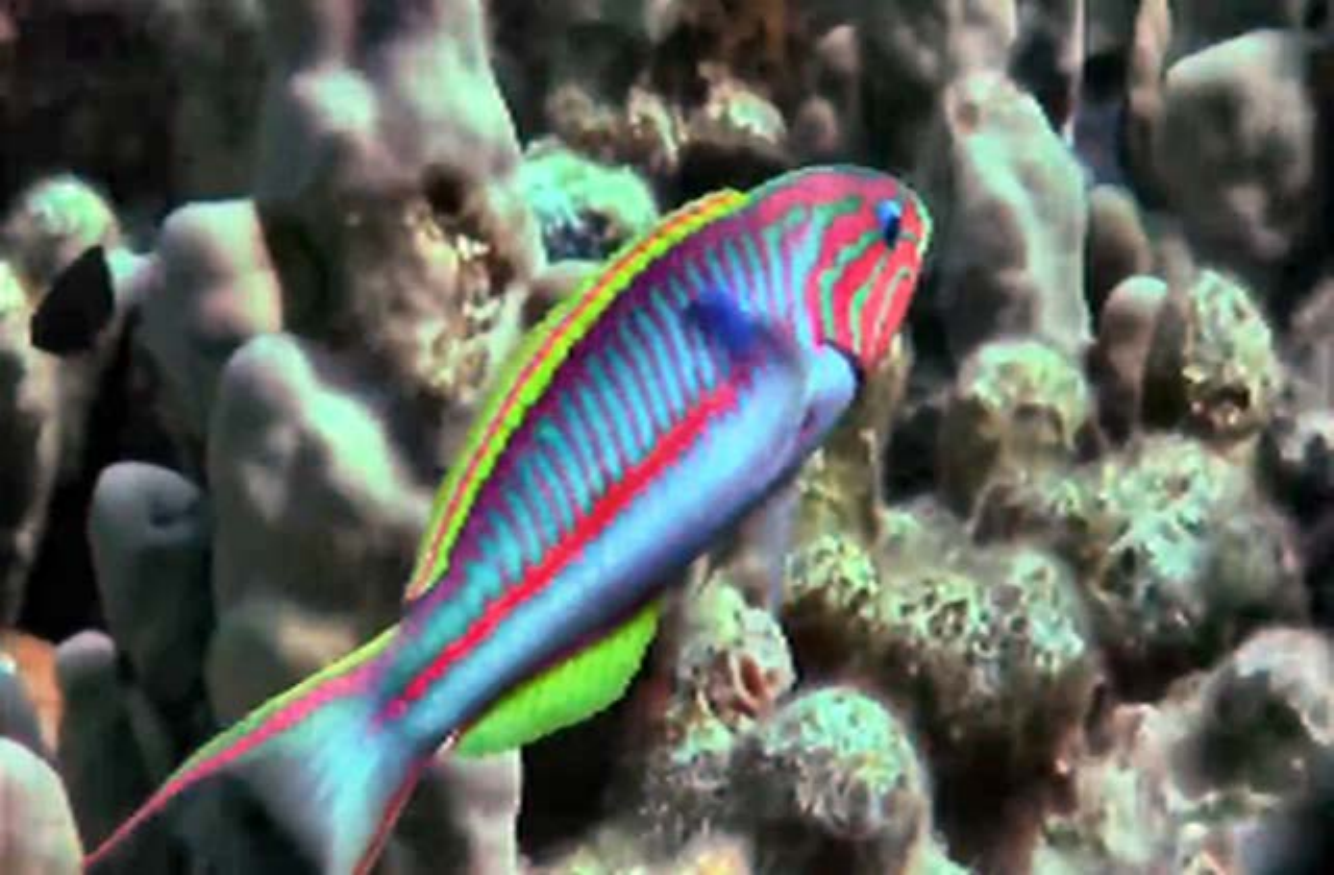}
		\\
		
		&\includegraphics[width=0.135\textwidth]{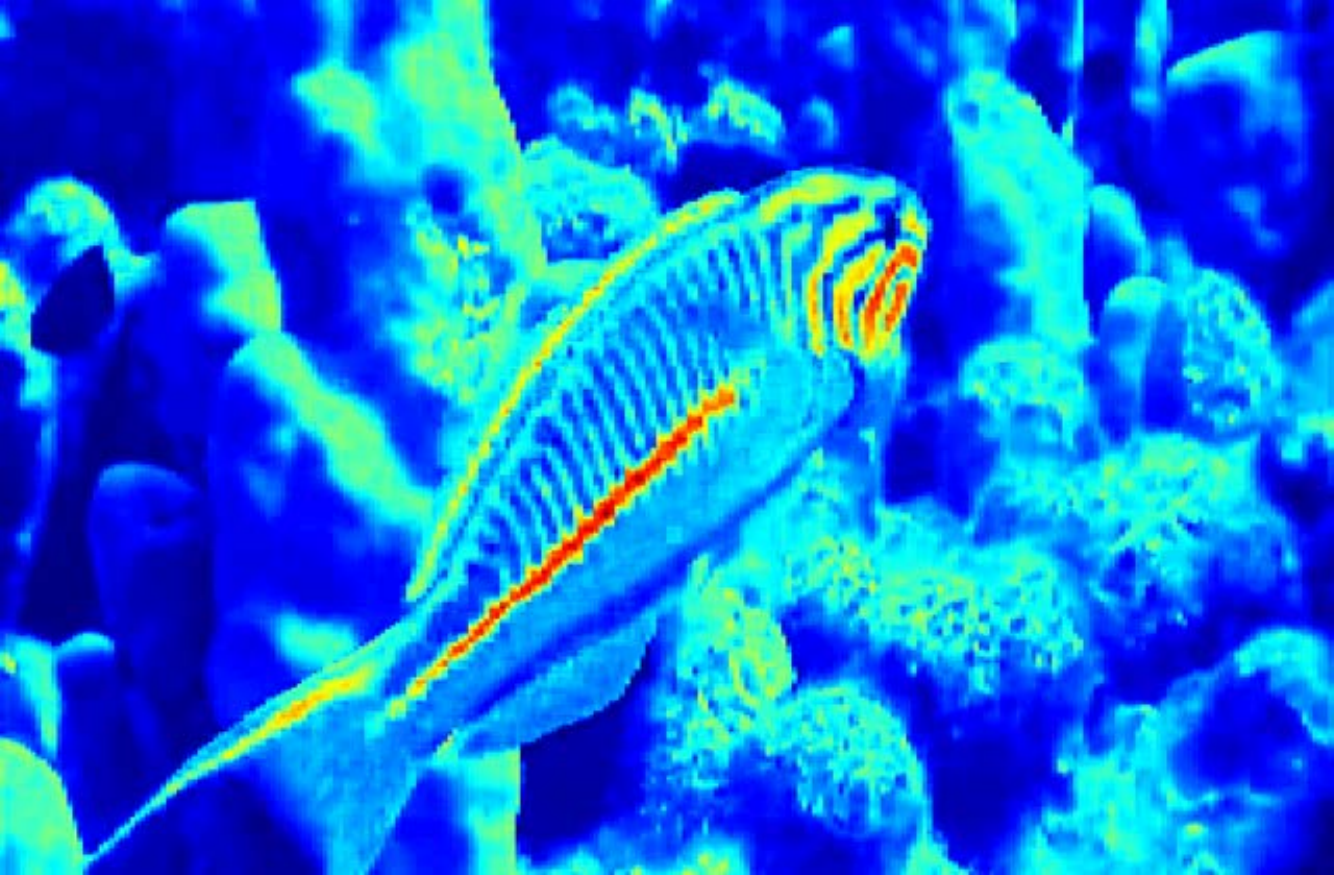}
		&\includegraphics[width=0.135\textwidth]{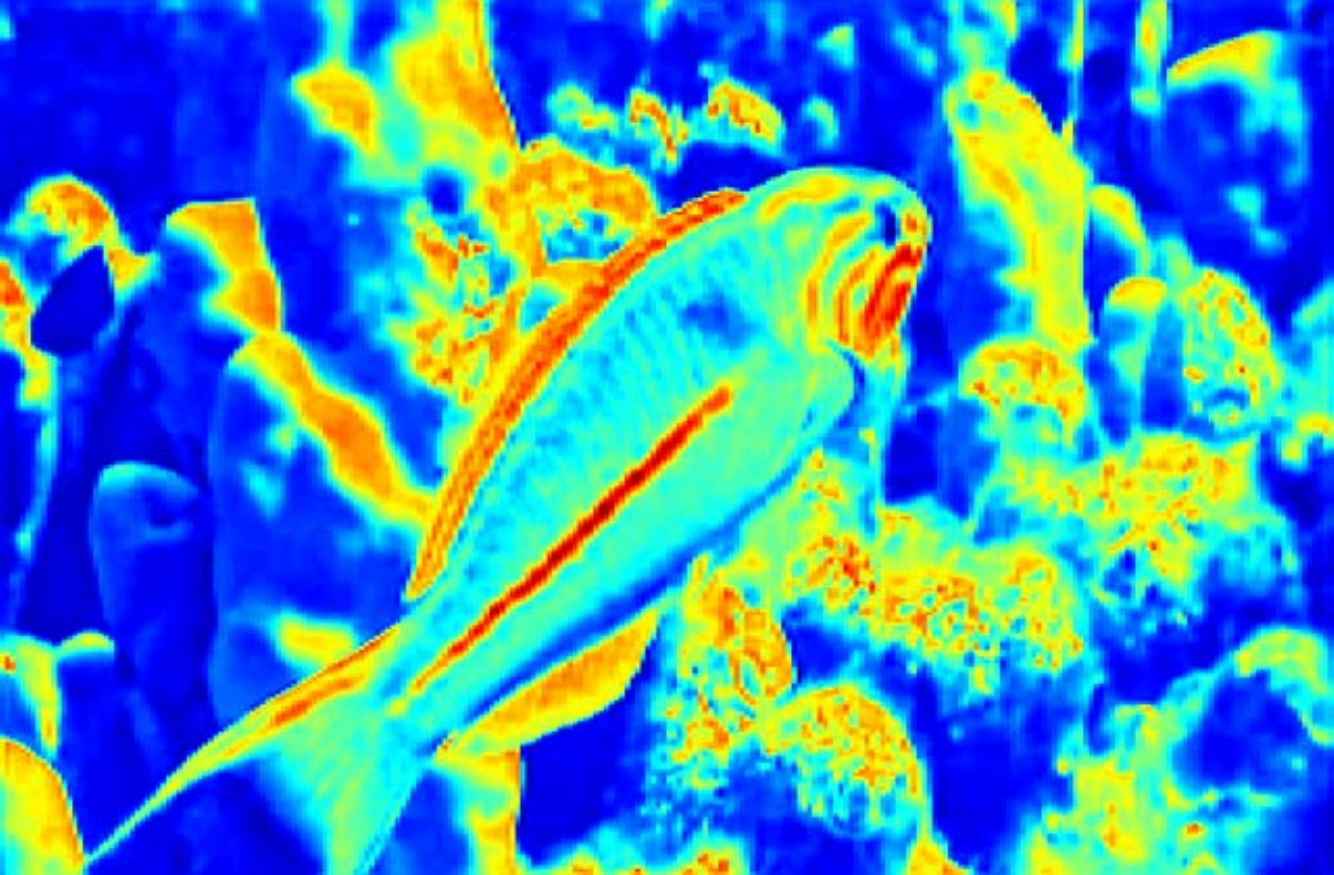}
		&\includegraphics[width=0.135\textwidth]{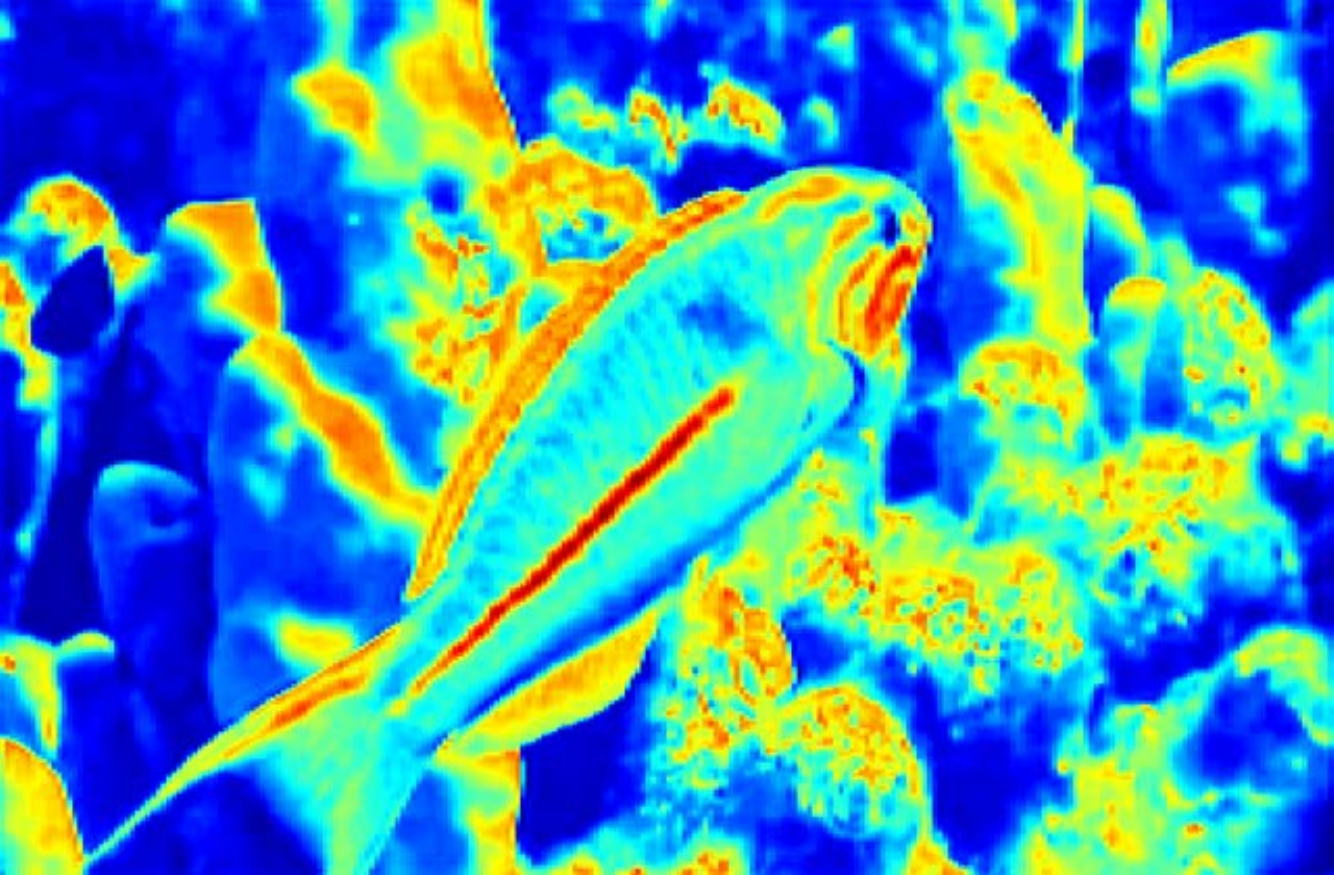}
		&\includegraphics[width=0.135\textwidth]{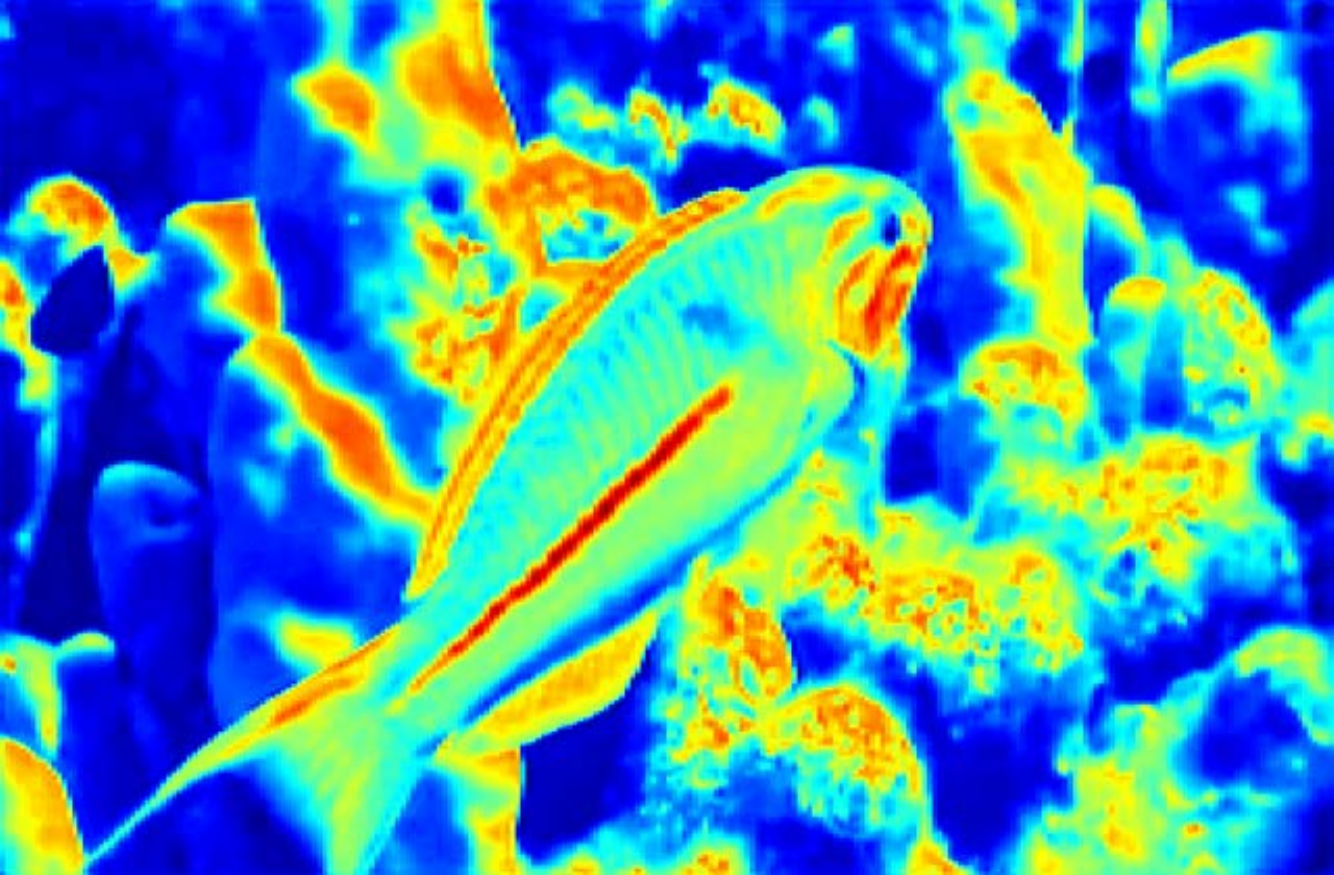}
		&\includegraphics[width=0.135\textwidth]{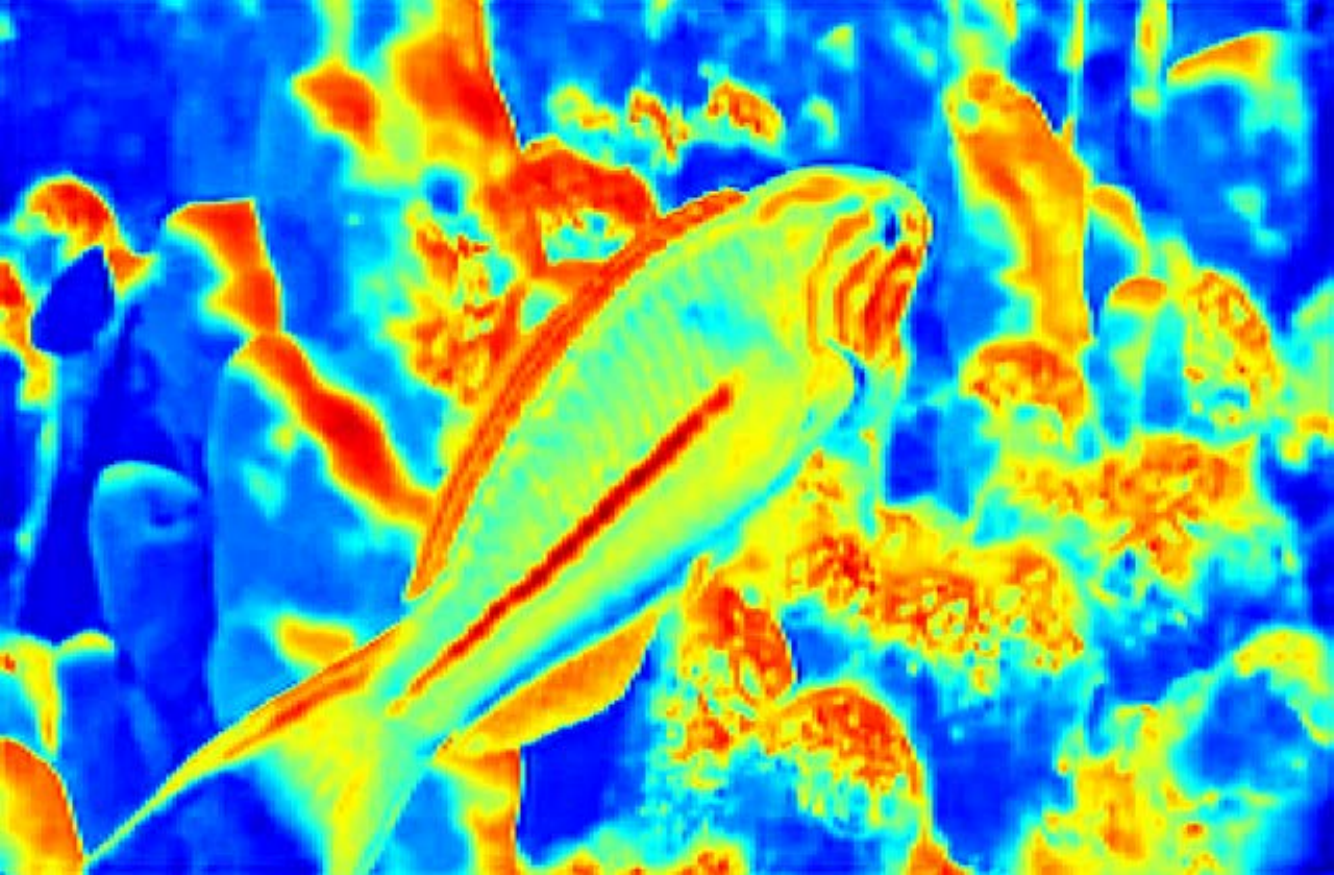}
		&\includegraphics[width=0.135\textwidth]{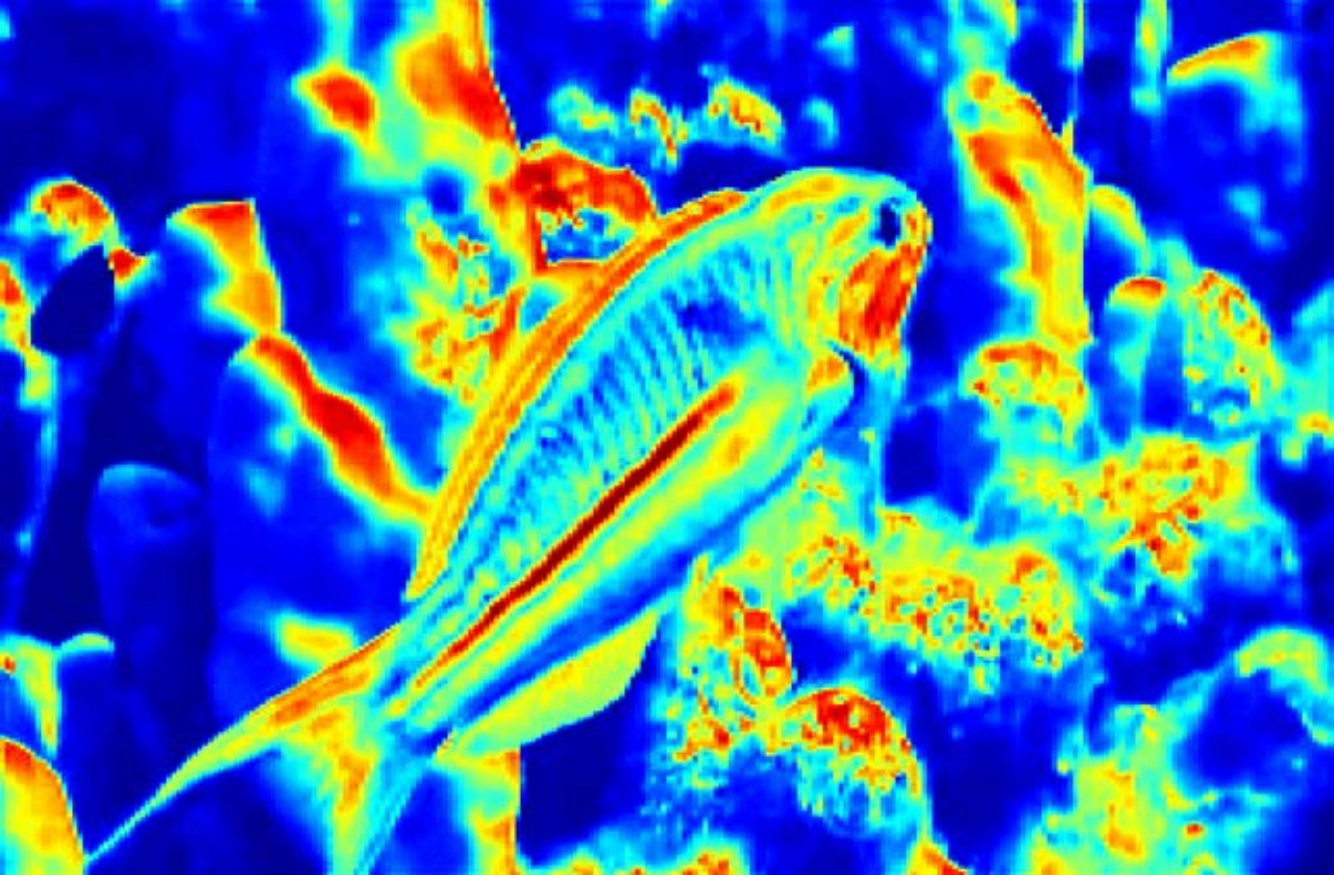}
		&\includegraphics[width=0.135\textwidth]{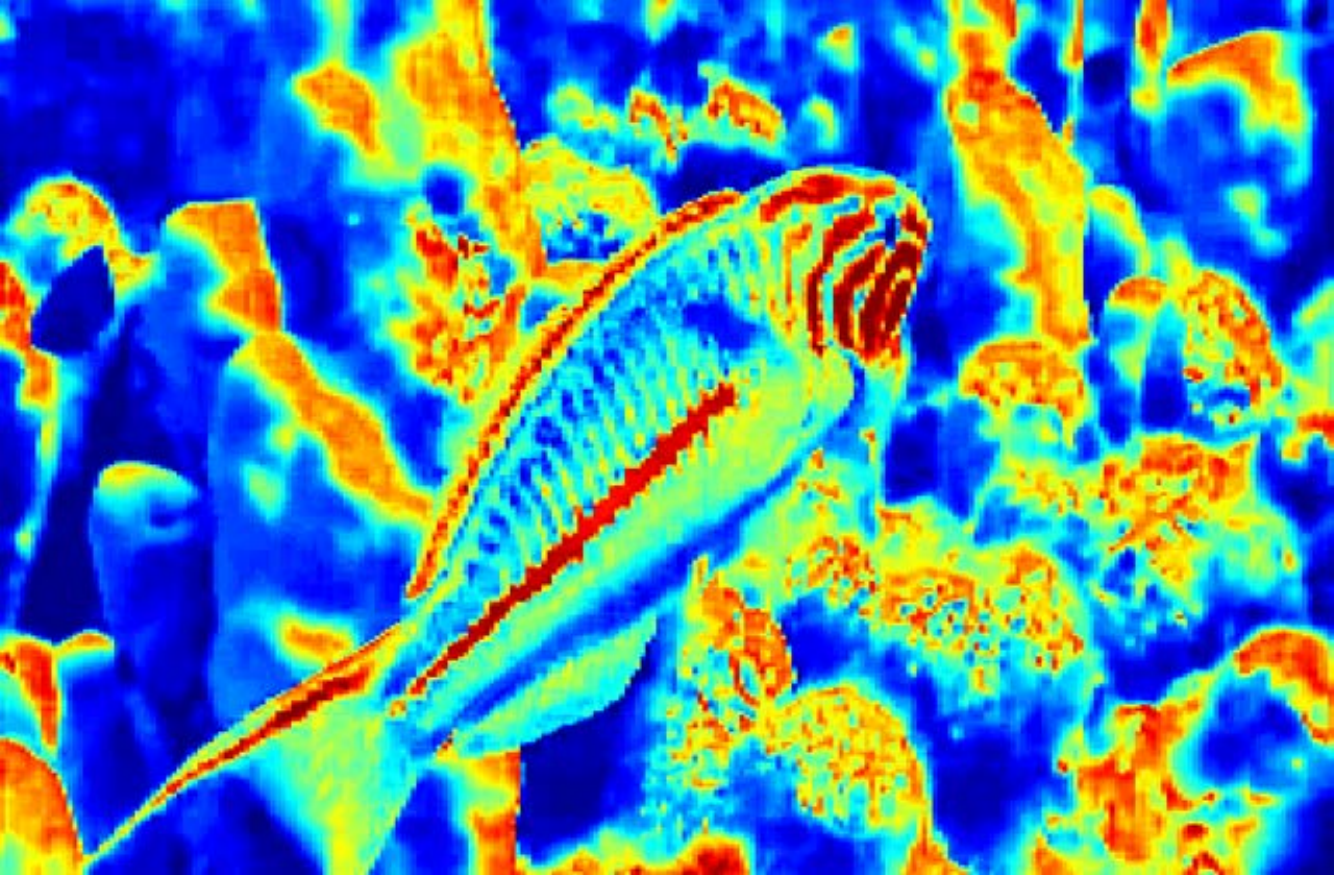}
		\\
		
		& \footnotesize Input & \footnotesize w/o ADS & \footnotesize w/o TDS & \footnotesize w/ ADS, w/o $\mathcal{G}_\mathbf{A}$  & \footnotesize  w/ TDS, w/o $\mathcal{G}_\mathbf{T}$ & \footnotesize Ours 	& \footnotesize Reference	
	\end{tabular}
	\caption{Visual results of ablation on ADS, TDS, $\mathcal{G}_\mathbf{A}$ and $\mathcal{G}_\mathbf{T}$. The second row show the corresponding heat feature map.
	}\label{fig:abl_ADS_TDS} 
\end{figure*}

\begin{figure}[t]
	\centering 
	\begin{tabular}{c@{\extracolsep{0.2em}}c@{\extracolsep{0.2em}}c@{\extracolsep{0.2em}}c@{\extracolsep{0.2em}}c}

		&\includegraphics[width=0.112\textwidth]{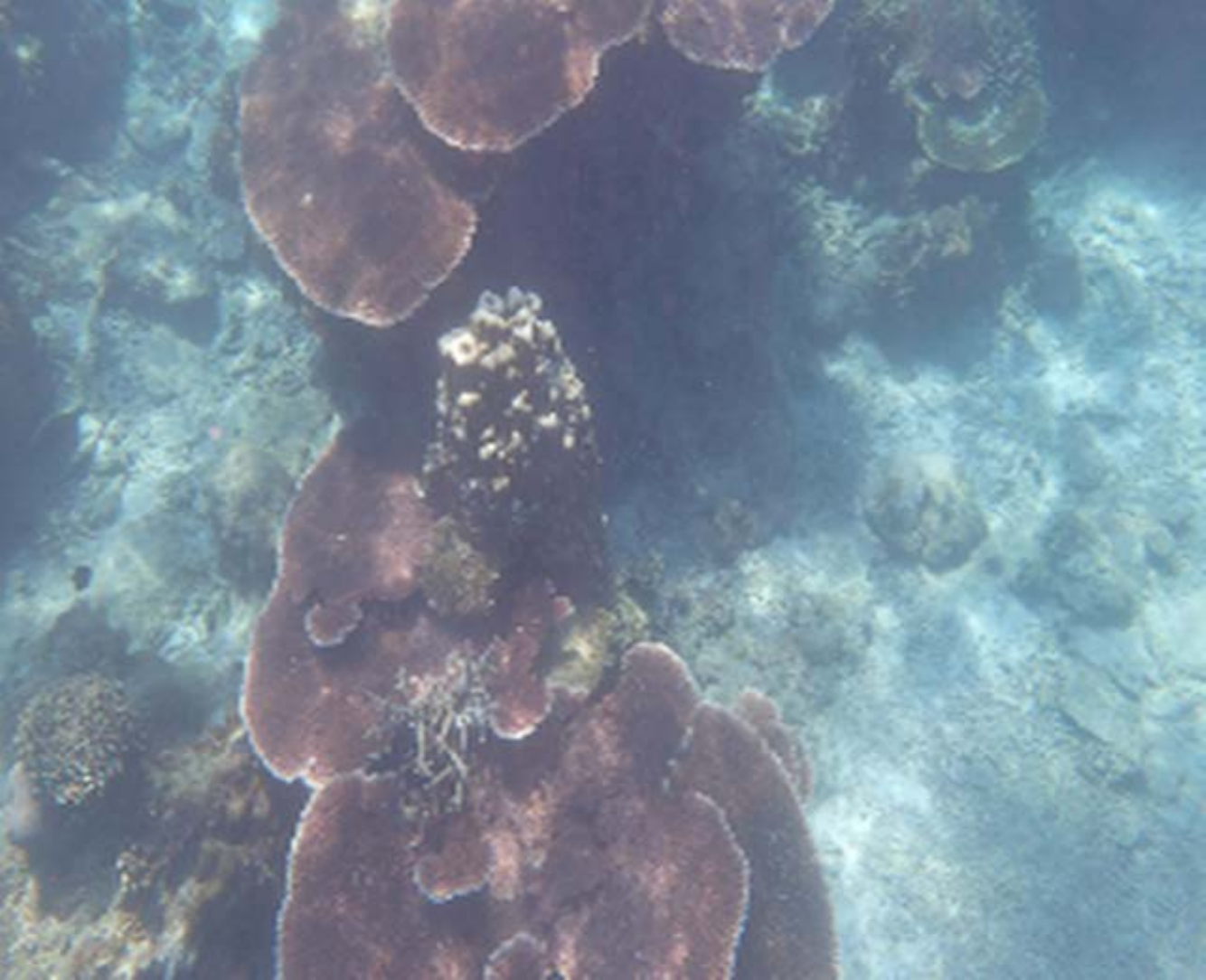}
		&\includegraphics[width=0.112\textwidth]{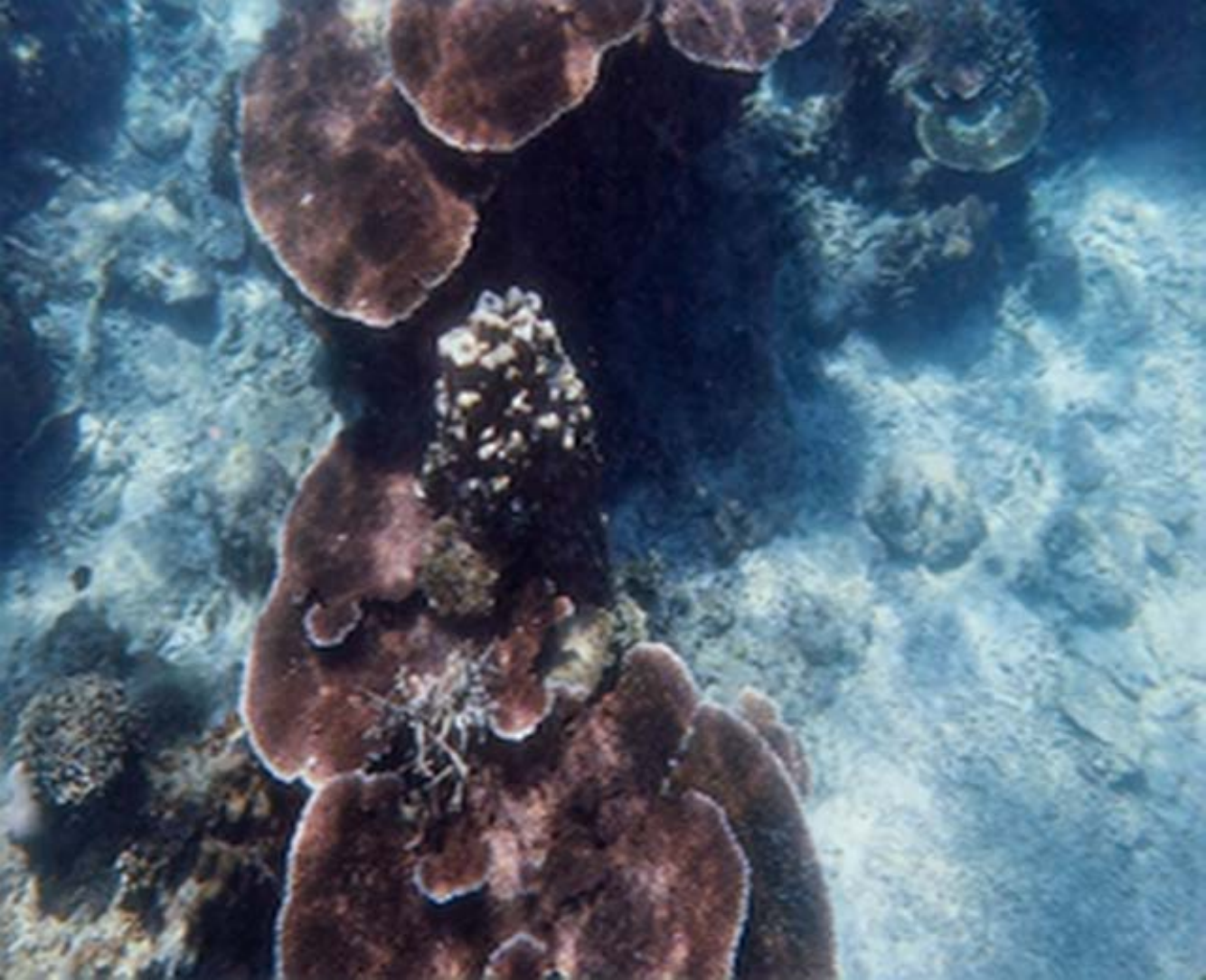}
		&\includegraphics[width=0.112\textwidth]{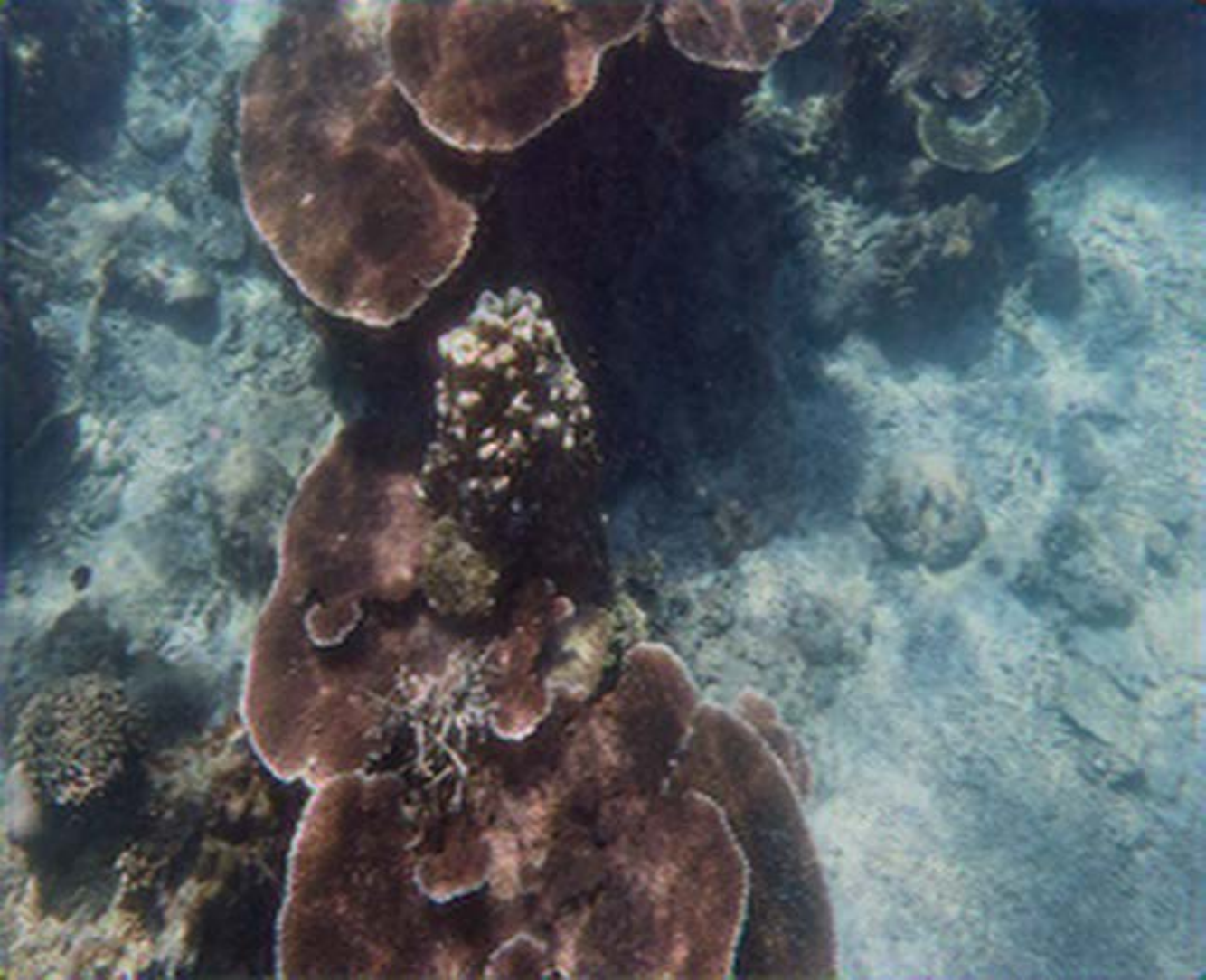}
		&\includegraphics[width=0.112\textwidth]{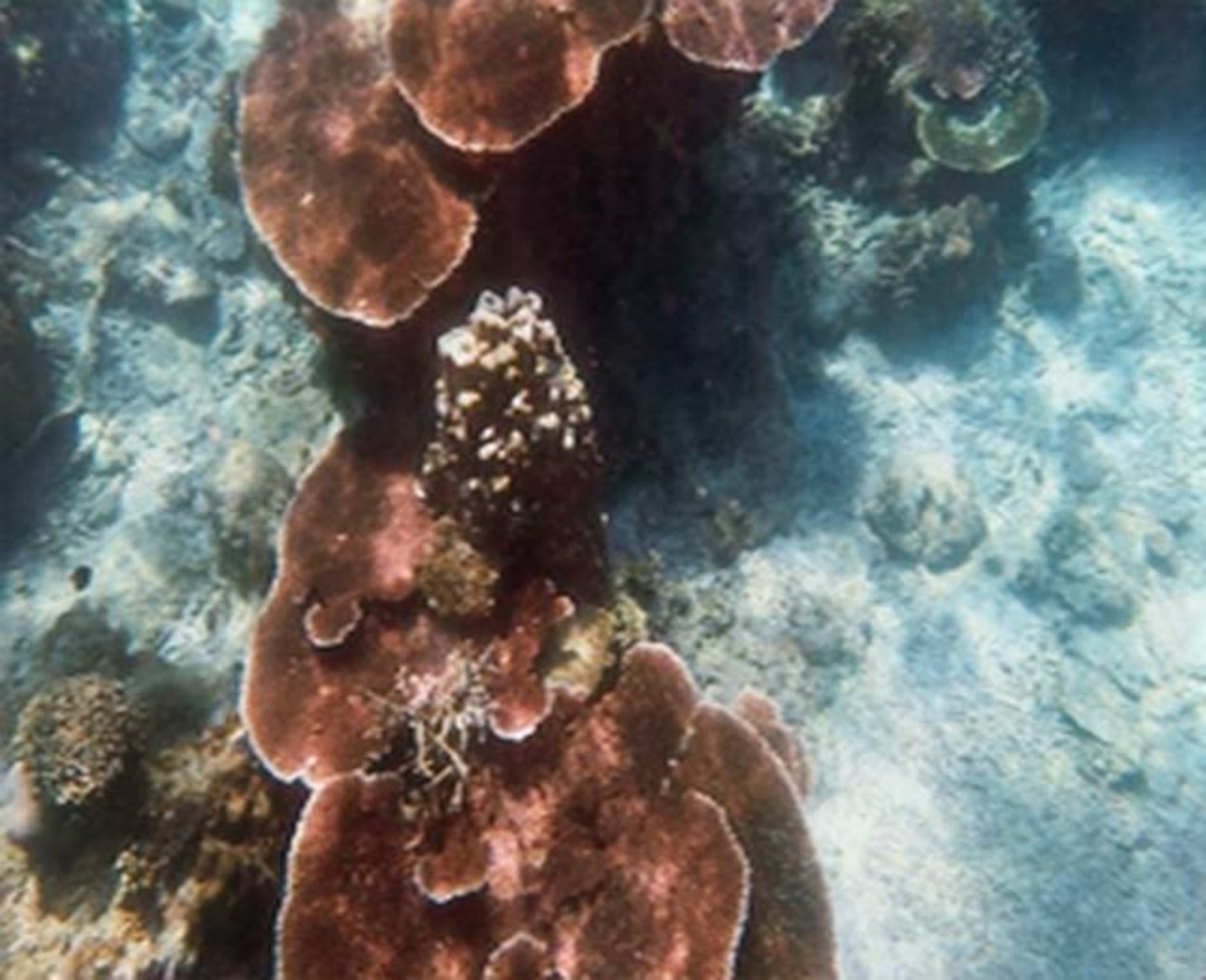}\\
		
		&\includegraphics[width=0.112\textwidth]{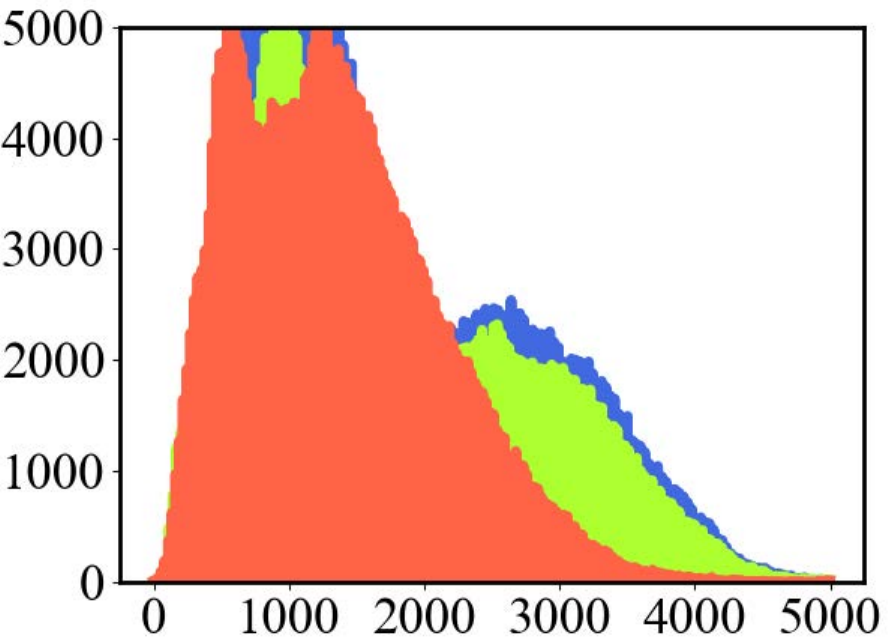}
		&\includegraphics[width=0.112\textwidth]{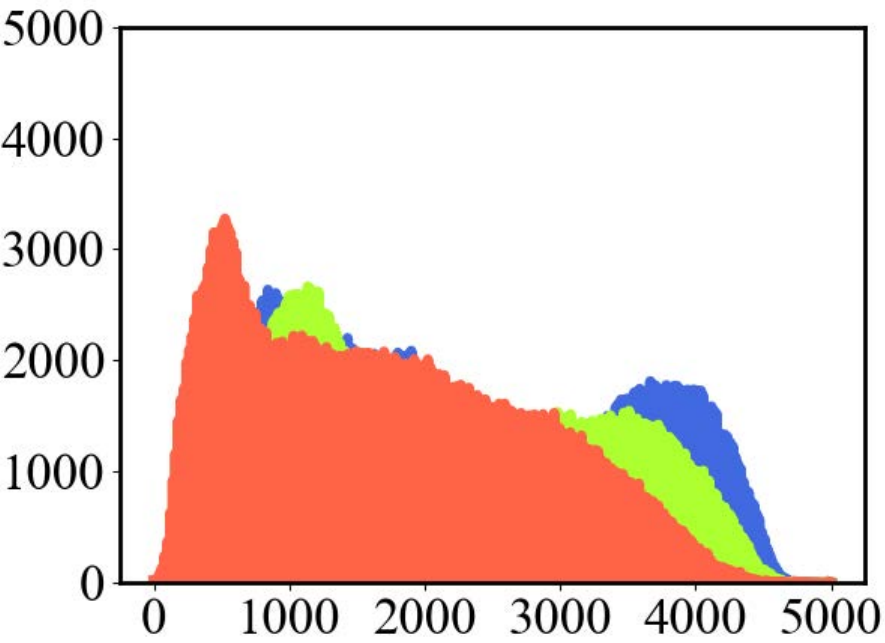}
		&\includegraphics[width=0.112\textwidth]{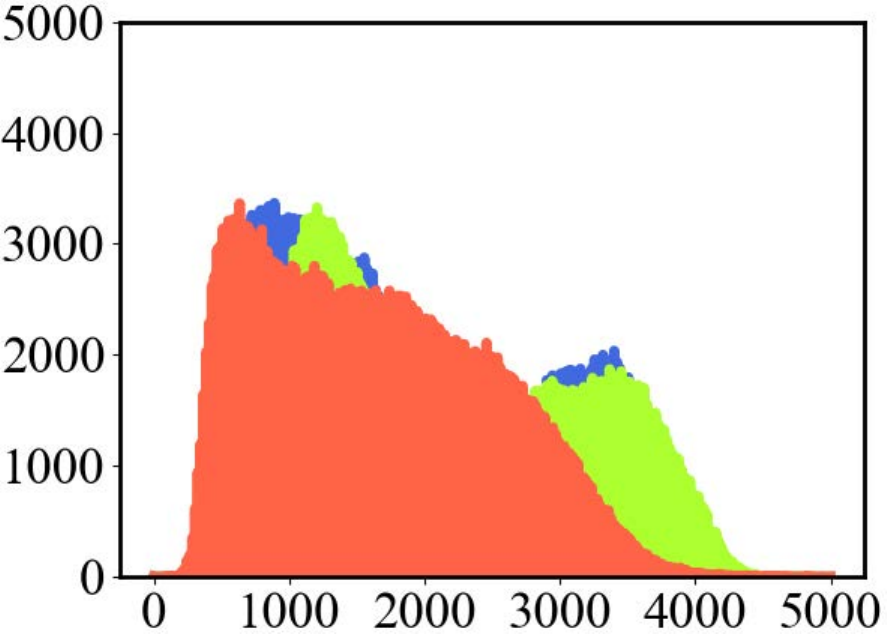}
		&\includegraphics[width=0.112\textwidth]{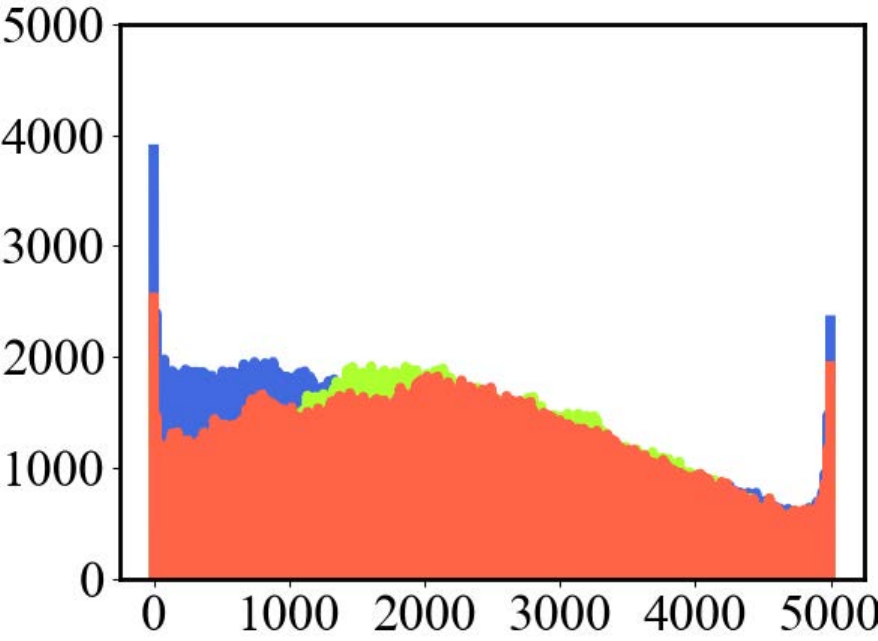}\\
		
		& \footnotesize Input & \footnotesize w/o 3H & \footnotesize w/o  HMH &  \footnotesize Ours  
	\end{tabular}
	\caption{Ablation study on 3H and HMH module. We show the corresponding histogram distance between enhanced images and references images in the second row. 
	}\label{fig:abl_3H_HWH} 
\end{figure}

\begin{table}[t]
	\centering
	\caption{Analyzing the effectiveness of ADS, TDS, $\mathcal{G}_\mathbf{A}$ and $\mathcal{G}_\mathbf{T}$.}
	\begin{tabular}{c||ccc}
		\Xhline{0.8pt}
		Baselines
		&PSNR$\uparrow$ & SSIM$\uparrow$ & MSE$\downarrow$  \\
		\hline
		w/o ADS & 0.84 &19.90 & 912   \\
		w/o TDS & 0.85 &20.67 & 789   \\
		\hline
		w/ ADS, w/o $\mathcal{G}_\mathbf{A}$  & 0.85 & 20.69 &774   \\
		w/ TDS, w/o $\mathcal{G}_\mathbf{T}$  & 0.84 & 20.30  &  830    \\
		\hline
		Ours & 0.86 & 21.53  &  721    \\
		
		\Xhline{0.8pt}
	\end{tabular}
\label{tab:abl_branch}
\end{table}

\begin{table}[t]
\centering
\caption{Analyzing the effectiveness of different Modules (including TGFE, 3H and HMH) on UIEB dataset.}
\begin{tabular}{c|c||ccc}
	\Xhline{1pt}
	\multirow{2}{*}{Modules} &\multirow{2}{*}{Ablation} & \multicolumn{3}{c}{Metrics} \\
	\cline{3-5}
	& &PSNR$\uparrow$ & SSIM$\uparrow$ & MSE$\downarrow$  \\
	\hline
	\multirow{4}{*}{TGFE} & w/o TGFE & 19.82 & 0.83 & 941   \\
	& w/o TC & 21.01 & 0.85 & 745   \\
	& w/o MFE & 20.86 & 0.85 & 780   \\
	& w/o HRB & 20.92 & 0.85  &  727    \\
	\hline
	
	3H & w/o 3H & 20.86 &0.85  & 764 \\
	
	\hline
	HMH& w/o HMH &  18.41  & 0.78 & 1143   \\
	\hline                    
	Ours & -  & 21.53    &  0.86   &  721   \\
	\Xhline{1pt}
\end{tabular}
\label{tab:abl_block}
\end{table}

\begin{table}[t]
\centering
\caption{Analyzing the effectiveness of $\mathbf{f}_{\text{ADS}}$ and $\mathbf{f}_{\text{TDS}}$.}
\begin{tabular}{c|c||ccc}
\Xhline{1pt}
\multirow{2}{*}{Modules} &\multirow{2}{*}{Baselines} & \multicolumn{3}{c}{ Average Metrics} \\
\cline{3-5}
& &PSNR$\uparrow$ & SSIM$\uparrow$ & MSE$\downarrow$  \\
\hline
3H  & w/o $\mathbf{f}_{\text{ADS}}$  & 0.85 & 23.87 & 404 \\

HMH & w/o $\mathbf{f}_{\text{ADS}}$ & 0.85 &23.85 & 411    \\
TGFE& w/o $\mathbf{f}_{\text{TDS}}$  & 0.84 & 23.65  &  427    \\
\hline                    
\multirow{2}{*}{	Full Model}  &   w/o $\mathbf{f}_{\text{ADS}}$,  w/o $\mathbf{f}_{\text{TDS}}$ & 0.84 &22.99 & 491   \\
&   w/ $\mathbf{f}_{\text{ADS}}$,  w/ $\mathbf{f}_{\text{TDS}}$  & 0.86 & 24.17  &  383   \\
\Xhline{1pt}
\end{tabular}
\label{tab:abl_meta}
\end{table}

\begin{table}[t]
\centering
\caption{Analyzing the effectiveness of training schema.}
\begin{tabular}{c||ccc}
\Xhline{0.8pt}
Baselines
&PSNR$\uparrow$ & SSIM$\uparrow$ & MSE$\downarrow$  \\
\hline
(a) end-to-end  & 0.84 & 20.09 &901   \\
(b) ADS and TDS together & 0.85 & 20.95 &726  \\
(c) ADS, TDS and PMS together &  0.85 &20.89 & 743   \\

\hline
(d) ADS, TDS and PMS separate & 0.86 & 21.53  &  721    \\

\Xhline{0.8pt}
\end{tabular}
\label{tab:abl_training}
\end{table}

\subsection{Ablation Study}\label{subsec:ablation}

\textbf{Analysis the effectiveness of ADS and TDS. }
We investigate the effectiveness of physical knowledge-based dynamic structure (i.e., ADS and TDS). We show the statistical results in Tab.~\ref{tab:abl_branch} and visual images in Fig.~\ref{fig:abl_ADS_TDS}. Without the whole ADS branch, the model cannot capture the atmosphere information. Without TDS branch providing ample transmission information, the performance of our model also deteriorates, resulting in less vivid images. Besides, ablating  $\mathcal{G}_\mathbf{A}$ and $\mathcal{G}_\mathbf{T}$ that generates various degraded priors  deprives the model from learning diverse atmosphere and transmission scenarios in ADS and TDS, respectively. 
Thus the produced images lack detail as shown in the body of fish in the heat map in  second row of Fig.~\ref{fig:abl_ADS_TDS}.

\textbf{Analysis the components of PMS. }
To verify how each module in our method plays a irreplaceable role, we ablate: the Transmission-map Guided Feature Extraction (TGFE), 3H and HMH. Furthermore, we exam the Transmission-based Connection (TC) block, Multi-scale Feature extraction (MFE) module and Hyper Residual Block (HRB) in TGFE to dig the effectiveness of each module. 
Note that when ablating the 3H and HMH module, we change all four modules posed in PMS. 
The corresponding results are listed in Tab.~\ref{tab:abl_block}.

Without MFE module extracting multiple feature from distinct scales, the performance of our model decreases which is reflected in reported scores. 
3H module, consisting of three hyper residual block, obtains weights from dynamic prior hyper net and serves as a color optimizer. Thus removing them leads to suboptimal scores. Furthermore, we display the visual results in Fig.~\ref{fig:abl_3H_HWH} which demonstrates that, without 3H and HMH integrating prior knowledge, the generated images are more blur and dull, the histogram distance to references is also worsen.

\textbf{Influence of $\mathbf{f}_{\text{ADS}}$ and $\mathbf{f}_{\text{TDS}}$ to generalization capacity. } 
We discuss the impact of $\mathbf{f}_{\text{ADS}}$ and $\mathbf{f}_{\text{TDS}}$ here. Note that removing them means that the weights of res-blocks in each setting are directly learned form training phrase, not obtained from hyper-nets. 
We perform experiments on res-blocks in 3H, HMH and TGFE. All  ablations are conducted on LSUI dataset for training and evaluated on four datasets: Test-E515, Test-U90, Test-L504 and Test-U120. 
We compute the average scores regarding the PSNR, SSIM and MSE metrics on four datasets  and report them in Tab.~\ref{tab:abl_meta}. 

We observe that, the full model ranks the first in all three measurements, indicating that we achieve the best generalization ability among datasets.
Without $\mathbf{f}_{\text{ADS}}$ or $\mathbf{f}_{\text{TDS}}$ in any position, the performance deteriorates.
Therefore, we keep $\mathbf{f}_{\text{ADS}}$ and $\mathbf{f}_{\text{TDS}}$ in those modules to ensure a better  generalization capability.

\textbf{Analysis on training strategy. }
We list different training methods and their results in Tab. ~\ref{tab:abl_training}. 
Specifically, 
(a) means that we train two hypernets and PMS in an end-to-end manner.
(b) means that we optimize ADS and TDS together, and then optimize PMS alternately.
(c) means that we optimize three networks (ADS, TDS and PMS) all together.
(d) is our strategy which separately optimizes three networks.
It can be seen that the our training strategy is of great important. We argue that this can be attributed to the inconsistent convergence speed of the hyper-nets and the main net. Thus, we train and optimize three nets in an alternate and separate way.

\section{Conclusion}

In this work we developed a novel approach named GUPDM for underwater image enhancement. We tailored ADS and TDS as hyper-nets to generate parameters for PMS via varied global atmosphere and transmission to cover complex underwater scenes.
Experimental results illustrated our superior performance in terms of quantitative scores, visual results and generalization ability.

\begin{acks}
This work is supported by Natural Science Foundation of China (Grant No. 62202429, U20A20196) and Zhejiang Provincial Natural Science Foundation of China under Grant No. LY23F020024, LR21F020002, LY23F020023.
\end{acks}

\bibliographystyle{ACM-Reference-Format}
\balance
\bibliography{sample-base}


\begin{thebibliography}{52}


\ifx \showCODEN    \undefined \def \showCODEN     #1{\unskip}     \fi
\ifx \showDOI      \undefined \def \showDOI       #1{#1}\fi
\ifx \showISBNx    \undefined \def \showISBNx     #1{\unskip}     \fi
\ifx \showISBNxiii \undefined \def \showISBNxiii  #1{\unskip}     \fi
\ifx \showISSN     \undefined \def \showISSN      #1{\unskip}     \fi
\ifx \showLCCN     \undefined \def \showLCCN      #1{\unskip}     \fi
\ifx \shownote     \undefined \def \shownote      #1{#1}          \fi
\ifx \showarticletitle \undefined \def \showarticletitle #1{#1}   \fi
\ifx \showURL      \undefined \def \showURL       {\relax}        \fi
\providecommand\bibfield[2]{#2}
\providecommand\bibinfo[2]{#2}
\providecommand\natexlab[1]{#1}
\providecommand\showeprint[2][]{arXiv:#2}

\bibitem[Akkaynak and Treibitz(2018)]%
        {2018formation}
\bibfield{author}{\bibinfo{person}{Derya Akkaynak} {and} \bibinfo{person}{Tali
  Treibitz}.} \bibinfo{year}{2018}\natexlab{}.
\newblock \showarticletitle{A revised underwater image formation model}. In
  \bibinfo{booktitle}{\emph{Proceedings of the IEEE conference on Computer
  Vision and Pattern Recognition}}. \bibinfo{pages}{6723--6732}.
\newblock


\bibitem[Akkaynak and Treibitz(2019a)]%
        {akkaynak2019sea}
\bibfield{author}{\bibinfo{person}{Derya Akkaynak} {and} \bibinfo{person}{Tali
  Treibitz}.} \bibinfo{year}{2019}\natexlab{a}.
\newblock \showarticletitle{Sea-thru: A method for removing water from
  underwater images}. In \bibinfo{booktitle}{\emph{Proceedings of the IEEE/CVF
  conference on Computer Vision and Pattern Recognition}}.
  \bibinfo{pages}{1682--1691}.
\newblock


\bibitem[Akkaynak and Treibitz(2019b)]%
        {2019Sea-thru}
\bibfield{author}{\bibinfo{person}{Derya Akkaynak} {and} \bibinfo{person}{Tali
  Treibitz}.} \bibinfo{year}{2019}\natexlab{b}.
\newblock \showarticletitle{Sea-thru: A method for removing water from
  underwater images}. In \bibinfo{booktitle}{\emph{Proceedings of the IEEE/CVF
  conference on Computer Vision and Pattern Recognition}}.
  \bibinfo{pages}{1682--1691}.
\newblock


\bibitem[Ancuti et~al\mbox{.}(2012)]%
        {2012fusion}
\bibfield{author}{\bibinfo{person}{Cosmin Ancuti},
  \bibinfo{person}{Codruta~Orniana Ancuti}, \bibinfo{person}{Tom Haber}, {and}
  \bibinfo{person}{Philippe Bekaert}.} \bibinfo{year}{2012}\natexlab{}.
\newblock \showarticletitle{Enhancing underwater images and videos by fusion}.
  In \bibinfo{booktitle}{\emph{2012 IEEE conference on Computer Vision and
  Pattern Recognition}}. IEEE, \bibinfo{pages}{81--88}.
\newblock


\bibitem[Ancuti et~al\mbox{.}(2019)]%
        {ancuti20193C}
\bibfield{author}{\bibinfo{person}{Codruta~O Ancuti}, \bibinfo{person}{Cosmin
  Ancuti}, \bibinfo{person}{Christophe De~Vleeschouwer}, {and}
  \bibinfo{person}{Mateu Sbert}.} \bibinfo{year}{2019}\natexlab{}.
\newblock \showarticletitle{Color channel compensation (3C): A fundamental
  pre-processing step for image enhancement}.
\newblock \bibinfo{journal}{\emph{IEEE Transactions on Image Processing}}
  \bibinfo{volume}{29} (\bibinfo{year}{2019}), \bibinfo{pages}{2653--2665}.
\newblock


\bibitem[Berman et~al\mbox{.}(2020)]%
        {2020SQUID}
\bibfield{author}{\bibinfo{person}{Dana Berman}, \bibinfo{person}{Deborah
  Levy}, \bibinfo{person}{Shai Avidan}, {and} \bibinfo{person}{Tali Treibitz}.}
  \bibinfo{year}{2020}\natexlab{}.
\newblock \showarticletitle{Underwater single image color restoration using
  haze-lines and a new quantitative dataset}.
\newblock \bibinfo{journal}{\emph{IEEE Transactions on Pattern Analysis and
  Machine Intelligence}} \bibinfo{volume}{43}, \bibinfo{number}{8}
  (\bibinfo{year}{2020}), \bibinfo{pages}{2822--2837}.
\newblock


\bibitem[Buchsbaum(1980)]%
        {1990GrayWorld}
\bibfield{author}{\bibinfo{person}{Gershon Buchsbaum}.}
  \bibinfo{year}{1980}\natexlab{}.
\newblock \showarticletitle{A spatial processor model for object colour
  perception}.
\newblock \bibinfo{journal}{\emph{Journal of the Franklin institute}}
  \bibinfo{volume}{310}, \bibinfo{number}{1} (\bibinfo{year}{1980}),
  \bibinfo{pages}{1--26}.
\newblock


\bibitem[Chen et~al\mbox{.}(2020)]%
        {chen2020dynamic}
\bibfield{author}{\bibinfo{person}{Yinpeng Chen}, \bibinfo{person}{Xiyang Dai},
  \bibinfo{person}{Mengchen Liu}, \bibinfo{person}{Dongdong Chen},
  \bibinfo{person}{Lu Yuan}, {and} \bibinfo{person}{Zicheng Liu}.}
  \bibinfo{year}{2020}\natexlab{}.
\newblock \showarticletitle{Dynamic convolution: Attention over convolution
  kernels}. In \bibinfo{booktitle}{\emph{Proceedings of the IEEE/CVF conference
  on Computer Vision and Pattern Recognition}}. \bibinfo{pages}{11030--11039}.
\newblock


\bibitem[Chiang and Chen(2011)]%
        {2012wavelength}
\bibfield{author}{\bibinfo{person}{John~Y Chiang} {and}
  \bibinfo{person}{Ying-Ching Chen}.} \bibinfo{year}{2011}\natexlab{}.
\newblock \bibinfo{title}{Underwater image enhancement by wavelength
  compensation and dehazing}.
\newblock , \bibinfo{numpages}{1756--1769}~pages.
\newblock


\bibitem[Drews et~al\mbox{.}(2013)]%
        {drews2013UDCP}
\bibfield{author}{\bibinfo{person}{Paul Drews}, \bibinfo{person}{Erickson
  Nascimento}, \bibinfo{person}{Filipe Moraes}, \bibinfo{person}{Silvia
  Botelho}, {and} \bibinfo{person}{Mario Campos}.}
  \bibinfo{year}{2013}\natexlab{}.
\newblock \showarticletitle{Transmission estimation in underwater single
  images}. In \bibinfo{booktitle}{\emph{Proceedings of the IEEE international
  conference on computer vision workshops}}. \bibinfo{pages}{825--830}.
\newblock


\bibitem[Fabbri et~al\mbox{.}(2018)]%
        {Fabbri2018UGAN}
\bibfield{author}{\bibinfo{person}{Cameron Fabbri}, \bibinfo{person}{Md~Jahidul
  Islam}, {and} \bibinfo{person}{Junaed Sattar}.}
  \bibinfo{year}{2018}\natexlab{}.
\newblock \showarticletitle{Enhancing Underwater Imagery Using Generative
  Adversarial Networks}. In \bibinfo{booktitle}{\emph{2018 IEEE International
  Conference on Robotics and Automation (ICRA)}}. \bibinfo{pages}{7159--7165}.
\newblock
\urldef\tempurl%
\url{https://doi.org/10.1109/ICRA.2018.8460552}
\showDOI{\tempurl}


\bibitem[Fu et~al\mbox{.}(2022a)]%
        {2022USUIR}
\bibfield{author}{\bibinfo{person}{Zhenqi Fu}, \bibinfo{person}{Huangxing Lin},
  \bibinfo{person}{Yan Yang}, \bibinfo{person}{Shu Chai},
  \bibinfo{person}{Liyan Sun}, \bibinfo{person}{Yue Huang}, {and}
  \bibinfo{person}{Xinghao Ding}.} \bibinfo{year}{2022}\natexlab{a}.
\newblock \showarticletitle{Unsupervised Underwater Image Restoration: From a
  Homology Perspective}. In \bibinfo{booktitle}{\emph{Proceedings of the AAAI
  Conference on Artificial Intelligence}}. \bibinfo{pages}{643--651}.
\newblock


\bibitem[Fu et~al\mbox{.}(2022b)]%
        {fu2022puieNet}
\bibfield{author}{\bibinfo{person}{Zhenqi Fu}, \bibinfo{person}{Wu Wang},
  \bibinfo{person}{Yue Huang}, \bibinfo{person}{Xinghao Ding}, {and}
  \bibinfo{person}{Kai-Kuang Ma}.} \bibinfo{year}{2022}\natexlab{b}.
\newblock \showarticletitle{Uncertainty Inspired Underwater Image Enhancement}.
  In \bibinfo{booktitle}{\emph{Computer Vision--ECCV 2022: 17th European
  Conference, Tel Aviv, Israel, October 23--27, 2022, Proceedings, Part
  XVIII}}. Springer, \bibinfo{pages}{465--482}.
\newblock


\bibitem[Galdran et~al\mbox{.}(2015)]%
        {galdran2015automatic}
\bibfield{author}{\bibinfo{person}{Adrian Galdran}, \bibinfo{person}{David
  Pardo}, \bibinfo{person}{Artzai Pic{\'o}n}, {and} \bibinfo{person}{Aitor
  Alvarez-Gila}.} \bibinfo{year}{2015}\natexlab{}.
\newblock \showarticletitle{Automatic red-channel underwater image
  restoration}.
\newblock \bibinfo{journal}{\emph{Journal of Visual Communication and Image
  Representation}}  \bibinfo{volume}{26} (\bibinfo{year}{2015}),
  \bibinfo{pages}{132--145}.
\newblock


\bibitem[Ghani and Isa(2015)]%
        {ghani2015contrast}
\bibfield{author}{\bibinfo{person}{Ahmad Shahrizan~Abdul Ghani} {and}
  \bibinfo{person}{Nor Ashidi~Mat Isa}.} \bibinfo{year}{2015}\natexlab{}.
\newblock \showarticletitle{Enhancement of low quality underwater image through
  integrated global and local contrast correction}.
\newblock \bibinfo{journal}{\emph{Applied Soft Computing}}
  \bibinfo{volume}{37} (\bibinfo{year}{2015}), \bibinfo{pages}{332--344}.
\newblock


\bibitem[Guo et~al\mbox{.}(2022)]%
        {guo2022URanker}
\bibfield{author}{\bibinfo{person}{Chunle Guo}, \bibinfo{person}{Ruiqi Wu},
  \bibinfo{person}{Xin Jin}, \bibinfo{person}{Linghao Han},
  \bibinfo{person}{Zhi Chai}, \bibinfo{person}{Weidong Zhang}, {and}
  \bibinfo{person}{Chongyi Li}.} \bibinfo{year}{2022}\natexlab{}.
\newblock \showarticletitle{Underwater ranker: learn which is better and how to
  be better}.
\newblock \bibinfo{journal}{\emph{arXiv preprint arXiv:2208.06857}}
  (\bibinfo{year}{2022}).
\newblock


\bibitem[Guo et~al\mbox{.}(2019)]%
        {guo2019DenseGAN}
\bibfield{author}{\bibinfo{person}{Yecai Guo}, \bibinfo{person}{Hanyu Li},
  {and} \bibinfo{person}{Peixian Zhuang}.} \bibinfo{year}{2019}\natexlab{}.
\newblock \showarticletitle{Underwater image enhancement using a multiscale
  dense generative adversarial network}.
\newblock \bibinfo{journal}{\emph{IEEE Journal of Oceanic Engineering}}
  \bibinfo{volume}{45}, \bibinfo{number}{3} (\bibinfo{year}{2019}),
  \bibinfo{pages}{862--870}.
\newblock


\bibitem[Ha et~al\mbox{.}(2016)]%
        {ha2016hypernetworks}
\bibfield{author}{\bibinfo{person}{David Ha}, \bibinfo{person}{Andrew Dai},
  {and} \bibinfo{person}{Quoc~V Le}.} \bibinfo{year}{2016}\natexlab{}.
\newblock \showarticletitle{Hypernetworks}.
\newblock \bibinfo{journal}{\emph{arXiv preprint arXiv:1609.09106}}
  (\bibinfo{year}{2016}).
\newblock


\bibitem[Huang et~al\mbox{.}(2023)]%
        {huang2023Semi-UIR}
\bibfield{author}{\bibinfo{person}{Shirui Huang}, \bibinfo{person}{Keyan Wang},
  \bibinfo{person}{Huan Liu}, \bibinfo{person}{Jun Chen}, {and}
  \bibinfo{person}{Yunsong Li}.} \bibinfo{year}{2023}\natexlab{}.
\newblock \showarticletitle{Contrastive Semi-supervised Learning for Underwater
  Image Restoration via Reliable Bank}.
\newblock \bibinfo{journal}{\emph{arXiv preprint arXiv:2303.09101}}
  (\bibinfo{year}{2023}).
\newblock


\bibitem[Iqbal et~al\mbox{.}(2010)]%
        {2010pixel-range-stretching}
\bibfield{author}{\bibinfo{person}{Kashif Iqbal}, \bibinfo{person}{Michael
  Odetayo}, \bibinfo{person}{Anne James}, \bibinfo{person}{Rosalina~Abdul
  Salam}, {and} \bibinfo{person}{Abdullah Zawawi~Hj Talib}.}
  \bibinfo{year}{2010}\natexlab{}.
\newblock \showarticletitle{Enhancing the low quality images using unsupervised
  colour correction method}. In \bibinfo{booktitle}{\emph{2010 IEEE
  International Conference on Systems, Man and Cybernetics}}. IEEE,
  \bibinfo{pages}{1703--1709}.
\newblock


\bibitem[Islam et~al\mbox{.}(2020a)]%
        {2020UFO}
\bibfield{author}{\bibinfo{person}{Md~Jahidul Islam}, \bibinfo{person}{Peigen
  Luo}, {and} \bibinfo{person}{Junaed Sattar}.}
  \bibinfo{year}{2020}\natexlab{a}.
\newblock \showarticletitle{Simultaneous enhancement and super-resolution of
  underwater imagery for improved visual perception}.
\newblock \bibinfo{journal}{\emph{arXiv preprint arXiv:2002.01155}}
  (\bibinfo{year}{2020}).
\newblock


\bibitem[Islam et~al\mbox{.}(2020b)]%
        {Islam2020FUnIE-GAN}
\bibfield{author}{\bibinfo{person}{Md~Jahidul Islam}, \bibinfo{person}{Youya
  Xia}, {and} \bibinfo{person}{Junaed Sattar}.}
  \bibinfo{year}{2020}\natexlab{b}.
\newblock \showarticletitle{Fast Underwater Image Enhancement for Improved
  Visual Perception}.
\newblock \bibinfo{journal}{\emph{IEEE Robotics and Automation Letters}}
  \bibinfo{volume}{5}, \bibinfo{number}{2} (\bibinfo{year}{2020}),
  \bibinfo{pages}{3227--3234}.
\newblock


\bibitem[Islam et~al\mbox{.}(2020c)]%
        {2020EUVP}
\bibfield{author}{\bibinfo{person}{Md~Jahidul Islam}, \bibinfo{person}{Youya
  Xia}, {and} \bibinfo{person}{Junaed Sattar}.}
  \bibinfo{year}{2020}\natexlab{c}.
\newblock \showarticletitle{Fast underwater image enhancement for improved
  visual perception}.
\newblock \bibinfo{journal}{\emph{IEEE Robotics and Automation Letters}}
  \bibinfo{volume}{5}, \bibinfo{number}{2} (\bibinfo{year}{2020}),
  \bibinfo{pages}{3227--3234}.
\newblock


\bibitem[Jiang et~al\mbox{.}(2022a)]%
        {jiang2022topal}
\bibfield{author}{\bibinfo{person}{Zhiying Jiang}, \bibinfo{person}{Zhuoxiao
  Li}, \bibinfo{person}{Shuzhou Yang}, \bibinfo{person}{Xin Fan}, {and}
  \bibinfo{person}{Risheng Liu}.} \bibinfo{year}{2022}\natexlab{a}.
\newblock \showarticletitle{Target Oriented Perceptual Adversarial Fusion
  Network for Underwater Image Enhancement}.
\newblock \bibinfo{journal}{\emph{IEEE Transactions on Circuits and Systems for
  Video Technology}} (\bibinfo{year}{2022}).
\newblock


\bibitem[Jiang et~al\mbox{.}(2022b)]%
        {jiang2022bilevel}
\bibfield{author}{\bibinfo{person}{Zhiying Jiang}, \bibinfo{person}{Zengxi
  Zhang}, \bibinfo{person}{Yiyao Yu}, {and} \bibinfo{person}{Risheng Liu}.}
  \bibinfo{year}{2022}\natexlab{b}.
\newblock \showarticletitle{Bilevel modeling investigated generative
  adversarial framework for image restoration}.
\newblock \bibinfo{journal}{\emph{The Visual Computer}} (\bibinfo{year}{2022}),
  \bibinfo{pages}{1--13}.
\newblock


\bibitem[Land(1977)]%
        {1977MaxRGB}
\bibfield{author}{\bibinfo{person}{Edwin~H Land}.}
  \bibinfo{year}{1977}\natexlab{}.
\newblock \showarticletitle{The retinex theory of color vision}.
\newblock \bibinfo{journal}{\emph{Scientific american}} \bibinfo{volume}{237},
  \bibinfo{number}{6} (\bibinfo{year}{1977}), \bibinfo{pages}{108--129}.
\newblock


\bibitem[Li et~al\mbox{.}(2021)]%
        {li2021Ucolor}
\bibfield{author}{\bibinfo{person}{Chongyi Li}, \bibinfo{person}{Saeed Anwar},
  \bibinfo{person}{Junhui Hou}, \bibinfo{person}{Runmin Cong},
  \bibinfo{person}{Chunle Guo}, {and} \bibinfo{person}{Wenqi Ren}.}
  \bibinfo{year}{2021}\natexlab{}.
\newblock \showarticletitle{Underwater image enhancement via medium
  transmission-guided multi-color space embedding}.
\newblock \bibinfo{journal}{\emph{IEEE Transactions on Image Processing}}
  \bibinfo{volume}{30} (\bibinfo{year}{2021}), \bibinfo{pages}{4985--5000}.
\newblock


\bibitem[Li et~al\mbox{.}(2020)]%
        {li2020UWCNN}
\bibfield{author}{\bibinfo{person}{Chongyi Li}, \bibinfo{person}{Saeed Anwar},
  {and} \bibinfo{person}{Fatih Porikli}.} \bibinfo{year}{2020}\natexlab{}.
\newblock \showarticletitle{Underwater scene prior inspired deep underwater
  image and video enhancement}.
\newblock \bibinfo{journal}{\emph{Pattern Recognition}}  \bibinfo{volume}{98}
  (\bibinfo{year}{2020}), \bibinfo{pages}{107038}.
\newblock


\bibitem[Li et~al\mbox{.}(2019a)]%
        {li2019WaterNet-UIEB-C60}
\bibfield{author}{\bibinfo{person}{Chongyi Li}, \bibinfo{person}{Chunle Guo},
  \bibinfo{person}{Wenqi Ren}, \bibinfo{person}{Runmin Cong},
  \bibinfo{person}{Junhui Hou}, \bibinfo{person}{Sam Kwong}, {and}
  \bibinfo{person}{Dacheng Tao}.} \bibinfo{year}{2019}\natexlab{a}.
\newblock \showarticletitle{An underwater image enhancement benchmark dataset
  and beyond}.
\newblock \bibinfo{journal}{\emph{IEEE Transactions on Image Processing}}
  \bibinfo{volume}{29} (\bibinfo{year}{2019}), \bibinfo{pages}{4376--4389}.
\newblock


\bibitem[Li et~al\mbox{.}(2016)]%
        {2016histogram_distribution}
\bibfield{author}{\bibinfo{person}{Chong-Yi Li}, \bibinfo{person}{Ji-Chang
  Guo}, \bibinfo{person}{Run-Min Cong}, \bibinfo{person}{Yan-Wei Pang}, {and}
  \bibinfo{person}{Bo Wang}.} \bibinfo{year}{2016}\natexlab{}.
\newblock \showarticletitle{Underwater image enhancement by dehazing with
  minimum information loss and histogram distribution prior}.
\newblock \bibinfo{journal}{\emph{IEEE Transactions on Image Processing}}
  \bibinfo{volume}{25}, \bibinfo{number}{12} (\bibinfo{year}{2016}),
  \bibinfo{pages}{5664--5677}.
\newblock


\bibitem[Li et~al\mbox{.}(2019b)]%
        {li2019FGAN}
\bibfield{author}{\bibinfo{person}{Hanyu Li}, \bibinfo{person}{Jingjing Li},
  {and} \bibinfo{person}{Wei Wang}.} \bibinfo{year}{2019}\natexlab{b}.
\newblock \showarticletitle{A fusion adversarial underwater image enhancement
  network with a public test dataset}.
\newblock \bibinfo{journal}{\emph{arXiv preprint arXiv:1906.06819}}
  (\bibinfo{year}{2019}).
\newblock


\bibitem[Lin et~al\mbox{.}(2021)]%
        {lin2021global}
\bibfield{author}{\bibinfo{person}{Runjia Lin}, \bibinfo{person}{Jinyuan Liu},
  \bibinfo{person}{Risheng Liu}, {and} \bibinfo{person}{Xin Fan}.}
  \bibinfo{year}{2021}\natexlab{}.
\newblock \showarticletitle{Global structure-guided learning framework for
  underwater image enhancement}.
\newblock \bibinfo{journal}{\emph{The Visual Computer}} (\bibinfo{year}{2021}),
  \bibinfo{pages}{1--16}.
\newblock


\bibitem[Lin et~al\mbox{.}(2020)]%
        {lin2020MTL}
\bibfield{author}{\bibinfo{person}{Xi Lin}, \bibinfo{person}{Zhiyuan Yang},
  \bibinfo{person}{Qingfu Zhang}, {and} \bibinfo{person}{Sam Kwong}.}
  \bibinfo{year}{2020}\natexlab{}.
\newblock \showarticletitle{Controllable pareto multi-task learning}.
\newblock \bibinfo{journal}{\emph{arXiv preprint arXiv:2010.06313}}
  (\bibinfo{year}{2020}).
\newblock


\bibitem[Liu et~al\mbox{.}(2018)]%
        {liu2018learning}
\bibfield{author}{\bibinfo{person}{Risheng Liu}, \bibinfo{person}{Xin Fan},
  \bibinfo{person}{Minjun Hou}, \bibinfo{person}{Zhiying Jiang},
  \bibinfo{person}{Zhongxuan Luo}, {and} \bibinfo{person}{Lei Zhang}.}
  \bibinfo{year}{2018}\natexlab{}.
\newblock \showarticletitle{Learning aggregated transmission propagation
  networks for haze removal and beyond}.
\newblock \bibinfo{journal}{\emph{IEEE transactions on neural networks and
  learning systems}} \bibinfo{volume}{30}, \bibinfo{number}{10}
  (\bibinfo{year}{2018}), \bibinfo{pages}{2973--2986}.
\newblock


\bibitem[Liu et~al\mbox{.}(2020a)]%
        {2020RUIE}
\bibfield{author}{\bibinfo{person}{Risheng Liu}, \bibinfo{person}{Xin Fan},
  \bibinfo{person}{Ming Zhu}, \bibinfo{person}{Minjun Hou}, {and}
  \bibinfo{person}{Zhongxuan Luo}.} \bibinfo{year}{2020}\natexlab{a}.
\newblock \showarticletitle{Real-world underwater enhancement: Challenges,
  benchmarks, and solutions under natural light}.
\newblock \bibinfo{journal}{\emph{IEEE Transactions on Circuits and Systems for
  Video Technology}} \bibinfo{volume}{30}, \bibinfo{number}{12}
  (\bibinfo{year}{2020}), \bibinfo{pages}{4861--4875}.
\newblock


\bibitem[Liu et~al\mbox{.}(2019)]%
        {liu2019compounded}
\bibfield{author}{\bibinfo{person}{Risheng Liu}, \bibinfo{person}{Minjun Hou},
  \bibinfo{person}{Jinyuan Liu}, \bibinfo{person}{Xin Fan}, {and}
  \bibinfo{person}{Zhongxuan Luo}.} \bibinfo{year}{2019}\natexlab{}.
\newblock \showarticletitle{Compounded layer-prior unrolling: A unified
  transmission-based image enhancement framework}. In
  \bibinfo{booktitle}{\emph{2019 IEEE International Conference on Multimedia
  and Expo (ICME)}}. IEEE, \bibinfo{pages}{538--543}.
\newblock


\bibitem[Liu et~al\mbox{.}(2022a)]%
        {liu2022tacl}
\bibfield{author}{\bibinfo{person}{Risheng Liu}, \bibinfo{person}{Zhiying
  Jiang}, \bibinfo{person}{Shuzhou Yang}, {and} \bibinfo{person}{Xin Fan}.}
  \bibinfo{year}{2022}\natexlab{a}.
\newblock \showarticletitle{Twin adversarial contrastive learning for
  underwater image enhancement and beyond}.
\newblock \bibinfo{journal}{\emph{IEEE Transactions on Image Processing}}
  \bibinfo{volume}{31} (\bibinfo{year}{2022}), \bibinfo{pages}{4922--4936}.
\newblock


\bibitem[Liu et~al\mbox{.}(2020b)]%
        {liu2020generic}
\bibfield{author}{\bibinfo{person}{Risheng Liu}, \bibinfo{person}{Pan Mu},
  \bibinfo{person}{Xiaoming Yuan}, \bibinfo{person}{Shangzhi Zeng}, {and}
  \bibinfo{person}{Jin Zhang}.} \bibinfo{year}{2020}\natexlab{b}.
\newblock \showarticletitle{A generic first-order algorithmic framework for
  bi-level programming beyond lower-level singleton}. In
  \bibinfo{booktitle}{\emph{International Conference on Machine Learning}}.
  PMLR, \bibinfo{pages}{6305--6315}.
\newblock


\bibitem[Liu et~al\mbox{.}(2022b)]%
        {liu2022general}
\bibfield{author}{\bibinfo{person}{Risheng Liu}, \bibinfo{person}{Pan Mu},
  \bibinfo{person}{Xiaoming Yuan}, \bibinfo{person}{Shangzhi Zeng}, {and}
  \bibinfo{person}{Jin Zhang}.} \bibinfo{year}{2022}\natexlab{b}.
\newblock \showarticletitle{A general descent aggregation framework for
  gradient-based bi-level optimization}.
\newblock \bibinfo{journal}{\emph{IEEE Transactions on Pattern Analysis and
  Machine Intelligence}} \bibinfo{volume}{45}, \bibinfo{number}{1}
  (\bibinfo{year}{2022}), \bibinfo{pages}{38--57}.
\newblock


\bibitem[Liu et~al\mbox{.}(1995)]%
        {liu1995whitebalance}
\bibfield{author}{\bibinfo{person}{Yung-Cheng Liu}, \bibinfo{person}{Wen-Hsin
  Chan}, {and} \bibinfo{person}{Ye-Quang Chen}.}
  \bibinfo{year}{1995}\natexlab{}.
\newblock \showarticletitle{Automatic white balance for digital still camera}.
\newblock \bibinfo{journal}{\emph{IEEE Transactions on Consumer Electronics}}
  \bibinfo{volume}{41}, \bibinfo{number}{3} (\bibinfo{year}{1995}),
  \bibinfo{pages}{460--466}.
\newblock


\bibitem[Mittal et~al\mbox{.}(2012)]%
        {2012NIQE}
\bibfield{author}{\bibinfo{person}{Anish Mittal}, \bibinfo{person}{Rajiv
  Soundararajan}, {and} \bibinfo{person}{Alan~C Bovik}.}
  \bibinfo{year}{2012}\natexlab{}.
\newblock \showarticletitle{Making a “completely blind” image quality
  analyzer}.
\newblock \bibinfo{journal}{\emph{IEEE Signal processing letters}}
  \bibinfo{volume}{20}, \bibinfo{number}{3} (\bibinfo{year}{2012}),
  \bibinfo{pages}{209--212}.
\newblock


\bibitem[Mu et~al\mbox{.}(2022)]%
        {Mu2022StructureInferredBM}
\bibfield{author}{\bibinfo{person}{Pan Mu}, \bibinfo{person}{Haotian Qian},
  {and} \bibinfo{person}{Cong Bai}.} \bibinfo{year}{2022}\natexlab{}.
\newblock \showarticletitle{Structure-Inferred Bi-level Model for Underwater
  Image Enhancement}.
\newblock \bibinfo{journal}{\emph{Proceedings of the 30th ACM International
  Conference on Multimedia (MM)}} (\bibinfo{year}{2022}).
\newblock


\bibitem[Panetta et~al\mbox{.}(2015)]%
        {2015UIQM}
\bibfield{author}{\bibinfo{person}{Karen Panetta}, \bibinfo{person}{Chen Gao},
  {and} \bibinfo{person}{Sos Agaian}.} \bibinfo{year}{2015}\natexlab{}.
\newblock \showarticletitle{Human-visual-system-inspired underwater image
  quality measures}.
\newblock \bibinfo{journal}{\emph{IEEE Journal of Oceanic Engineering}}
  \bibinfo{volume}{41}, \bibinfo{number}{3} (\bibinfo{year}{2015}),
  \bibinfo{pages}{541--551}.
\newblock


\bibitem[Peng et~al\mbox{.}(2023)]%
        {2021LSUI}
\bibfield{author}{\bibinfo{person}{Lintao Peng}, \bibinfo{person}{Chunli Zhu},
  {and} \bibinfo{person}{Liheng Bian}.} \bibinfo{year}{2023}\natexlab{}.
\newblock \showarticletitle{U-shape transformer for underwater image
  enhancement}. In \bibinfo{booktitle}{\emph{Computer Vision--ECCV 2022
  Workshops: Tel Aviv, Israel, October 23--27, 2022, Proceedings, Part II}}.
  Springer, \bibinfo{pages}{290--307}.
\newblock


\bibitem[Peng and Cosman(2017)]%
        {2017light_absorption}
\bibfield{author}{\bibinfo{person}{Yan-Tsung Peng} {and}
  \bibinfo{person}{Pamela~C Cosman}.} \bibinfo{year}{2017}\natexlab{}.
\newblock \showarticletitle{Underwater image restoration based on image
  blurriness and light absorption}.
\newblock \bibinfo{journal}{\emph{IEEE Transactions on Image Processing}}
  \bibinfo{volume}{26}, \bibinfo{number}{4} (\bibinfo{year}{2017}),
  \bibinfo{pages}{1579--1594}.
\newblock


\bibitem[Pizer(1990)]%
        {pizer1990contrast}
\bibfield{author}{\bibinfo{person}{Stephen~M Pizer}.}
  \bibinfo{year}{1990}\natexlab{}.
\newblock \showarticletitle{Contrast-limited adaptive histogram equalization:
  Speed and effectiveness stephen m. pizer, r. eugene johnston, james p.
  ericksen, bonnie c. yankaskas, keith e. muller medical image display research
  group}. In \bibinfo{booktitle}{\emph{Proceedings of the first conference on
  visualization in biomedical computing, Atlanta, Georgia}},
  Vol.~\bibinfo{volume}{337}. \bibinfo{pages}{1}.
\newblock


\bibitem[Qian et~al\mbox{.}(2022)]%
        {qian2022real}
\bibfield{author}{\bibinfo{person}{Haotian Qian}, \bibinfo{person}{Wentao
  Tong}, \bibinfo{person}{Pan Mu}, \bibinfo{person}{Zheyuan Liu}, {and}
  \bibinfo{person}{Hanning Xu}.} \bibinfo{year}{2022}\natexlab{}.
\newblock \showarticletitle{Real-world Underwater Image Enhancement via
  Degradation-aware Dynamic Network}. In \bibinfo{booktitle}{\emph{PRICAI 2022:
  Trends in Artificial Intelligence: 19th Pacific Rim International Conference
  on Artificial Intelligence, PRICAI 2022, Shanghai, China, November 10--13,
  2022, Proceedings, Part III}}. Springer, \bibinfo{pages}{530--541}.
\newblock


\bibitem[Rahman et~al\mbox{.}(1996)]%
        {rahman1996retinex}
\bibfield{author}{\bibinfo{person}{Zia-ur Rahman}, \bibinfo{person}{Daniel~J
  Jobson}, {and} \bibinfo{person}{Glenn~A Woodell}.}
  \bibinfo{year}{1996}\natexlab{}.
\newblock \showarticletitle{Multi-scale retinex for color image enhancement}.
  In \bibinfo{booktitle}{\emph{Proceedings of 3rd IEEE International Conference
  on Image Processing}}, Vol.~\bibinfo{volume}{3}. IEEE,
  \bibinfo{pages}{1003--1006}.
\newblock


\bibitem[Wang et~al\mbox{.}(2017)]%
        {2017attenuation-curve}
\bibfield{author}{\bibinfo{person}{Yi Wang}, \bibinfo{person}{Hui Liu}, {and}
  \bibinfo{person}{Lap-Pui Chau}.} \bibinfo{year}{2017}\natexlab{}.
\newblock \showarticletitle{Single underwater image restoration using adaptive
  attenuation-curve prior}.
\newblock \bibinfo{journal}{\emph{IEEE Transactions on Circuits and Systems I:
  Regular Papers}} \bibinfo{volume}{65}, \bibinfo{number}{3}
  (\bibinfo{year}{2017}), \bibinfo{pages}{992--1002}.
\newblock


\bibitem[Yang and Sowmya(2015)]%
        {2015UCIQE}
\bibfield{author}{\bibinfo{person}{Miao Yang} {and} \bibinfo{person}{Arcot
  Sowmya}.} \bibinfo{year}{2015}\natexlab{}.
\newblock \showarticletitle{An underwater color image quality evaluation
  metric}.
\newblock \bibinfo{journal}{\emph{IEEE Transactions on Image Processing}}
  \bibinfo{volume}{24}, \bibinfo{number}{12} (\bibinfo{year}{2015}),
  \bibinfo{pages}{6062--6071}.
\newblock


\bibitem[Yin et~al\mbox{.}(2022)]%
        {yin2022conditional}
\bibfield{author}{\bibinfo{person}{Guanghao Yin}, \bibinfo{person}{Wei Wang},
  \bibinfo{person}{Zehuan Yuan}, \bibinfo{person}{Wei Ji},
  \bibinfo{person}{Dongdong Yu}, \bibinfo{person}{Shouqian Sun},
  \bibinfo{person}{Tat-Seng Chua}, {and} \bibinfo{person}{Changhu Wang}.}
  \bibinfo{year}{2022}\natexlab{}.
\newblock \showarticletitle{Conditional hyper-network for blind
  super-resolution with multiple degradations}.
\newblock \bibinfo{journal}{\emph{IEEE Transactions on Image Processing}}
  \bibinfo{volume}{31} (\bibinfo{year}{2022}), \bibinfo{pages}{3949--3960}.
\newblock


\bibitem[Zhang et~al\mbox{.}(2017)]%
        {2017multi-scale_retinex}
\bibfield{author}{\bibinfo{person}{Shu Zhang}, \bibinfo{person}{Ting Wang},
  \bibinfo{person}{Junyu Dong}, {and} \bibinfo{person}{Hui Yu}.}
  \bibinfo{year}{2017}\natexlab{}.
\newblock \showarticletitle{Underwater image enhancement via extended
  multi-scale Retinex}.
\newblock \bibinfo{journal}{\emph{Neurocomputing}}  \bibinfo{volume}{245}
  (\bibinfo{year}{2017}), \bibinfo{pages}{1--9}.
\newblock


\end{thebibliography}

%

\end{document}